\documentclass[11pt, a4paper, copyright, gdm]{google}

\usepackage[numbers, sort&compress]{natbib}
\uselogo{} 

\title{The Cartesian Shortcut: Re-evaluate Vision Reasoning in Polar Coordinate Space}

\correspondingauthor{xiahu@google.com}

\usepackage[utf8]{inputenc} 
\usepackage[T1]{fontenc}    
\usepackage{hyperref}       
\usepackage{url}            
\usepackage{booktabs}       
\usepackage{amsfonts}       
\usepackage{nicefrac}       
\usepackage{microtype}      
\usepackage{xcolor}         

\usepackage{graphicx}       
\usepackage{amsmath}        
\usepackage{amssymb}

\usepackage{booktabs}
\usepackage{enumitem}
\usepackage{multirow}
\usepackage{makecell}
\usepackage{graphicx}
\usepackage{xspace}
\usepackage{wrapfig}

\usepackage{amssymb}
\usepackage{pifont}

\newcommand{\cmark}{\ding{51}}  
\newcommand{\xmark}{\ding{55}}  

\newcommand{\polar}{Polaris-Bench\xspace} 
\newcommand{\mllm}{MLLMs}
\newcommand{\dt}[1]{\enskip{\tiny\color{darkgray}(#1)}}

\newcommand{\howardzhou}[1]{\textcolor{green}{[#1]}}
\newcommand{\nop}[1]{}
\newcommand{\eg}{e.g.,\xspace}
\newcommand{\ie}{i.e.,\xspace}

\reportnumber{0001} 

\renewcommand{\today}{2026-05-10}

\author[1]{Xia Hu}
\author[1]{Zhenrui Yue}
\author[1]{Brian Potetz}
\author[1]{Howard Zhou}
\author[1,2]{Leonidas Guibas}
\author[3]{Chun-Ta Lu}
\author[1]{Zhicheng Wang}

\affil[1]{\thepa{}{}}
\affil[2]{Stanford University}
\affil[3]{Google Research}

\begin{abstract}
As current Multimodal Large Language Models rapidly saturate canonical visual reasoning benchmarks, a key question emerges: do these strong scores genuinely reflect robust visual understanding? We identify a pervasive vulnerability, the \textbf{Cartesian Shortcut}: models frequently discretize the orthogonal grid-based layouts prevalent in visual reasoning benchmarks into explicit textual coordinates, offloading reasoning from visual perception to text-based deduction. To re-evaluate visual reasoning when this shortcut is unavailable, we introduce \textbf{Polaris-Bench}, which re-formulates 53 visual reasoning tasks in Polar coordinate space with paired Cartesian counterparts that preserve task semantics, disrupting the orthogonal structure that models exploit. Comprehensive evaluation across $14$ state-of-the-art MLLMs reveals that frontier models achieving $69$--$83\%$ on Cartesian layouts collapse to $31$--$39\%$ on Polar equivalents. Moreover, thinking gains largely vanish on Polar layouts, prompting interventions fail to close the gap, and comparable drops arise on other non-orthogonal layouts. These findings reveal that current MLLMs' visual reasoning performance is strongly coupled to orthogonal grid structure, a fragility that is consistent across model families but far smaller in humans, who maintain 88.8\% accuracy on Polar layouts.
The benchmark, evaluation suite, and leaderboard are publicly available at \url{https://google-deepmind.github.io/polaris-bench}.
\end{abstract}

\begin{document}

\maketitle

\section{Introduction}

Recent Multimodal Large Language Models (MLLMs)~\citep{gemini3pro2025,singh2025openai,anthropic2024claude} achieve remarkable, often human-level performance across canonical visual reasoning benchmarks~\citep{yue2025mmmu,chen2025megabench,hao2025emma,tong2024cambrian}. 
However, as these datasets rapidly saturate, a critical question emerges: have models genuinely acquired visual reasoning abilities, or are they merely exploiting text-based heuristics~\citep{asadi2026mirage,xu2025more,chen2024we,kanade-ganu-2026-see}? 
In this study, we expose a pervasive systemic vulnerability that 
creates an illusion of profound visual capability: the \textbf{Cartesian Shortcut}.


\begin{figure}[t]
    \centering
    \includegraphics[trim=1.4cm 2.2cm 1.8cm 2.2cm, clip, width=\textwidth]{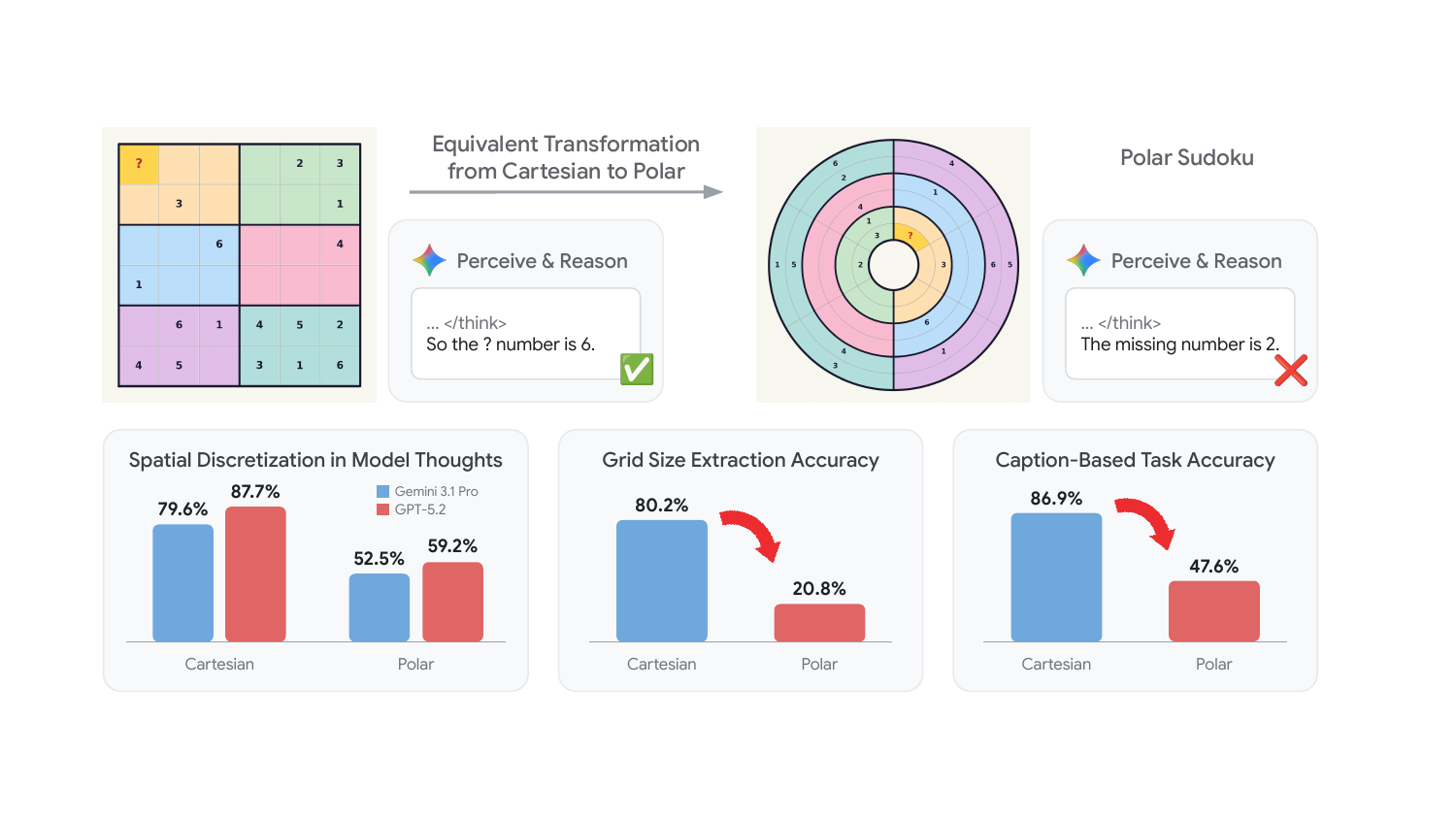}
    \caption{Illustration and empirical evidence of the Cartesian Shortcut. \textit{(Top)} An equivalent sudoku question in Cartesian and Polar spaces (row to ring, column to angular sector). The model correctly solves the orthogonal layout but fails on its polar counterpart.\nop{\howardzhou{Mention row 1 maps to ring 1 (in --> out), column 1 maps to sector 1?}} \textit{(Bottom Left)} The percentage of intermediate CoT that explicitly invokes textual coordinates. Gemini-3.1-Pro and GPT-5.2 both exhibit high frequencies on Cartesian layouts. \textit{(Bottom Middle \& Right)} Two-stage perception and reasoning analysis on Gemini-3.1-Pro: grid size extraction accuracy estimates visual perception capability~(\textit{Middle}), while task accuracy is evaluated on correctly perceived instances~(\textit{Right}). High Cartesian accuracy indicates the effectiveness of the shortcut, while the sharp decline on Polar layouts is consistent with the corruption of orthogonal structure.}
    \label{fig:cartesian_shortcut}
\end{figure}

Orthogonal grid-based layouts, which underlie many prominent visual reasoning benchmarks, can be readily discretized into explicit textual coordinates. 
We find that {\mllm} systematically exploit this property: across a synthetic subset of 9 prominent benchmarks~(\eg \citet{hao2025emma,jia2026omnispatial}), frontier model explicitly invoke Cartesian coordinates (e.g., ``row 2'', ``(x,y)'') in over $56\%$ of intermediate Chain-of-Thought~\citep{wei2022cot} reasoning, converting visual layouts into textual representations to assist problem-solving. 
By substantially offloading reasoning from visual perception to text-based deduction, this shortcut significantly confounds the evaluation of visual reasoning on grid-based testbeds.

Crucially, performance drops sharply once the orthogonal layout is distorted: on a logically identical Sudoku puzzle rendered in Polar coordinates, frontier models that correctly solve the Cartesian version fail on its coordinate equivalent~(Figure~\ref{fig:cartesian_shortcut}).
Drawing on this observation, we introduce \textbf{{\polar}}, comprising 53 procedurally generated visual reasoning tasks in Polar coordinate space, each accompanied by a Cartesian counterpart as controlled reference under consistent logical constraints.
The Cartesian-to-Polar coordinate transformation breaks the orthogonal structure underlying the Cartesian shortcut, disrupting models' ability to easily discretize visual layouts into textual coordinates. 
This design provides a controlled diagnostic testbed to re-evaluate visual reasoning under coordinate transformation, with the consistent logical constraints enabling direct cross-topology evaluation and analysis.

We comprehensively evaluate {\polar} across 14 state-of-the-art {\mllm} spanning both proprietary and open-source families. 
The results reveal a striking and consistent pattern that frontier models achieving $69$--$83\%$ accuracy on Cartesian layouts see their performance collapse to $31$--$39\%$ on Polar equivalents, with the largest degradation approaching 47 points. 
Comparatively, human performance drop only 5.7 points on Polar layout. 
Critically, the degradation persists even on tasks where the transformation preserves complete logical equivalence~(\ie identical rules, constraints, and ground truths), confirming that the coordinate transformation alone suffices to disrupt model reasoning.
Moreover, while enabling high reasoning mode yields substantial accuracy gains on Cartesian layouts, these gains are severely diminished on Polar equivalents. 
Further analyses show that comparable drops arise on non-Polar layouts (hexagonal tilings, wave and arc deformations), and the gap could not be recovered by prompting intervention~(conversion hints, in-context examples). 

Our main contributions are:
\begin{itemize}[leftmargin=*,itemsep=0pt,topsep=0pt]
    \item We identify and empirically validate the Cartesian Shortcut, whereby {\mllm} systematically discretize orthogonal grid-based layouts into explicit textual coordinates, offloading visual reasoning onto text-based deduction and confounding existing visual reasoning evaluation.
    \item We introduce {\polar}, which re-formulates 53 visual reasoning tasks in Polar coordinate space to systematically disrupt the Cartesian Shortcut, with paired Cartesian counterparts enabling cross-topology evaluation.
    \item Our comprehensive evaluation across 14 state-of-the-art MLLMs reveals drastic and consistent performance collapse on Polar equivalents, with degradation persisting across diverse model families, scales, and reasoning configurations, while remaining far smaller in human participants.
\end{itemize}

\begin{figure}[t]
    \centering
    \includegraphics[width=\linewidth]{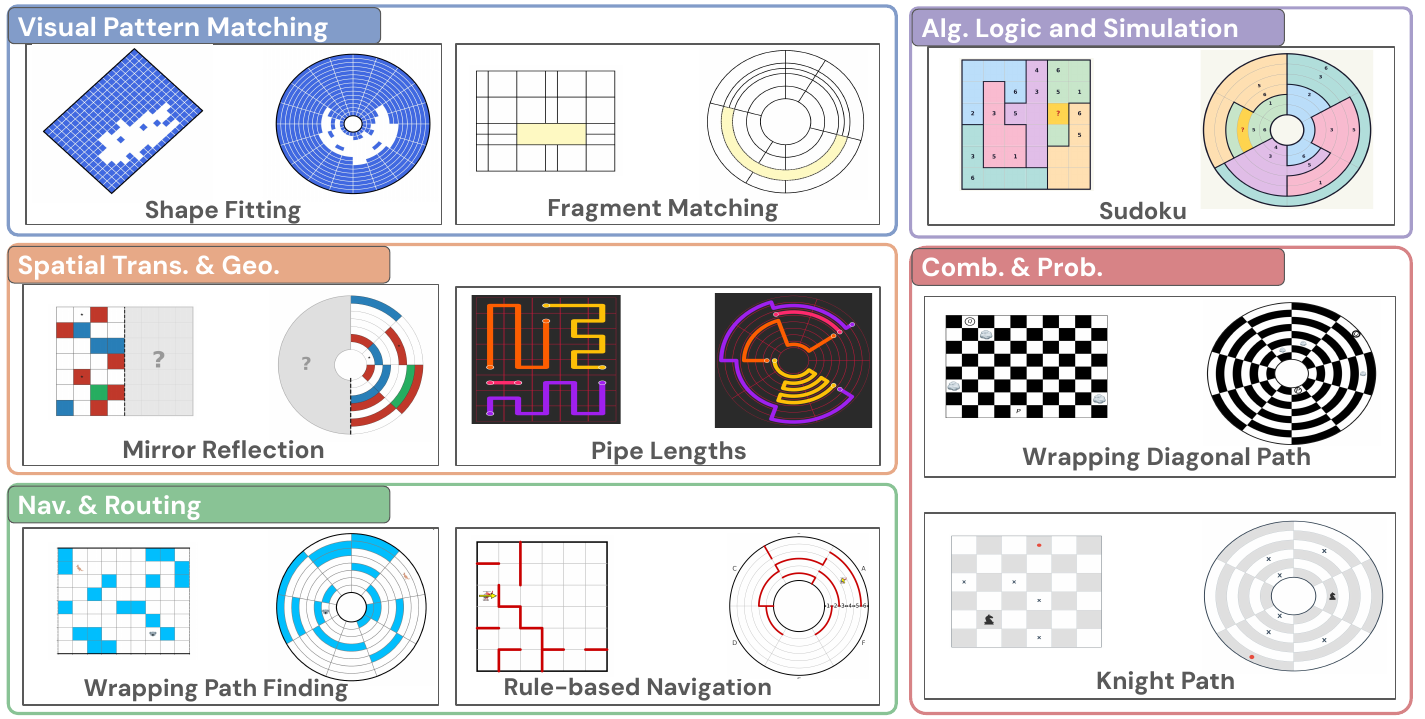}
    \caption{Representative examples for each of the five core taxonomies in {\polar}.}
    \label{fig:representative_examples}
\end{figure}

\section{Rethinking vision reasoning: the Cartesian Shortcut}
\label{sec:cartesian_shortcut}


Current {\mllm} excel on vision reasoning benchmarks, where many of these tasks are built upon synthetic, grid-based layouts~\citep{tang2025grasp,ren2025vgrp}. To investigate how models approach them, we analyzed over 3,800 synthetic-based questions across 9 prominent visual reasoning benchmarks~(\eg \citet{hao2025emma,jia2026omnispatial}). 
Our Chain-of-Thought (CoT)~\citep{wei2022cot} analysis reveals a striking pattern: in over $56\%$ examples, frontier models (\eg Gemini-3-Flash) explicitly discretize the input image into Cartesian coordinates, generating text such as ``\textit{starting at (2,3), moving to (3,5)}.'' 
By translating these visual layouts into discrete matrices, models heavily offload visual logic onto text-based deductive reasoning, reducing the need for persistent visual grounding. We define this phenomenon as \textbf{Cartesian Shortcut}.
\nop{\howardzhou{here we use bold, we should be consistent}.}
While this behavior demonstrates robust deductive capability, it exposes a critical flaw in current evaluations: models can leverage textual discretization to assist visual reasoning, projecting an inflated estimate of visual capability.

To further empirically validate this Cartesian Shortcut, we analyzed model behavior on our paired Cartesian-Polar benchmark. 
Intermediate CoT from frontier models reveal a deep dependency on textual discretization, explicitly invoking Cartesian coordinates in $80\%$-88\% of Cartesian examples, and still attempting to in over half of Polar ones (Figure~\ref{fig:cartesian_shortcut}(left)). 
We further probed this dependency by a two-stage experiment on Gemini-3.1-Pro~\citep{gemini3pro2025}. 
The model first generates a descriptive caption, then solves the task based on the caption. 
As shown in Figure~\ref{fig:cartesian_shortcut}(middle, right), the model 
achieves $80.2\%$ accuracy on visual layout perception, estimated via grid 
dimension extraction from model-generated captions, and task accuracy on the 
subset of correctly perceived instances reaches $86.9\%$.
This indicates that Cartesian Shortcut is not merely a behavioral pattern but a highly effective problem-solving strategy for\nop{current} {\mllm}.
This effectiveness raises a fundamental concern: benchmarks built on orthogonal grids may systematically overestimate visual reasoning capabilities.

Drawing on the duality of coordinate systems~\citep{arfken2011mathematical,zetzsche1999atoms}, we introduce {\polar} to dismantle the Cartesian shortcut. 
{\polar} re-formulates visual reasoning tasks in Polar coordinate space, pairing each instance with a Cartesian counterpart as controlled reference. 
This mapping is a visual and coordinate transformation that preserves the underlying logical constraints, rather than a mathematical translation.
The coordinate transformation 
disrupts the orthogonal grid structure that enables the Cartesian Shortcut. 
Applying the same two-stage analysis to Polar layouts reveals a stark contrast: visual layout perception accuracy falls to $20.8\%$, and task accuracy on correctly perceived instances to $47.6\%$~(Figure~\ref{fig:cartesian_shortcut}). 
This is consistent with the observation: on Polar layouts, models can neither reliably discretize the grid into text nor maintain high accuracy.

\section{{\polar}}
\label{sec:benchmark}


\subsection{Benchmark overview}
\label{subsec:bench_overview}

\textbf{{\polar}} comprises 53 procedurally generated visual reasoning tasks, each re-formulated in Polar coordinate space with paired Cartesian counterparts as controlled reference, yielding over 10,600 core test cases. 
This paired design is central to our methodology: by presenting each task in two coordinate systems that share identical rules, question context, and identical ground truths for the majority, 
we construct a controlled diagnostic testbed where 
performance differences directly measure sensitivity to the coordinate transformation.
To ensure valid spatial logic across coordinate systems, physical concepts are contextually adapted~(\eg Cartesian grid distances are translated to unit edge counting in Polar space).
All task outputs are constrained to deterministic formats (multiple-choice options, exact digits, coordinates, string, and lists) to facilitate rigorous automated evaluation.


\begin{wrapfigure}[18]{r}{0.4\textwidth}
    \centering
    \includegraphics[trim=0.3cm 0.3cm 0.3cm 0.3cm, clip, width=\linewidth]{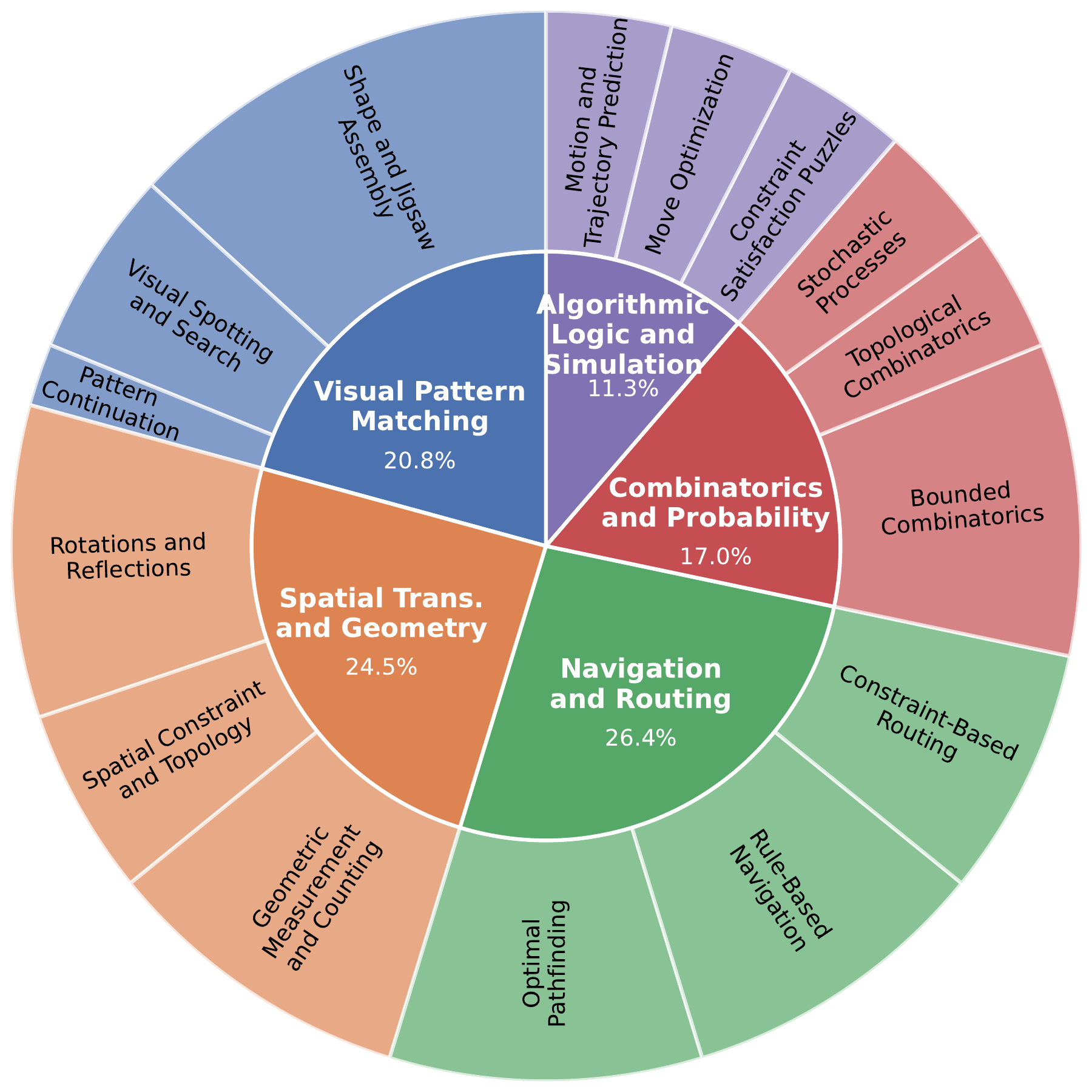}
    \caption{Distribution of 53 tasks across the five core categories.}
    \label{fig:taxonomy_sunhurst}
\end{wrapfigure}
\textbf{Task Taxonomy.} To evaluate vision reasoning across diverse cognitive dimensions, we organize the $53$ tasks into five categories based on their core cognitive operation (Figure~\ref{fig:taxonomy_sunhurst}). 
\textit{Visual Pattern Matching} (11 tasks) evaluates pattern recognition and structural discrimination within organized visual layouts. 
\textit{Spatial Transformation and Geometry} (13 tasks) evaluates geometric transformation reasoning (e.g., rotation, reflection, folding) and quantitative property inference. 
\textit{Navigation and Routing} (14 tasks) evaluates sequential positional state tracking, requiring models to execute or trace multi-step movements under environmental constraints. 
\textit{Combinatorics and Probability} (9 tasks) evaluates systematic combinatorial enumeration and probabilistic reasoning. 
\textit{Algorithmic Logic and Simulation} (6 tasks) evaluates constraint satisfaction and deterministic forward simulation. This categorization enables fine-grained analysis of how the Cartesian Shortcut differentially affects distinct reasoning capabilities.

\textbf{Topology Alignment.}
The paired Cartesian-Polar evaluation introduces a design constraint: the angular dimension of Polar coordinates is inherently cyclic, with $360°$ wrapping seamlessly to $0°$. This cylindrical connectivity can alter valid solutions for adjacency-sensitive tasks, creating a potential confound between perceptual distortion and genuine topological divergence.

We address this through two boundary conditions. 
Under the \textit{bounded} condition, boundary rings and radial seams act as impenetrable barriers, producing a closed Polar grid coordinate equivalent to its Cartesian counterpart with identical ground truths. 
Under the \textit{wrapping} condition, we preserve the natural cyclic connectivity of Polar space, retaining its intrinsic topological properties as an additional dimension of the evaluation; 
for a subset of these tasks, we additionally introduce cyclic boundaries in the Cartesian version~(\eg exit left will enter right) to both align ground truths and examine whether models can handle wrapping within standard grid layouts. 
This yields two alignment categories: the large majority are \textit{fully aligned} where coordinate transformation is the only variance, while a deliberately preserved subset remains \textit{partially aligned}, where cylindrical connectivity alters the solution space. 
This stratification enables controlled ablation: gaps on fully aligned tasks reflect the coordinate transformation itself, with rules and ground truths held fixed;
while gaps on partially aligned tasks additionally capture failures under altered topology.

\subsection{Benchmark construction}
\label{subsec:construction}

\begin{figure}[h]
    \centering
    \includegraphics[trim=2.5cm 5.2cm 2.5cm 5.2cm, clip, width=0.9\textwidth]{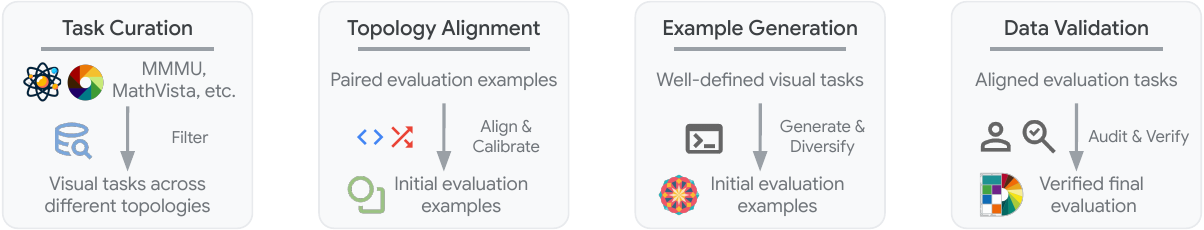}
    \caption{An overview of the construction process of the {\polar}.}
    \label{fig:process}
\end{figure}

Given the scarcity of Polar tasks in the wild, we systematically generated our benchmark data instead of collecting it manually. 
Specifically, we synthesize {\polar} through a four-stage pipeline illustrated in Figure~\ref{fig:process}: task curation from existing benchmarks, task design with cross-topology alignment, \nop{procedural} example generation, and multi-stage quality assurance. Key facts are provided below with comprehensive details in Appendix.

\textbf{Task Curation.}
We first surveyed prominent visual reasoning benchmarks (\eg \nop{MMMU Pro, MathVista, EMMA}\citet{yue2025mmmu, lu2024mathvista, hao2025emma}) to identify tasks that heavily rely on grid-based spatial logic. 
Using a combination of LLM-assisted filtering and manual review, we selected tasks whose core reasoning logic can be faithfully re-formulated across different coordinate topologies(Section~\ref{subsec:bench_overview}).

\textbf{Task Design and Cross-Topology Alignment.}
For each curated task, we designed paired Cartesian and Polar implementations that preserve equivalent evaluation conditions. 
This involves three interrelated design considerations.
First, \emph{logical alignment}: for each task, we analyzed and designed the wrapping behavior and topology alignment to ensure consistent evaluation across coordinate systems (Section~\ref{subsec:bench_overview}).
Second, \emph{visual calibration}: since the Polar mapping introduces nonlinear distortion, we calibrate the rendering of each task so that size-sensitive visual elements remain as comparable as possible across the two layouts, ensuring that performance differences stem from the coordinate transformation rather than perceptual disadvantage.
Third, \emph{narrative design}: beyond adapting existing task formats, we substantially redesigned task rules, visual presentations, and problem structures to introduce richer reasoning demands, and contextualized them within diverse textual scenarios (\eg robotic navigation, physics simulations).

\textbf{Procedural Generation.}
Each task is implemented as a self-contained generative script capable of producing unlimited unique instances. 
We heavily randomize both visual and logical axes~(\eg grid dimensions, starting positions, path topologies, distractors, color palettes), while enforcing programmatic solvability checks to prevent degenerate configurations. 
Each task contributes 100 paired Cartesian–Polar instances to the final benchmark.

\textbf{Quality Assurance and Human Validation.}
We apply a multi-stage verification pipeline to ensure the correctness and clarity of generated data.
(1)~We iteratively refined the rendering to ensure visual clarity and legibility across both Cartesian and Polar layouts.
(2)~The authors verified task logic through collaborative review and by independently spot-checking and solving sampled instances of every tasks.
(3)~We employed LLM-assisted audits of the procedural code to identify unhandled edge cases.
Steps (1)--(3) were repeated until no further issues identified.
(4)~As a final safeguard, we conducted a human validation study with independent raters manually solved 20 randomly sampled Cartesian--Polar pairs per task, confirming the correctness and quality of the final benchmark data.

\section{Experiments}
\label{sec:experiments}


\subsection{Experimental setup}

\textbf{Evaluated Models.}
To comprehensively evaluate visual reasoning capabilities, we test a diverse suite of 14 state-of-the-art MLLMs spanning both proprietary and open-source families.
For proprietary models, we include the \textit{Gemini family}~\citep{gemini3pro2025}, \textit{GPT-5.2}~\citep{singh2025openai}, \textit{Claude-Sonnet-4.6}~\citep{anthropic2024claude}, and \textit{Grok-4 family}~\citep{grok4modelcard2025}.
For open-source models, we evaluate \textit{Gemma-4-31B}, \textit{Gemma-4-26B}~\citep{gemma4}, \textit{Kimi-k2.5}~\citep{team2026kimi}, \textit{Qwen-3.5-397B-A17B}~\citep{qwen3.5}, and \textit{Mistral-Small-3.1-24B}~\citep{mistral2025small31}.
This diverse model selection spans different model families at varying architectures, scales, and training recipes.
All models are accessed via public APIs with default parameter settings and maximum supported context length.

\textbf{Reasoning Modes.}
For the subset of models supporting configurable reasoning, we evaluate under two inference settings:
(1)~a \textit{high reasoning} mode that activates the model's highest supported Chain-of-Thought setting; and
(2)~a \textit{non-reasoning} mode that uses the minimal reasoning configuration.

\textbf{Evaluation Metrics.}
Since ground-truth answers in {\polar} are deterministic and span well-defined formats~(option labels, digits, coordinate positions, strings and lists), we employ exact-match accuracy as our primary metric.
For the responses whose format deviates from the expected structure, we apply LLM-as-a-judge~(\ie Gemini-3.1-Pro) as a calibration fallback to ensure fair evaluation~\citep{zheng2023judging,gu2024surveyllmjudge}.
To contextualize model performance, we additionally report a human performance reference alongside a theoretical random baseline. 
All models and tasks share the same prompt templates, with full setup details provided in Appendix.

\begin{table*}[t]
  \centering
  \caption{Main results on {\polar}. All models are evaluated under \textit{high} reasoning mode. C and P denote Cartesian and Polar accuracy(\%). Accuracy is averaged over tasks in each category. ($-\Delta$) indicates the C$\to$P accuracy drop. The \textbf{highest} and \underline{second-highest} scores are highlighted.}
  \label{tab:overall_results}
  \setlength{\tabcolsep}{3pt}
  \resizebox{\textwidth}{!}{
    \begin{tabular}{@{} l cc cc cc cc cc cc @{}}
      \toprule
      \multirow{2}{*}{\textbf{Model}} & \multicolumn{2}{c}{\textbf{\makecell{Algorithmic Logic \\ \& Simulation}}} & \multicolumn{2}{c}{\textbf{\makecell{Combinatorics \\ \& Probability}}} & \multicolumn{2}{c}{\textbf{\makecell{Navigation \\ \& Routing}}} & \multicolumn{2}{c}{\textbf{\makecell{Spatial Trans. \\ \& Geometry}}} & \multicolumn{2}{c}{\textbf{\makecell{Visual Pattern \\ Matching}}} & \multicolumn{2}{c}{\textbf{Overall}} \\
      \cmidrule(lr){2-3} \cmidrule(lr){4-5} \cmidrule(lr){6-7} \cmidrule(lr){8-9} \cmidrule(lr){10-11} \cmidrule(lr){12-13}
      & \textbf{C} & \textbf{P\dt{$-\Delta$}} & \textbf{C} & \textbf{P\dt{$-\Delta$}} & \textbf{C} & \textbf{P\dt{$-\Delta$}} & \textbf{C} & \textbf{P\dt{$-\Delta$}} & \textbf{C} & \textbf{P\dt{$-\Delta$}} & \textbf{C} & \textbf{P\dt{$-\Delta$}} \\
      \midrule
      Human  & 92.5 & 87.5 \dt{$-5.0$} & 77.8 & 64.4 \dt{$-13.4$} & 97.5 & 92.5 \dt{$-5.0$} & 98.8 & 94.6 \dt{$-4.2$} & 100.0 & 97.3 \dt{$-2.7$} & 94.5 & 88.8 \dt{$-5.7$} \\
      Random & 17.0 & 17.0 & 6.7 & 6.7 & 18.4 & 18.4 & 16.2 & 16.2 & 19.0 & 19.0 & 15.8 & 15.8 \\
      \midrule
      \multicolumn{13}{c}{\textit{Close Source MLLMs}} \\
      \midrule
      Gemini-3.1-Pro        & \textbf{85.0} & \textbf{52.6} \dt{$-32.4$} & \textbf{92.4} & \underline{24.9} \dt{$-67.6$} & \textbf{83.4} & 39.4 \dt{$-44.0$} & \textbf{83.7} & \underline{31.4} \dt{$-52.3$} & \textbf{71.0} & \textbf{36.6} \dt{$-34.4$} & \textbf{82.6} & \underline{35.9} \dt{$-46.7$} \\ 
      Gemini-3-Flash        & 68.8 & 41.5 \dt{$-27.3$} & 74.8 & 20.6 \dt{$-54.2$} & 76.2 & \underline{41.9} \dt{$-34.4$} & 68.7 & 29.8 \dt{$-38.8$} & 65.0 & \underline{34.8} \dt{$-30.2$} & 71.0 & 33.8 \dt{$-37.2$} \\ 
      Gemini-3-Flash-lite   & 58.0 & 28.3 \dt{$-29.7$} & 40.4 & 15.9 \dt{$-24.6$} & 54.6 & 28.8 \dt{$-25.8$} & 40.6 & 21.7 \dt{$-18.9$} & 43.3 & 27.7 \dt{$-15.5$} & 46.8 & 24.6 \dt{$-22.2$} \\ 
      Gemini-2.5-Pro        & 54.3 & 30.2 \dt{$-24.2$} & 23.3 & 12.7 \dt{$-10.7$} & 45.6 & 31.1 \dt{$-14.5$} & 33.8 & 23.9 \dt{$-9.8$}  & 38.4 & 27.0 \dt{$-11.4$} & 38.4 & 25.3 \dt{$-13.2$} \\ 
      Gemini-2.5-Flash      & 51.3 & 27.7 \dt{$-23.7$} & 15.9 & 9.4 \dt{$-6.4$}   & 38.7 & 26.8 \dt{$-11.9$} & 26.4 & 18.5 \dt{$-7.9$}  & 33.8 & 22.7 \dt{$-11.1$} & 32.2 & 21.1 \dt{$-11.2$} \\ 
      Gpt-5.2              & \underline{70.7} & \underline{46.7} \dt{$-24.0$} & \underline{84.8} & \textbf{30.7} \dt{$-54.1$} & \underline{82.2} & \textbf{42.8} \dt{$-39.4$} & \underline{76.5} & \textbf{41.6} \dt{$-34.9$} & \underline{70.0} & 34.5 \dt{$-35.5$} & \underline{77.4} & \textbf{39.2} \dt{$-38.2$} \\ 
      Claude-Sonnet-4.6     & 60.0 & 30.3 \dt{$-29.7$} & 42.7 & 16.8 \dt{$-25.9$} & 45.0 & 28.5 \dt{$-16.4$} & 42.3 & 25.2 \dt{$-17.0$} & 38.9 & 28.3 \dt{$-10.6$} & 44.4 & 25.9 \dt{$-18.5$} \\ 
      Grok-4-0709          & 53.2 & 32.8 \dt{$-20.4$} & 20.8 & 12.0 \dt{$-8.8$}  & 38.0 & 24.9 \dt{$-13.1$} & 29.9 & 22.2 \dt{$-7.7$}  & 29.5 & 19.6 \dt{$-9.9$}  & 33.0 & 21.8 \dt{$-11.2$} \\ 
      Grok-4-Fast-Reasoning  & 49.7 & 30.8 \dt{$-18.8$} & 17.7 & 12.3 \dt{$-5.3$}  & 37.7 & 27.1 \dt{$-10.6$} & 28.4 & 21.5 \dt{$-6.8$}  & 26.1 & 20.5 \dt{$-5.5$}  & 31.0 & 22.3 \dt{$-8.7$} \\ 
      \midrule 
      \multicolumn{13}{c}{\textit{Open Source MLLMs}} \\
      \midrule
      Qwen3.5-397B-A17B     & \underline{72.2} & \textbf{43.2} \dt{$-29.0$} & \underline{65.2} & \underline{18.1} \dt{$-47.1$} & \textbf{77.6} & \textbf{40.9} \dt{$-36.7$} & \textbf{77.2} & \textbf{39.5} \dt{$-37.8$} & \textbf{68.5} & \underline{31.5} \dt{$-37.0$} & \textbf{72.9} & \textbf{35.0} \dt{$-37.9$} \\ 
      Kimi-k2.5            & \textbf{73.5} & \underline{37.3} \dt{$-36.1$} & \textbf{68.7} & 16.2 \dt{$-52.5$} & \underline{73.0} & \underline{38.3} \dt{$-34.7$} & \underline{71.7} & \underline{31.1} \dt{$-40.6$} & \underline{58.4} & 30.8 \dt{$-27.5$} & \underline{69.0} & \underline{31.1} \dt{$-37.8$} \\ 
      Gemma-4-31B          & 67.7 & 36.0 \dt{$-31.7$} & 62.3 & \textbf{19.4} \dt{$-42.8$} & 63.9 & 37.6 \dt{$-26.2$} & 60.1 & 28.5 \dt{$-31.6$} & 51.5 & \textbf{32.5} \dt{$-19.1$} & 60.5 & 31.0 \dt{$-29.5$} \\ 
      Gemma-4-26B          & 56.0 & 29.5 \dt{$-26.5$} & 45.8 & 14.8 \dt{$-30.9$} & 48.9 & 26.5 \dt{$-22.4$} & 48.5 & 19.4 \dt{$-29.2$} & 40.0 & 25.4 \dt{$-14.6$} & 47.2 & 22.9 \dt{$-24.4$} \\ 
      Mistral-Small-2503    & 25.0 & 20.7 \dt{$-4.4$}  & 9.6  & 10.9 \dt{$1.3$}   & 22.6 & 21.2 \dt{$-1.4$}  & 19.5 & 21.2 \dt{$1.7$}   & 20.3 & 19.4 \dt{$-0.9$}  & 19.4 & 19.0 \dt{$-0.4$} \\ 
      \bottomrule
    \end{tabular}
  }
\end{table*} 

\subsection{Main results}
\label{subsec:main_results}

\begin{figure}[t]
    \centering
    \includegraphics[width=0.95\linewidth]{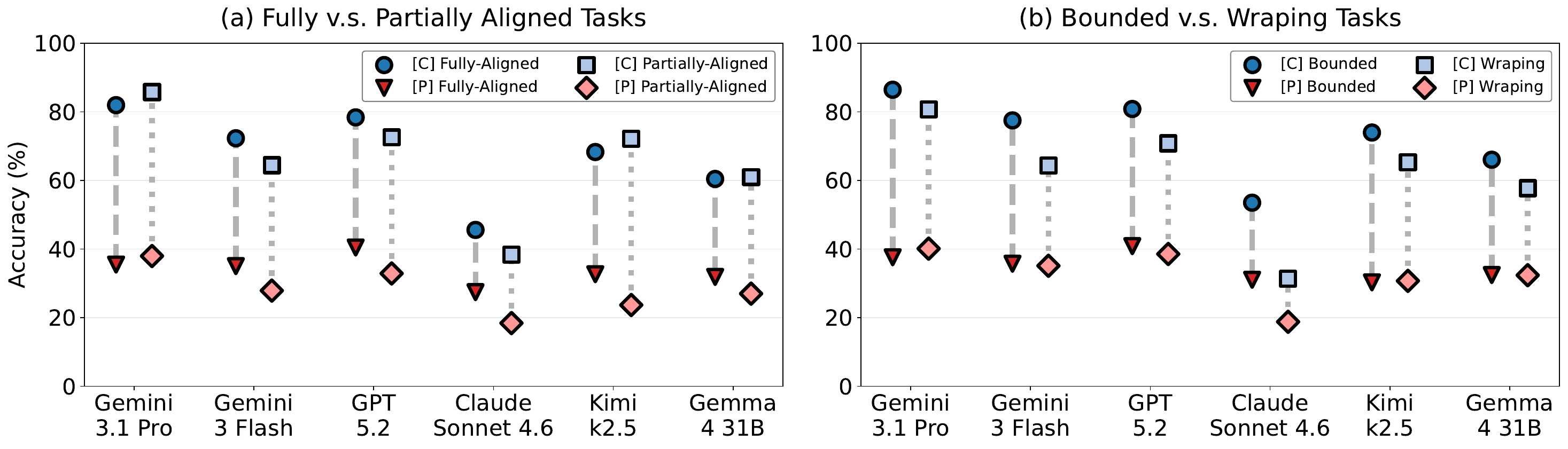}
    \caption{Cartesian (C) and Polar (P) accuracy under two partitions of the 53 tasks: \textit{(a)} fully vs.\ partially aligned; \textit{(b)} bounded vs.\ wrapping. The C$\rightarrow$P drop persists in every subset; wrapping mainly lowers Cartesian accuracy.}
    \label{fig:alignment_wrap_comparison}
\end{figure}

\begin{figure}[t]
    \centering
    \includegraphics[width=0.9\linewidth]{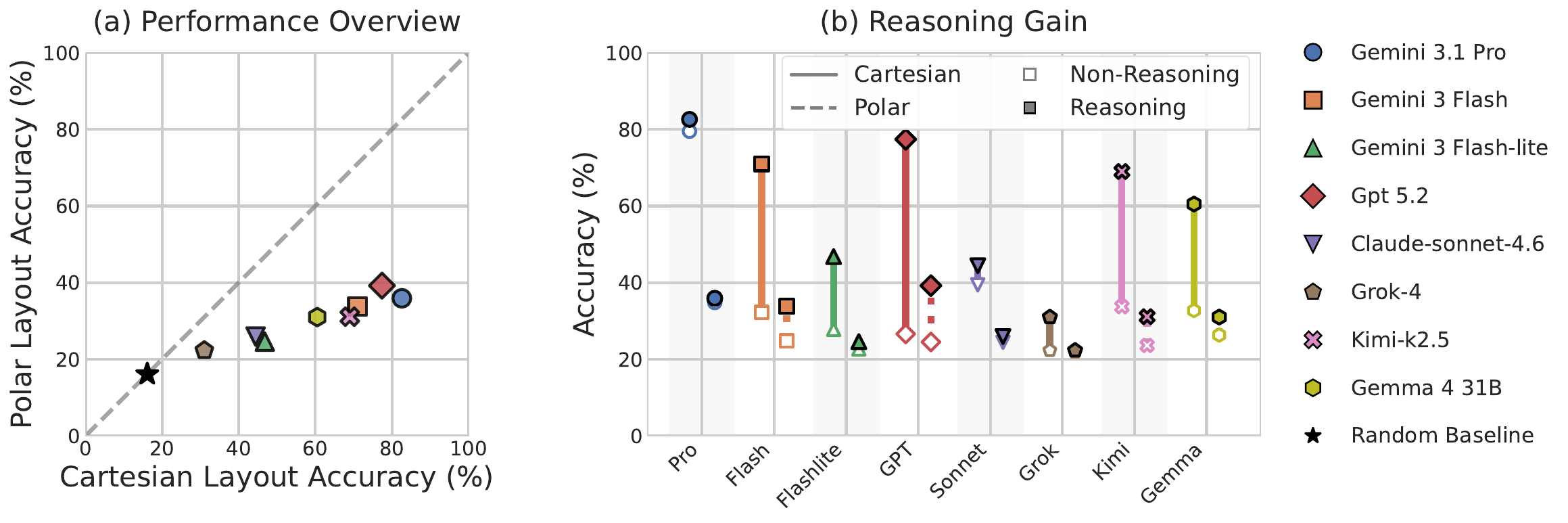}
    \caption{Performance overview and reasoning mode effect. \textit{(a)}~Overall Cartesian vs.\ Polar accuracy for each model~(reasoning: high). \textit{(b)}~Reasoning effect on Cartesian and Polar accuracy\nop{\howardzhou{the Cartesian/Polar legend in the diagram is not correct}}. Connected pairs show the reasoning gains on Cartesian layouts consistently diminish on Polar equivalents.}
    \label{fig:overall_n_thinkinggain}
\end{figure}

The main results in Table~\ref{tab:overall_results} report overall and per-category accuracy for all evaluated models under high reasoning mode, comparing Cartesian (C) and Polar (P) layouts. 
Figure~\ref{fig:alignment_wrap_comparison} examines topology alignment conditions.
Table~\ref{tab:reasoning_performance} reports reasoning vs.\ non-reasoning inference for configurable models. Figure~\ref{fig:overall_n_thinkinggain} visualize the comparison in Table~\ref{tab:overall_results},\ref{tab:reasoning_performance}. 
Figure~\ref{fig:discussion}(a) provides a representative failure case.

\textbf{Models consistently collapse on Polar layouts.}
Table~\ref{tab:overall_results} and Figure~\ref{fig:overall_n_thinkinggain}(a) present the Cartesian and Polar accuracy across all evaluated models. 
Frontier models achieving $69$--$83\%$ on Cartesian layouts collapse to $31$--$39\%$ on Polar equivalents, with the largest \nop{overall} drop over $45$ points, whereas human accuracy decreases by only 5.7 points from 94.5\% to 88.8\%.
Notably, while Cartesian accuracy varies widely across model families (\eg $82.6\%$ for Gemini-3.1-Pro, $33.0\%$ for Grok-4-0709), 
Polar scores compress into a narrow band, suggesting a shared performance floor that transcends model-specific capacity, consistent with a common dependence on orthogonal grid structure.
Per-category analysis reveals that this collapse is non-uniform: 
\textit{Combinatorics \& Probability} suffers the steepest degradation (\eg $92.4\to24.9$ for Gemini-3.1-Pro, $68.7 \to 16.2$ for Kimi-k2.5), which we hypothesize relates to the tasks' deeper dependency on structured grid enumeration. 
\textit{Algorithmic Logic \& Simulation} retains comparatively higher Polar accuracy (\eg $52.6\%$ for Gemini-3.1-Pro).
This performance collapse stay consistency across diverse model families, scales, and training paradigms, 
suggests that the performance collapse reflects a shared dependence on orthogonal structure rather than a model-specific limitation.


\textbf{Performance collapse persists across fully and partial alignment conditions.}
Figure~\ref{fig:alignment_wrap_comparison} examines whether the collapse is attributable to the coordinate transformation itself or to confounds introduced by altered solution spaces or boundary conditions (Section~\ref{sec:benchmark}).
On \textit{fully aligned} tasks~(Figure~\ref{fig:alignment_wrap_comparison}(a)), where the coordinate transformation is the sole variable, all models still exhibit severe collapse~(\eg Gemini-3.1-Pro: $82.0 \to 35.5$, Kimi-k2.5: $68.3 \to 32.6$) comparable in magnitude to the overall and partially aligned results.
This shows that the performance collapse persists even when the transformation preserves complete logical equivalence, indicating that altered solution spaces are not a primary confound.
\textit{Partially aligned} tasks show marginally larger gaps (\eg Gemini-3.1-Pro: $85.8 \to 38.0$), yet the collapse trend remains consistent.
The \textit{bounded} vs.\ \textit{wrapping} comparison (Figure~\ref{fig:alignment_wrap_comparison}(b)) similarly 
shows consistent C to P drops under both conditions; wrapping mainly lowers Cartesian accuracy.
Together, these results show that the performance collapse is robust across topological configurations, with boundary conditions (bounded vs. wrapping) exerting limited additional influence.

\begin{table*}[t]
\centering
\caption{Model performance across categories, comparing reasoning modes. ($-\Delta$) are shown alongside the Polar column scores. The third row for each model represents the thinking gain.}
\label{tab:reasoning_performance}
\setlength{\tabcolsep}{3pt} 
\resizebox{0.95\textwidth}{!}{
    \begin{tabular}{@{} l c c r@{\ }l c r@{\ }l c r@{\ }l c r@{\ }l c r@{\ }l c r@{\ }l @{}}
    \toprule
    \multirow{2}{*}{\textbf{Model}} & \multirow{2}{*}{\textbf{Reason}} & \multicolumn{3}{c}{\makecell{\textbf{Algorithmic Logic} \\ \textbf{\& Simulation}}} & \multicolumn{3}{c}{\makecell{\textbf{Combinatorics} \\ \textbf{\& Probability}}} & \multicolumn{3}{c}{\makecell{\textbf{Navigation} \\ \textbf{\& Routing}}} & \multicolumn{3}{c}{\makecell{\textbf{Spatial Trans.} \\ \textbf{\& Geometry}}} & \multicolumn{3}{c}{\makecell{\textbf{Visual Pattern} \\ \textbf{Matching}}} & \multicolumn{3}{c}{\textbf{Overall}} \\
    \cmidrule(lr){3-5} \cmidrule(lr){6-8} \cmidrule(lr){9-11} \cmidrule(lr){12-14} \cmidrule(lr){15-17} \cmidrule(lr){18-20}
    & & \textbf{C} & \multicolumn{2}{c}{\textbf{P}\dt{$-\Delta$}} & \textbf{C} & \multicolumn{2}{c}{\textbf{P}\dt{$-\Delta$}} & \textbf{C} & \multicolumn{2}{c}{\textbf{P}\dt{$-\Delta$}} & \textbf{C} & \multicolumn{2}{c}{\textbf{P}\dt{$-\Delta$}} & \textbf{C} & \multicolumn{2}{c}{\textbf{P}\dt{$-\Delta$}} & \textbf{C} & \multicolumn{2}{c}{\textbf{P}\dt{$-\Delta$}} \\
    \midrule
    \multirow{3}{*}{Gemini-3.1-Pro} 
    & $\checkmark$ & 85.0 & 52.6 & \dt{$-32.4$} & 92.4 & 24.9 & \dt{$-67.6$} & 83.4 & 39.4 & \dt{$-44.0$} & 83.7 & 31.4 & \dt{$-52.3$} & 71.0 & 36.6 & \dt{$-34.4$} & 82.6 & 35.9 & \dt{$-46.7$} \\
    & $\times$ & 66.7 & 35.6 & \dt{$-31.1$} & 83.4 & 22.7 & \dt{$-60.7$} & 82.9 & 42.6 & \dt{$-40.3$} & 86.2 & 31.8 & \dt{$-54.4$} & 71.3 & 37.6 & \dt{$-33.7$} & 79.5 & 34.8 & \dt{$-44.8$} \\
    & & 18.3 & 16.9 & & 9.1 & 2.2 & & 0.5 & -3.2 & & -2.5 & -0.4 & & -0.3 & -1.0 & & 3.1 & 1.1 & \\
    \midrule
    \multirow{3}{*}{Gemini-3-Flash} 
    & $\checkmark$ & 68.8 & 41.5 & \dt{$-27.3$} & 74.8 & 20.6 & \dt{$-54.2$} & 76.2 & 41.9 & \dt{$-34.4$} & 68.7 & 29.8 & \dt{$-38.8$} & 65.0 & 34.8 & \dt{$-30.2$} & 71.0 & 33.8 & \dt{$-37.2$} \\
    & $\times$ & 34.2 & 19.3 & \dt{$-14.8$} & 14.2 & 11.7 & \dt{$-2.6$} & 42.9 & 37.1 & \dt{$-5.9$} & 28.5 & 24.1 & \dt{$-4.5$} & 36.7 & 24.0 & \dt{$-12.7$} & 32.2 & 24.8 & \dt{$-7.4$} \\
    & & 34.7 & 22.2 & & 60.6 & 8.9 & & 33.3 & 4.8 & & 40.2 & 5.8 & & 28.3 & 10.8 & & 38.7 & 9.0 & \\
    \midrule
    \multirow{3}{*}{Gemini-3-Flash-lite} 
    & $\checkmark$ & 58.0 & 28.3 & \dt{$-29.7$} & 40.4 & 15.9 & \dt{$-24.6$} & 54.6 & 28.8 & \dt{$-25.8$} & 40.6 & 21.7 & \dt{$-18.9$} & 43.3 & 27.7 & \dt{$-15.5$} & 46.8 & 24.6 & \dt{$-22.2$} \\
    & $\times$ & 26.5 & 19.2 & \dt{$-7.3$} & 9.9 & 9.3 & \dt{$-0.6$} & 36.7 & 33.0 & \dt{$-3.7$} & 23.7 & 19.3 & \dt{$-4.4$} & 36.7 & 26.5 & \dt{$-10.3$} & 27.8 & 22.7 & \dt{$-5.1$} \\
    & & 31.5 & 9.2 & & 30.6 & 6.6 & & 17.9 & -4.2 & & 16.9 & 2.4 & & 6.5 & 1.3 & & 19.0 & 1.9 & \\
    \midrule
    \multirow{3}{*}{GPT-5.2} 
    & $\checkmark$ & 70.7 & 46.7 & \dt{$-24.0$} & 84.8 & 30.7 & \dt{$-54.1$} & 82.2 & 42.8 & \dt{$-39.4$} & 76.5 & 41.6 & \dt{$-34.9$} & 70.0 & 34.5 & \dt{$-35.5$} & 77.4 & 39.2 & \dt{$-38.2$} \\
    & $\times$ & 32.9 & 28.0 & \dt{$-5.0$} & 11.2 & 10.7 & \dt{$-0.5$} & 29.3 & 28.6 & \dt{$-0.8$} & 24.1 & 25.2 & \dt{$1.1$} & 35.4 & 28.1 & \dt{$-7.3$} & 26.6 & 24.5 & \dt{$-2.1$} \\
    & & 37.8 & 18.7 & & 73.6 & 20.0 & & 52.8 & 14.2 & & 52.4 & 16.5 & & 34.6 & 6.4 & & 50.8 & 14.6 & \\
    \midrule
    \multirow{3}{*}{Claude-Sonnet-4.6} 
    & $\checkmark$ & 60.0 & 30.3 & \dt{$-29.7$} & 42.7 & 16.8 & \dt{$-25.9$} & 45.0 & 28.5 & \dt{$-16.4$} & 42.3 & 25.2 & \dt{$-17.0$} & 38.9 & 28.3 & \dt{$-10.6$} & 44.4 & 25.9 & \dt{$-18.5$} \\
    & $\times$ & 45.4 & 26.9 & \dt{$-18.5$} & 42.6 & 16.7 & \dt{$-25.9$} & 45.5 & 25.5 & \dt{$-20.0$} & 29.6 & 24.0 & \dt{$-5.6$} & 37.5 & 28.4 & \dt{$-9.1$} & 39.4 & 24.4 & \dt{$-15.0$} \\
    & & 14.6 & 3.4 & & 0.2 & 0.2 & & -0.5 & 3.0 & & 12.6 & 1.2 & & 1.5 & -0.0 & & 4.9 & 1.5 & \\
    \midrule
    \multirow{3}{*}{Grok-4}\nop{~\footnote{We compare Grok-4-Fast-Reasoning and Grok-4-Fast-Non-Reasoning.}}
    & $\checkmark$ & 49.7 & 30.8 & \dt{$-18.8$} & 17.7 & 12.3 & \dt{$-5.3$} & 37.7 & 27.1 & \dt{$-10.6$} & 28.4 & 21.5 & \dt{$-6.8$} & 26.1 & 20.5 & \dt{$-5.5$} & 31.0 & 22.3 & \dt{$-8.7$} \\
    & $\times$ & 18.2 & 17.8 & \dt{$-0.4$} & 10.5 & 10.7 & \dt{$0.2$} & 35.3 & 36.8 & \dt{$1.5$} & 19.1 & 18.9 & \dt{$-0.3$} & 21.5 & 18.1 & \dt{$-3.4$} & 22.3 & 21.9 & \dt{$-0.4$} \\
    & & 31.5 & 13.0 & & 7.1 & 1.6 & & 2.4 & -9.7 & & 9.2 & 2.7 & & 4.6 & 2.4 & & 8.6 & 0.3 & \\
    \midrule
    \multirow{3}{*}{Kimi-k2.5} 
    & $\checkmark$ & 73.5 & 37.3 & \dt{$-36.1$} & 68.7 & 16.2 & \dt{$-52.5$} & 73.0 & 38.3 & \dt{$-34.7$} & 71.7 & 31.1 & \dt{$-40.6$} & 58.4 & 30.8 & \dt{$-27.5$} & 69.0 & 31.1 & \dt{$-37.8$} \\
    & $\times$ & 44.5 & 25.4 & \dt{$-19.1$} & 20.9 & 8.0 & \dt{$-12.9$} & 37.2 & 29.9 & \dt{$-7.4$} & 27.8 & 23.2 & \dt{$-4.6$} & 40.9 & 27.7 & \dt{$-13.2$} & 33.7 & 23.6 & \dt{$-10.2$} \\
    & & 29.0 & 12.0 & & 47.8 & 8.2 & & 35.7 & 8.4 & & 43.9 & 8.0 & & 17.5 & 3.1 & & 35.2 & 7.6 & \\
    \midrule
    \multirow{3}{*}{Gemma-4-31B} 
    & $\checkmark$ & 67.7 & 36.0 & \dt{$-31.7$} & 62.3 & 19.4 & \dt{$-42.8$} & 63.9 & 37.6 & \dt{$-26.2$} & 60.1 & 28.5 & \dt{$-31.6$} & 51.5 & 32.5 & \dt{$-19.1$} & 60.5 & 31.0 & \dt{$-29.5$} \\
    & $\times$ & 32.5 & 21.8 & \dt{$-10.7$} & 19.3 & 11.3 & \dt{$-8.0$} & 44.8 & 38.1 & \dt{$-6.7$} & 24.2 & 22.2 & \dt{$-2.0$} & 38.2 & 30.9 & \dt{$-7.3$} & 32.7 & 26.3 & \dt{$-6.3$} \\
    & & 35.2 & 14.2 & & 42.9 & 8.1 & & 19.1 & -0.4 & & 35.8 & 6.2 & & 13.4 & 1.5 & & 27.9 & 4.7 & \\
    \bottomrule
    \end{tabular}%
    }
\end{table*}

\textbf{Thinking gains on Cartesian layouts are severely diminished on Polar equivalents.}
Table~\ref{tab:reasoning_performance} and Figure~\ref{fig:overall_n_thinkinggain}(b) compare model performance under high reasoning and non-reasoning modes.
On Cartesian layouts, enabling reasoning yields substantial accuracy improvements, with gains up to $50.8$ points (GPT-5.2) and several models exceeding $30$ points.
However, on Polar layouts these gains are systematically compressed, with most models improving by fewer than $10$ points: Gemini-3-Flash's gain shrinks from $38.7$ to $9.0$, Kimi-k2.5 from $35.2$ to $7.6$.
At the category level, this pattern mirrors the findings in Table~\ref{tab:overall_results}: 
\textit{Combinatorics \& Probability}, which benefits more from reasoning on Cartesian (\eg Gemini-3-Flash: $60.6$ points), sees these gains collapse more severely on Polar ($8.9$ points), while \textit{Algorithmic Logic \& Simulation} retains comparatively more of its reasoning benefit.
This asymmetry shows that thinking gains are strongly coupled to the presence of orthogonal grid structure: reasoning substantially amplifies performance on Cartesian layouts, but this amplification diminishes severely on Polar equivalents.

\section{Discussions}
\label{sec:discussion}

\begin{figure}[t]
    \centering
    \begin{minipage}{0.5\linewidth}
        \centering
        \includegraphics[width=\linewidth]{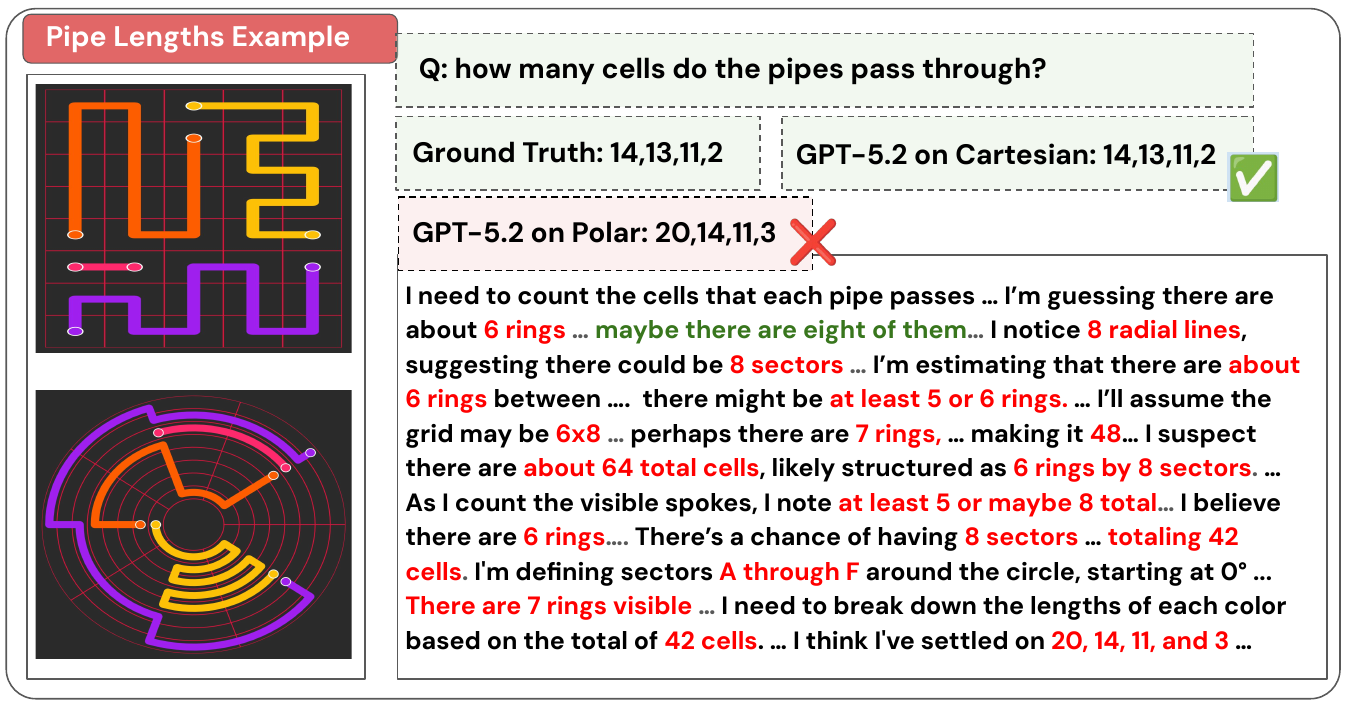}
        {(a)}
    \end{minipage}
    \begin{minipage}{0.24\linewidth}
        \centering
        \includegraphics[width=\linewidth]{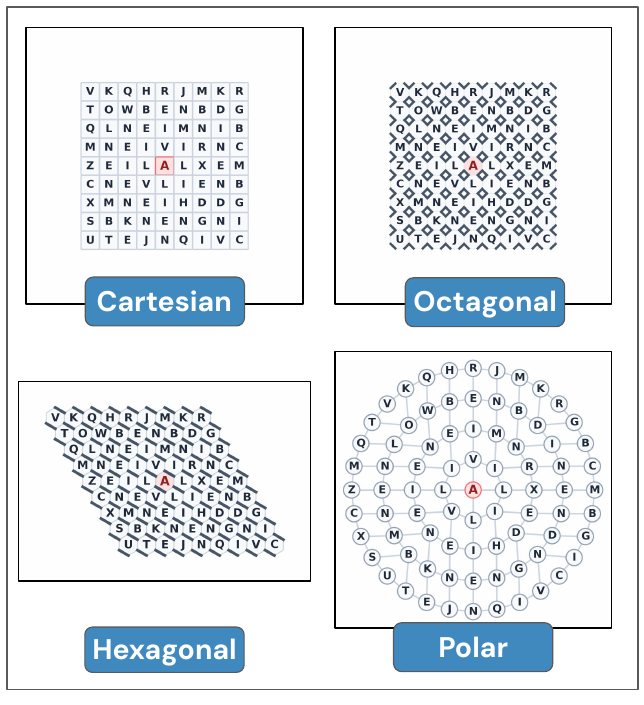}
        {(b)}
    \end{minipage}
    \begin{minipage}{0.24\linewidth}
        \centering
        \includegraphics[width=\linewidth]{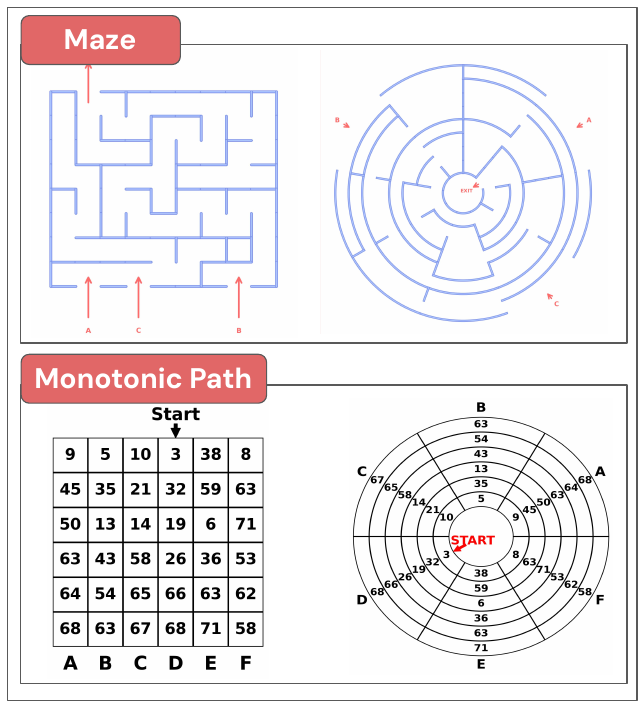}
        {(c)}
    \end{minipage}
    \begin{minipage}{0.33\linewidth}
        \centering
        \includegraphics[width=\linewidth]{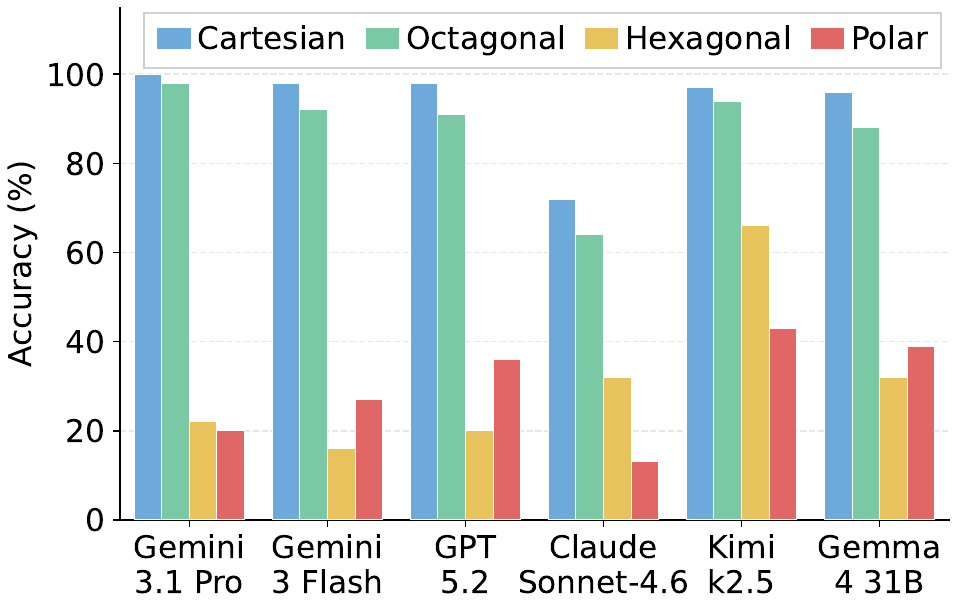}
        {(d)}
    \end{minipage}
    \begin{minipage}{0.325\linewidth}
        \centering
        \includegraphics[width=\linewidth]{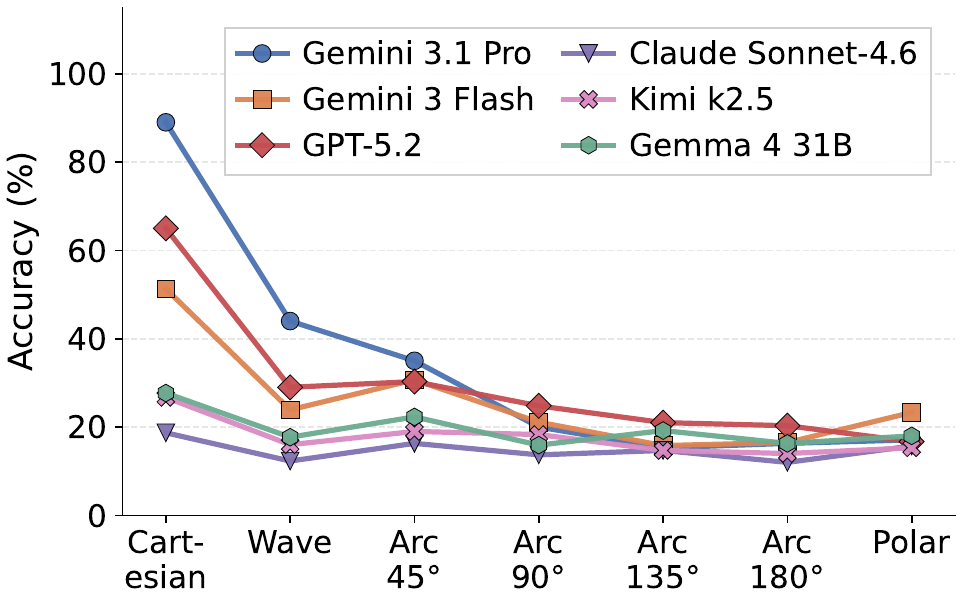}
        {(e)}
    \end{minipage}
    \begin{minipage}{0.325\linewidth}
        \centering
        \includegraphics[width=\linewidth]{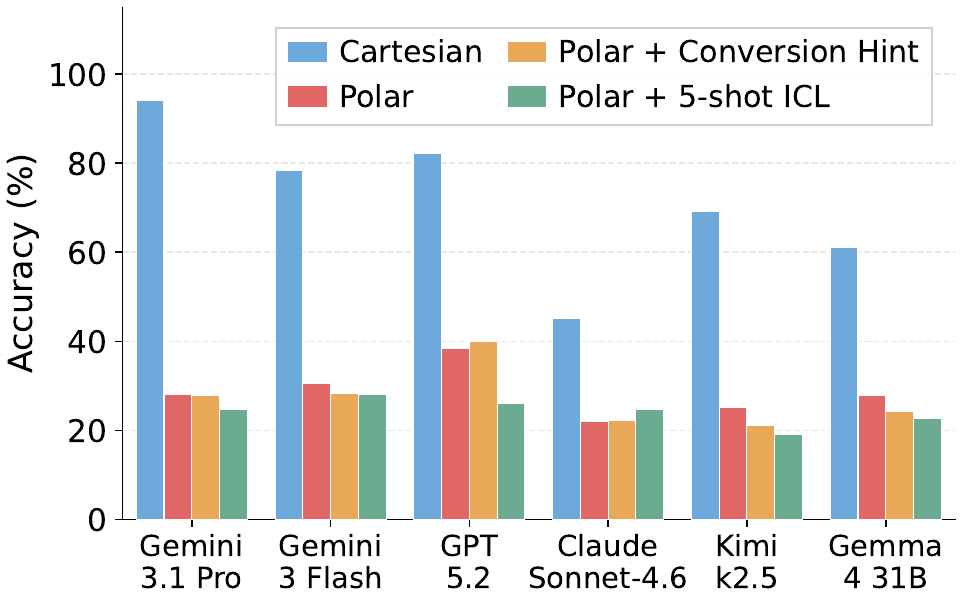}
        {(f)}
    \end{minipage}
    \caption{Case study and analyses for the discussion.
        \textit{(a)} Pipe Lengths case study: GPT-5.2 answers Cartesian correctly but repeatedly revises its ring and sector counts on Polar.
        \textit{(b)} Word Search in Cartesian, Octagonal, Hexagonal, and graph-based Polar layouts.
        \textit{(c)} Maze and Monotonic Path in both layouts.
        \textit{(d)} Word Search accuracy on the four layouts in (b).
        \textit{(e)} Accuracy under progressive deformation of Cartesian layouts, averaged over 3 tasks.
        \textit{(f)} Accuracy on Cartesian, Polar, and Polar with a conversion hint or 5-shot ICL, averaged over 6 tasks.
        (d-f) report 6 representative models.
        \nop{\howardzhou{We should order these 4 consistently both in the bar chart and in the task diagram, e.g. Cartesian, Octagonal, Hexagonal, Polar}}}
    \label{fig:discussion}
\end{figure}

We further characterize the Cartesian-Polar gap along three axes: whether it is specific to Polar layouts; whether it can be closed by test-time guidance; whether it appears uniformly across tasks.

\textbf{The gap is not specific to Polar layouts.} 
On Word Search, we compare Cartesian and graph-based Polar layouts with Octagonal and Hexagonal tilings (Figure~\ref{fig:discussion}(b)). 
Octagonal tilings, which preserve local orthogonal adjacency, stay close to Cartesian accuracy, whereas Hexagonal and Polar layouts show large drops for most models (Figure~\ref{fig:discussion}(d)). 
We further apply progressive geometric deformations to the Cartesian images of three tasks (Figure~\ref{fig:discussion}(e)).
For models with substantial Cartesian accuracy, a sine-wave deformation, which introduces neither wrap-around nor radial compression, already causes a large drop (e.g., Gemini 3.1 Pro: 89.0\% to 44.0\%), a 45 arc has a similar effect, and accuracy approaches the Polar level by the 90 arc. 
Models near floor on Cartesian for these tasks~(Sonnet 4.6, Kimi K2.5, Gemma 4 31B) vary little across conditions. Comparable drops therefore arise without Polar-specific properties, on several layouts that depart from the standard orthogonal grid.

\textbf{The gap is not closed by test-time guidance.} 
We evaluate two interventions on six tasks with large drops across six models (Figure~\ref{fig:discussion}(f)):
(1) a conversion hint instructing the model to re-map the Polar layout onto a Cartesian grid before solving, 
and (2) 5-shot in-context learning~\citep{dong2024surveyicl} with solved Polar examples. Conversion hints change accuracy by less than 4 points relative to the Polar baseline, and 5-shot examples lower accuracy for five of the six models (\eg GPT-5.2: 38.3\% to 26.0\%). Test-time instruction and demonstrations therefore do not close the gap between familiar orthogonal layouts and Polar layouts, which humans solve with a modest drop (Table~\ref{tab:overall_results}).

\textbf{Two tasks show no Cartesian-Polar gap.} 
In \textit{Monotonic Path}, leading models exceed 80\% on both Cartesian and Polar layouts; while in \textit{Maze}, most models score below 40\% on both (Figure~\ref{fig:discussion}(c)). 
Inspection of thinking traces on Monotonic Path reveals that models consistently reference the numerical labels in cells during reasoning~(e.g., ``currently at $35$, move to $45$''); Maze contains no such labels.
These cases show that the gap does not appear uniformly whenever the layout changes: models can reach high accuracy on Polar layouts and can score low on Cartesian ones. 
They are consistent with the use of textual representations observed in Section~\ref{sec:cartesian_shortcut}, although we present them as illustrative cases rather than a systematic analysis.

\section{Related works}

\textbf{Vision reasoning in {\mllm}.}
The rapid advancement of {\mllm}~\citep{gemini3pro2025,singh2025openai,team2026kimi,qwen3.5,anthropic2024claude} has driven a proliferation of visual reasoning benchmarks evaluating mathematical reasoning~\citep{lu2024mathvista, hao2025emma}, multi-discipline understanding~\citep{yue2025mmmu, yue2024mmmu}, spatial reasoning~\citep{jia2026omnispatial}, and broader, deeper vision reasoning~\citep{chen2025megabench, xu2025visulogic,meng2024mmiu, tong2024cambrian,zou2026unimmmu,wang2024charxiv,wang2024muirbench}, with several adopting structured grid-based layouts for precise assessment of spatial and algorithmic problem-solving~\citep{tang2025grasp, ren2025vgrp}.
Alongside this progress, growing evidence questions whether current models genuinely reason over visual content.
One line of work reveals that {\mllm} fail at perceptual tasks humans find trivial~\citep{fu2024blink, tong2024eyes,ye2026blinktwice,chen2026babyvision,saxena2025losttime}.
Another line questions whether strong benchmark scores genuinely reflect visual understanding~\citep{asadi2026mirage,chen2024we,xu2025more}: evaluation practices may overestimate true capability~\citep{chen2024we}, reasoning modes can amplify hallucination rather than improve visual grounding~\citep{xu2025more}, and frontier models generate elaborate reasoning for images never provided~\citep{asadi2026mirage}.
These concerns connect to shortcut learning~\citep{geirhos2020shortcut,xia2025visionaryr1}: models exploiting superficial regularities that generalize within benchmarks but collapse under distribution shift. 
However, the specific mechanisms by which models circumvent genuine visual reasoning remain underexplored. 
In this work, we identify the Cartesian Shortcut, a pervasive pattern in which models discretize orthogonal layouts into textual coordinates, and show that model performance collapses consistently when this structure is disrupted.

\textbf{Coordinate systems and geometric foundations.}
Standard neural architectures are built upon grid-structured representations with inherent translational symmetry~\citep{bronstein2021geometric}, an architectural prior that may contribute to model's Cartesian deduction behavior.
The equivalence and duality between different coordinate representations has long been studied in mathematics~\citep{arfken2011mathematical} and vision science~\citep{zetzsche1999atoms}.
To our knowledge, this is the first work to leverage coordinate transformation as a systematic diagnostic probe for {\mllm} visual reasoning.



\section{Conclusion}

In this work, we identify the Cartesian Shortcut, a pervasive pattern whereby MLLMs systematically discretize orthogonal grid layouts into textual coordinates, offloading reasoning from visual perception to text-based deduction. We introduce \polar, a controlled diagnostic benchmark of 53 paired Cartesian-Polar visual reasoning tasks. 
We choose the Cartesian Polar pair because the bijective mapping preserves task semantics while cleanly disrupting orthogonal structure, enabling a rigorous paired evaluation. 
Our experiments reveal that current MLLMs' visual reasoning performance is strongly coupled to orthogonal grid structure: when this structure is disrupted, performance collapses consistently across model families, is not recovered by prompting interventions, and is far smaller in humans. We do not claim that Polar conversion is either necessary or sufficient for performance collapse: several non-Polar deformations produce comparable drops (Section~\ref{sec:discussion}), and Polar tasks with textual anchors show no gap. Rather, this work establishes the scope and severity of models' dependence on orthogonal structure, and provides a controlled testbed for studying it. Separating the contributions of familiarity, geometry, and orthogonal structure through training-level interventions, extending to 3D and naturally occurring non-orthogonal visuals, and developing visual reasoning that generalizes across layout geometries all remain important open directions.

\section*{Acknowledgments}
The authors would like to express their deep appreciation to Tom Duerig and Huizhong Chen for their valuable suggestions, insightful comments, and constructive advice, which greatly benefited this work. 
We also extend our gratitude to the Kaggle Benchmarks team for their exceptional support
throughout this project, including providing computational resources for experiments. In particular,
we thank Nicholas Kang and Goeff Thomas for their outstanding and sustained partnership,
as well as Mohamed Amin, Andrew Wang, and Vincent Roseberry for their valuable support. Their
dedicated assistance with platform infrastructure, experiment execution, and dataset deployment, on
Kaggle was instrumental to the success of this work.

\bibliography{reference}

\newpage
\appendix

\clearpage
\section{Benchmark Details}


\subsection{Task Taxonomy}
\label{sec:task-taxonomy}

The {\polar} comprises 53 procedurally generated visual reasoning tasks, each yielding 200 QA pairs (100 Cartesian + 100 Polar), totaling 10,800 QA pairs across the suite (Word Search additionally includes Hexagonal and Octagonal variants, contributing 400 QA pairs). We organize these tasks into five cognitive domains based on the core reasoning ability each task primarily evaluates:

\begin{itemize}[leftmargin=*,itemsep=2pt]
    \item \textbf{Visual Pattern Matching} (11 tasks) evaluates pattern recognition and structural discrimination, including shape assembly, visual search, and sequential pattern continuation.
    \item \textbf{Spatial Transformation and Geometry} (13 tasks) evaluates geometric transformation reasoning~(e.g., rotation, reflection, folding) alongside quantitative geometric measurement such as area counting and curve estimation.
    \item \textbf{Navigation and Routing} (14 tasks) evaluates sequential positional state tracking, requiring models to execute or trace multi-step movement paths under environmental constraints such as walls, wrapping boundaries, and monotonicity rules.
    \item \textbf{Combinatorics and Probability} (9 tasks) evaluates systematic combinatorial enumeration and probabilistic reasoning, including path counting under topological boundary conditions and stochastic trajectory analysis.
    \item \textbf{Algorithmic Logic and Simulation} (6 tasks) evaluates constraint satisfaction, discrete optimization, and deterministic forward simulation of physical or rule-based dynamics.
\end{itemize}

\begin{figure}[h]
    \centering
    \includegraphics[width=0.99\linewidth]{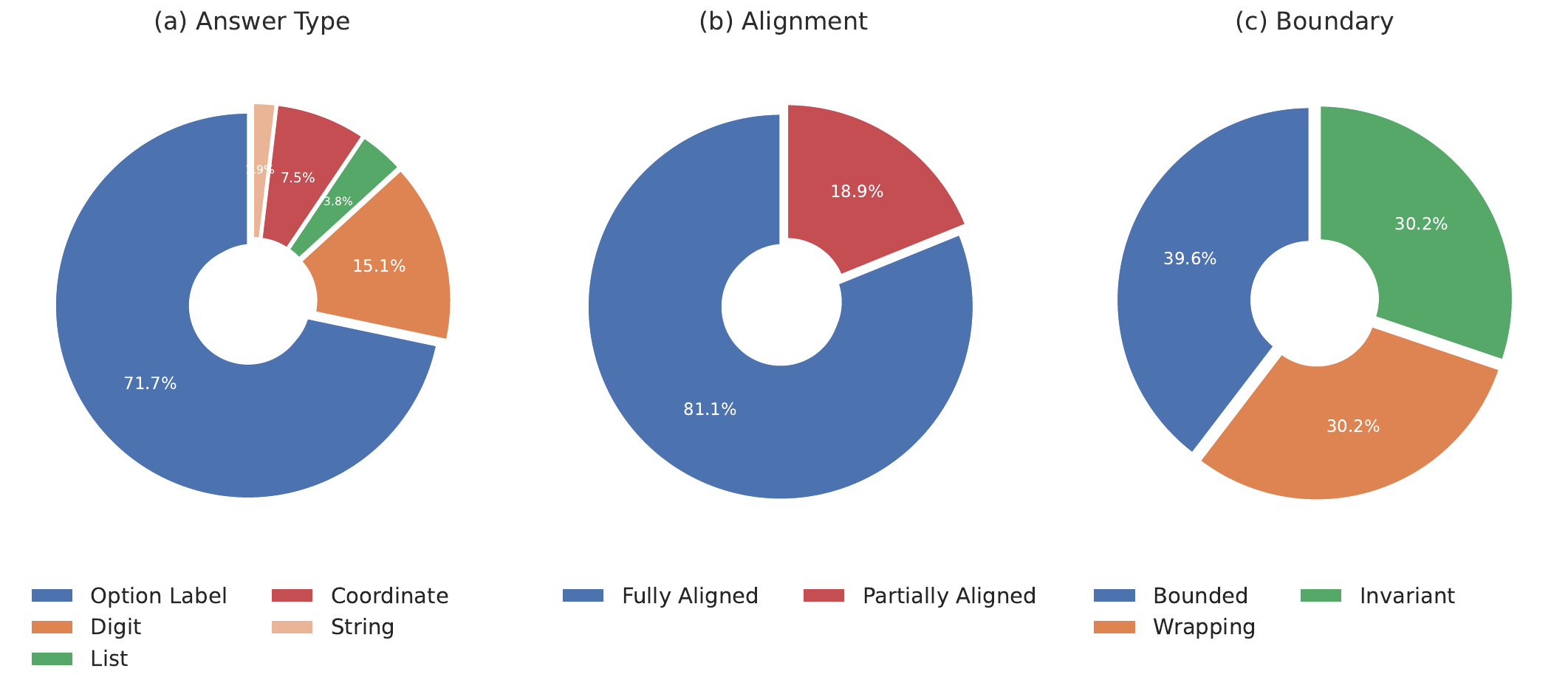}
    \caption{Task distribution of {\polar} across \textit{(a)} answer type, \textit{(b)} alignment, \textit{(c)} boundary condition.}
    \label{fig:pie_chart_taxonomy}
\end{figure}

Tables~\ref{tab:tasks-vpm}--\ref{tab:tasks-algo} provide a complete per-task breakdown. For each task, we report its sub-category, answer type, a one-sentence description, and two topological properties. \textit{Alignment} indicates whether the Cartesian and Polar versions share identical ground-truth answers (\cmark~= Fully Aligned) or intentionally diverge due to intrinsic coordinate differences (\xmark~= Partially Aligned). \textit{Boundary} specifies the Polar layout's boundary condition: \textit{Bounded} (closed, non-wrapping edges), \textit{Wrapping} (360\textdegree{} angular wrap-around), or ``---'' (tasks unaffected by the boundary condition). Figure~\ref{fig:pie_chart_taxonomy} provides a statistic of task distribution among answer type, alignment between Cartesian and Polar, and boundary condition.

\paragraph{Partially Aligned Tasks}
Below we provide task-specific explanation to partial aligned tasks, where the Cartesian and Polar instances share same problem rules, but the valid solution and ground-truth answers are different:
\begin{itemize}
    \item {Impossible Shape.} Finding a perfectly identical shape correspondence between Cartesian (rectangular cells) and Polar (sector-shaped cells) layouts is inherently infeasible. We therefore control for identical grid dimensions and overall shape silhouette, but due to the geometric differences between cell types, the set of valid rotation and assembly configurations may differ.
    \item {Knight Paths.} Under identical problem setups, the Polar layout permits angular wrapping, granting the knight additional boundary-crossing moves that do not exist on a bounded Cartesian grid. This results in more valid paths to reach the destination in Polar.
    \item {Longest Path.} The set of candidate paths and the correct answer are identical across both versions. However, to prevent candidate paths from inadvertently connecting head-to-tail through angular wrapping, the Polar layout uses one additional sector compared to the number of Cartesian columns.
    \item {Maximum Collection.} The Polar layout permits angular wrapping, expanding the reachable region along the trajectory. Consequently, more numbers can be collected in Polar under the same movement rules, yielding a higher optimal collection value.
    \item {Minimum Flips.} Angular wrapping makes the first and last columns effectively adjacent, introducing additional parity constraints on the boundary. This alters the minimum number of flips required to achieve the target pattern.
    \item {Random Walk.} The problem formulation and correct answer are kept identical across both versions. To achieve answer alignment despite wrapping, the Polar layout uses a slightly larger number of sectors, with the key intersections (A, B, C) placed within a contiguous sector region whose internal connectivity mirrors that of the Cartesian grid.
    \item {Rotation Center.} To ensure that the answer and key point positions align correctly between the two coordinate systems, the Polar grid uses a different grid size from the Cartesian version.
    \item {Wrapping Diagonal Path.} Angular wrapping in Polar allows diagonal paths to wrap around the angular boundary, creating additional valid trajectories that do not exist in the bounded Cartesian grid. This increases the total count of valid diagonal paths.
\end{itemize}

\begin{table}[ht]
\centering
\caption{Details of 11 tasks for the \textbf{Visual Pattern Matching} domain.
Each task yields 200 QA pairs (100 Cartesian + 100 Polar) unless otherwise noted.}
\label{tab:tasks-vpm}
\vspace{4pt}
\resizebox{\textwidth}{!}{%
\renewcommand{\arraystretch}{1.3}
\begin{tabular}{@{} p{3.2cm} p{2.7cm} p{2.0cm} p{7.5cm} c c c @{}}
\toprule
\textbf{Task Name} & \textbf{Sub-Category} & \textbf{Answer Type} & \textbf{Description} & \textbf{\#QA} & \textbf{Alignment} & \textbf{Boundary} \\
\midrule
Pattern Completion   & Shape and Jigsaw Assembly & Option label & Fill a missing contiguous region in a bi-color diagonally partitioned layout & 200 & \cmark & --- \\
Shape Fitting        & Shape and Jigsaw Assembly & Option label & Choose the piece that exactly fills an irregular hole in a cell layout & 200 & \cmark & Bounded \\
Layer Completion     & Shape and Jigsaw Assembly & Option label & Select the piece that fills a missing chunk in a concentric-layer colored grid & 200 & \cmark & Bounded \\
Fragment Matching    & Shape and Jigsaw Assembly & Option label & Match a grid fragment to a hole based on non-uniform line spacing patterns & 200 & \cmark & --- \\
Template Matching    & Shape and Jigsaw Assembly & Option label & Find which puzzle piece (with rotation/flip) is hidden in a symbol grid & 200 & \cmark & Wrapping \\
Jigsaw Matching      & Shape and Jigsaw Assembly & Option label & Determine which two of four pieces combine to form a complete shape & 200 & \cmark & Bounded \\
Shape Completion     & Shape and Jigsaw Assembly & Option label & Identify which piece fills the missing gap to complete a solid shape & 200 & \cmark & Wrapping \\
Odd Piece Out        & Visual Spotting and Search & Option label & Find the imposter piece that doesn't fit when assembling 4 pieces into a shape & 200 & \cmark & --- \\
Anomaly Detection    & Visual Spotting and Search & Coordinate & Find the one subtly different item in a large grid of identical items & 200 & \cmark & --- \\
Letter Collection    & Visual Spotting and Search & String & Walk along a spiral path and collect letters only on your right-hand side & 200 & \cmark & Wrapping \\
Pattern Prediction   & Pattern Continuation & Option label & Predict which point lies on a continuing periodic bump pattern & 200 & \cmark & Wrapping \\
\bottomrule
\end{tabular}%
}
\end{table}

\begin{table}[ht]
\centering
\caption{Task details for the \textbf{Spatial Transformation and Geometry} domain (13 tasks).
Notation follows Table~\ref{tab:tasks-vpm}.}
\label{tab:tasks-stg}
\vspace{4pt}
\resizebox{\textwidth}{!}{%
\renewcommand{\arraystretch}{1.3}
\begin{tabular}{@{} p{3.2cm} p{2.7cm} p{2.0cm} p{7.5cm} c c c @{}}
\toprule
\textbf{Task Name} & \textbf{Sub-Category} & \textbf{Answer Type} & \textbf{Description} & \textbf{\#QA} & \textbf{Alignment} & \textbf{Boundary} \\
\midrule
Rotation Matching    & Rotations and Reflections & List & Match 3 rotated versions of a colored ring/polygon pattern to 6 candidates & 200 & \cmark & Wrapping \\
Mirror Reflection    & Rotations and Reflections & Option label & Identify the correct mirror reflection of a half-filled colored layout & 200 & \cmark & Bounded \\
Grid Rotation        & Rotations and Reflections & Option label & Identify a grid's appearance after a 90\textdegree{} rotation from 5 options & 200 & \cmark & --- \\
Rotation Center      & Rotations and Reflections & Option label & Determine which point is the center of rotation between two segments & 200 & \xmark & --- \\
Pivot Rotation       & Rotations and Reflections & Option label & Identify the appearance of a shape after 180\textdegree{} rotation around a marked pivot & 200 & \cmark & --- \\
Impossible Shape     & Spatial Constraint and Topology & Option label & Find which combined grid cannot be made by inserting a piece into a frame & 200 & \xmark & Wrapping \\
Grid Folding         & Spatial Constraint and Topology & Option label & Fold a grid along a line; find which labeled cell is not covered & 200 & \cmark & Bounded \\
Four-Color           & Spatial Constraint and Topology & Option label & Insert a colored connected piece into a gap without violating adjacency rules & 200 & \cmark & --- \\
Area Counting        & Geometric Measurement and Counting & Digit & Count the number of gray shaded cells forming a connected region & 200 & \cmark & Bounded \\
Pipe Lengths         & Geometric Measurement and Counting & List & Count the lengths of colored pipes filling a layout; report in descending order & 200 & \cmark & --- \\
Uncut Cells          & Geometric Measurement and Counting & Option label & Count cells not intersected by two boundary-spanning diagonal cuts & 200 & \cmark & Bounded \\
Area Balancing       & Geometric Measurement and Counting & Option label & Choose the cell pattern that makes total black area equal total white area & 200 & \cmark & --- \\
Curve Length          & Geometric Measurement and Counting & Option label & Calculate the total length of a snake tracing cell boundary edges & 200 & \cmark & --- \\
\bottomrule
\end{tabular}%
}
\end{table}

\begin{table}[ht]
\centering
\caption{Task details for the \textbf{Navigation and Routing} domain (14 tasks).
Notation follows Table~\ref{tab:tasks-vpm}.}
\label{tab:tasks-nav}
\vspace{4pt}
\resizebox{\textwidth}{!}{%
\renewcommand{\arraystretch}{1.3}
\begin{tabular}{@{} p{3.2cm} p{2.7cm} p{2.0cm} p{7.5cm} c c c @{}}
\toprule
\textbf{Task Name} & \textbf{Sub-Category} & \textbf{Answer Type} & \textbf{Description} & \textbf{\#QA} & \textbf{Alignment} & \textbf{Boundary} \\
\midrule
Maze                  & Optimal Pathfinding & Option label & Find which labeled entrance leads to the exit through a maze & 200 & \cmark & Bounded \\
Shortest Path         & Optimal Pathfinding & Option label & Compare 4 diagonal zigzag routes and find the shortest (or all-equal) & 200 & \cmark & --- \\
Longest Path          & Optimal Pathfinding & Option label & Compare 5 paths on grids and identify which is the longest & 200 & \xmark & Bounded \\
Bounded Path Finding  & Optimal Pathfinding & Option label & Select the correct move sequence for a penguin reaching a fish on a bounded grid & 200 & \cmark & Bounded \\
Wrapping Path Finding & Optimal Pathfinding & Option label & Select the correct move sequence on a cylindrical wrapping grid & 200 & \cmark & Wrapping \\
Wrapping Navigation   & Rule-Based Navigation & Option label & Trace an absolute-direction movement sequence on a wrapping grid to find the target animal & 200 & \cmark & Wrapping \\
Egocentric Navigation & Rule-Based Navigation & Option label & Mouse traces egocentric commands (Fwd/Back/Turn) through cacti to find food & 200 & \cmark & Bounded \\
Absolute Navigation   & Rule-Based Navigation & Option label & Turtle traces absolute directions through cacti to find flowers (bounded) & 200 & \cmark & Bounded \\
Wall Follower         & Rule-Based Navigation & Coordinate & Simulate a robot that turns right at walls; find where it stops or loops & 200 & \cmark & Bounded \\
Rule-Based Navigation & Rule-Based Navigation & Coordinate & Simulate a drone with alternating R/L banking rules; find where it lands & 200 & \xmark & Bounded \\
Monotonic Path        & Constraint-Based Routing & Option label & Navigate rooms with strictly increasing numbers from START to the correct exit & 200 & \cmark & Wrapping \\
Turn Counting         & Constraint-Based Routing & Option label & Identify the path with the exact specified number of left and right turns & 200 & \cmark & Bounded \\
Word Search           & Constraint-Based Routing & Digit & Count valid adjacent-cell paths that spell a target word from the center & 400 & \cmark & --- \\
Largest Number Path   & Constraint-Based Routing & Option label & Trace digit paths through a number grid; identify which forms the largest number & 200 & \cmark & --- \\
\bottomrule
\end{tabular}%
}
\end{table}

\begin{table}[ht]
\centering
\caption{Task details for the \textbf{Combinatorics and Probability} domain (9 tasks).
Notation follows Table~\ref{tab:tasks-vpm}.}
\label{tab:tasks-comb}
\vspace{4pt}
\resizebox{\textwidth}{!}{%
\renewcommand{\arraystretch}{1.3}
\begin{tabular}{@{} p{3.2cm} p{2.7cm} p{2.0cm} p{7.5cm} c c c @{}}
\toprule
\textbf{Task Name} & \textbf{Sub-Category} & \textbf{Answer Type} & \textbf{Description} & \textbf{\#QA} & \textbf{Alignment} & \textbf{Boundary} \\
\midrule
Path Counting          & Bounded Combinatorics & Option label & Count distinct shortest paths from P to Q avoiding wall obstacles & 200 & \cmark & Bounded \\
Bounded Diagonal Paths & Bounded Combinatorics & Digit & Count diagonal paths from P to Q with a wall blocking Polar wrap-around & 200 & \cmark & Bounded \\
Bounded Knight Paths   & Bounded Combinatorics & Digit & Count knight paths from start to target with a wall blocking sector wrap-around & 200 & \cmark & Bounded \\
Checkpoint Paths       & Bounded Combinatorics & Digit & Count right/down paths from X to Y passing through checkpoint $\nabla$, avoiding gaps & 200 & \cmark & Bounded \\
Lattice Paths          & Bounded Combinatorics & Digit & Count paths on a dot grid allowing right, down, and diagonal-down-right moves & 200 & \cmark & Bounded \\
Wrapping Diagonal Paths & Topological Combinatorics & Digit & Count diagonal paths from P to Q on an alternating-color layout with wrapping & 200 & \xmark & Wrapping \\
Knight Paths            & Topological Combinatorics & Digit & Count paths for a chess knight from start to target in exactly K steps & 200 & \xmark & Wrapping \\
Edge Counting          & Stochastic Processes & Option label & Count the number of cell edges an ant traverses from P to Q & 200 & \cmark & --- \\
Random Walk            & Stochastic Processes & Option label & Calculate the probability of passing through point C on a random walk A$\to$B & 200 & \xmark & Wrapping \\
\bottomrule
\end{tabular}%
}
\end{table}

\begin{table}[ht]
\centering
\caption{Task details for the \textbf{Algorithmic Logic and Simulation} domain (6 tasks).
Notation follows Table~\ref{tab:tasks-vpm}.}
\label{tab:tasks-algo}
\vspace{4pt}
\resizebox{\textwidth}{!}{%
\renewcommand{\arraystretch}{1.3}
\begin{tabular}{@{} p{3.2cm} p{2.7cm} p{2.0cm} p{7.5cm} c c c @{}}
\toprule
\textbf{Task Name} & \textbf{Sub-Category} & \textbf{Answer Type} & \textbf{Description} & \textbf{\#QA} & \textbf{Alignment} & \textbf{Boundary} \\
\midrule
N-Queens            & Constraint Satisfaction & Option label & Place missing queens on a partial N-Queens board; deduce the target coordinate & 200 & \cmark & Bounded \\
Sudoku              & Constraint Satisfaction & Option label & Solve for the value of a highlighted cell in a partially filled Sudoku & 200 & \cmark & --- \\
Minimum Flips       & Move Optimization & Option label & Find minimum moves to achieve an alternating two-color pattern by flipping 3-cell strips & 200 & \xmark & Wrapping \\
Maximum Collection  & Move Optimization & Option label & Find the maximum cheese a mouse can eat on a non-repeating maze path & 200 & \xmark & Wrapping \\
Collision Detection & Motion and Trajectory Prediction & Option label & Determine which pair of moving cars will crash first based on velocity arrows & 200 & \cmark & Wrapping \\
Bouncing Point      & Motion and Trajectory Prediction & Coordinate & Compute coordinates after N steps of a bouncing/looping point & 200 & \xmark & Wrapping \\
\bottomrule
\end{tabular}%
}
\end{table}


\section{Benchmark Construction Details}
\subsection{Visual Calibration}
\label{app:visual_calibration}

The Polar coordinate mapping introduces nonlinear spatial distortion that, without mitigation, could confound the comparison by adding irrelevant legibility differences.
We apply task-specific visual calibrations to ensure that critical visual elements remain perceptually comparable across Cartesian and Polar layouts.

For example, for tasks where font size is a key perceptual factor~(\eg word search, odd one out), we enforce matched letter sizes across both layouts. Since Polar cells near the pole are significantly compressed, we adjust the Polar rendering to use cells large enough for legible text. To maintain size parity, we correspondingly introduce additional white space in the Cartesian layout, forcing the core visual elements (\ie the letters) to occupy a comparable rendered area rather than filling the entire grid. This ensures that any performance gap reflects reasoning difficulty rather than a legibility disadvantage.

For tasks involving geometric shapes or icons, we apply analogous area-matching calibrations: element sizes in Cartesian layouts are scaled to approximate the rendered area of their Polar counterparts, rather than naively filling the available cell space. These calibrations are tuned on a per-task basis through manual inspection.

\subsection{Procedural Generation}
\label{app:procedural_generation}

Each of the 53 tasks is implemented as a self-contained Python script that procedurally generates paired Cartesian--Polar instances. The generation process involves several carefully tuned components.

\paragraph{Randomization Tuning.}
We randomize multiple axes per task, including grid dimensions, starting positions, path topologies, target placements, distractor counts, and color palettes. The randomization bounds for each axis were manually calibrated through iterative testing to satisfy two constraints simultaneously: (1)~all generated instances must be solvable with a unique correct answer, and (2)~no configuration should produce visual elements that are too small or cluttered to be perceived, particularly under Polar distortion. This tuning process was performed by the authors through systematic manual review.

\paragraph{Solvability and Corner Case Resolution.}
Each script includes programmatic solvability checks that verify the generated instance has a valid, unique solution before it is accepted. Beyond these automated checks, we conducted extensive manual and LLM-assisted reviews to identify and resolve corner cases. For example, we discovered through literature review that N-Queens under wrapping (\ie toroidal) boundary conditions has no valid solution for most board sizes; we therefore constrained this task to use bounded topology exclusively. Similar task-specific adaptations were applied across the benchmark wherever edge cases were identified.

\paragraph{Distractor Validation.}
For tasks with multiple-choice or enumeration-based answers, we implement explicit distractor validation to ensure that incorrect answer options are genuinely incorrect. The generative scripts verify that each distractor does not accidentally coincide with the ground-truth answer under alternative interpretations or boundary conditions. This prevents ambiguous evaluation where a model might select a ``wrong'' answer that is in fact defensible.

\subsection{Rendering and Legibility}
\label{app:rendering}

Rendering Polar layouts introduces several unique visual challenges that require dedicated engineering solutions.

\paragraph{Anti-Collision Constraints.}
We enforce strict anti-collision rules to prevent overlapping visual elements. Grid lines, labels, icons, and path traces are rendered with minimum separation guarantees. Under Polar distortion, cells near the pole are significantly smaller than those near the boundary; we dynamically adjust label placement and font sizing to prevent overlap in these compressed regions.

\paragraph{Boundary Wrapping Artifacts.}
Certain visual patterns that appear straightforward in Cartesian layouts create rendering artifacts under Polar mapping. For instance, a polyline that spans both the left and right edges of a Cartesian grid will form a closed circle in Polar space, fundamentally altering its visual appearance. We address such cases through multiple strategies: constraining the generative script to produce polylines that touch at most one boundary, adding an extra column to the Polar layout to break the visual closure, or programmatically detecting and excluding instances where unwanted closed loops arise.

\paragraph{Inner Ring Radius Optimization.}
If the inner ring radius is too small, elements near the pole become illegibly compressed; if too large, the outer rings waste space. We tune the inner ring radius on a per-task basis to find the optimal balance between visual clarity near the pole and efficient use of the rendering canvas. This parameter is adjusted through manual inspection of representative instances for each task.

\subsection{Quality Assurance Pipeline}
\label{app:qa}

Our quality assurance process consists of three iterative stages followed by a final human validation study.

\paragraph{Author Review.}
The authors collaboratively reviewed the logical design of each task, verifying that the rules, constraints, and ground-truth computations are correct for both Cartesian and Polar versions. In addition, we independently spot-checked randomly sampled instances for each of the 53 tasks by manually solving them and comparing our answers against the computed ground truths.

\paragraph{LLM-Assisted Code Audit.}
We used Gemini-3.1-Pro to systematically audit the procedural generation code, prompting the model to identify potential edge cases, off-by-one errors, and boundary condition failures. This process surfaced several subtle bugs that were not apparent during manual review, including incorrect wrapping behavior at grid boundaries and degenerate configurations for specific grid dimensions.

\paragraph{Iterative Convergence.}
The above two stages were repeated in multiple rounds. Each round produced a list of identified issues, which were resolved before the next round of review. This process continued until no further issues were identified across all 53 tasks.

\paragraph{Representative Examples of Resolved Issues.}
To illustrate the depth of this process, we highlight several representative cases:
\begin{itemize}[leftmargin=20pt]
    \item \emph{N-Queens wrapping:} Literature review revealed that N-Queens under toroidal (wrapping) boundary conditions has no valid solution for most board sizes. We adapted this task to use bounded topology exclusively.
    \item \emph{Polyline boundary closure:} Polylines spanning both edges of a Cartesian grid form closed circles under Polar mapping. We constrained generation to avoid this artifact (Section~\ref{app:rendering}).
    \item \emph{Distractor collision:} In several combinatorial tasks, randomly generated distractors occasionally coincided with the correct answer under alternative boundary interpretations. We added explicit distractor validation checks to eliminate these cases.
\end{itemize}

\paragraph{Representative Design Refinements.}
Beyond resolving correctness issues, we iteratively refined task designs in this stage based on observations and results, for more comprehensive evaluation and reduce possible shortcut for answering. Below are some representative examples:
\begin{itemize}[leftmargin=20pt]
    \item \emph{Four Color:} This task originally used numerical labels (1--4) to fill cells, presenting a layout close to the Monotonic Path task. Pilot evaluation revealed that both tasks exhibited similarly high accuracy across Cartesian and Polar layouts. After comparative analysis on their CoT reasoning chain, we retained numerical labels for monotonic path and switched four color to use color-filled cells, creating a controlled contrast pair. Under this design, four color exhibits sharp performance collapse on Polar layouts, further corroborating our observation that Monotonic Path does not show this drop (Section~\ref{sec:discussion}).
    \item \emph{Letter Collection:} This task was originally framed as multiple-choice. We observed that models could exploit an elimination strategy: identifying letters unique to each option, finding them on the image and verifying their positioning correctness, rather than genuinely tracing the path to collect letters. We reformulated the task to require sequential string output, forcing models to actually traverse the path and aggregate the collected letters in order. This adjustment caused a sharp performance decline, confirming that the original format permitted a reasoning shortcut. Notably, under the multiple-choice setting, increasing the length of candidate strings does not degrade performance, indicating that the difficulty of the underlying task is not really be evaluated in the multi-choice setting. After converting to string output format, we subsequently reduced task difficulty in multiple iterations to ensure performance remained sufficiently above the random baseline to yield informative signal.
\end{itemize}

\subsection{Human Validation Study}
\label{app:human_validation}



As a final safeguard, we conducted a human validation study~(pool of over 50 human raters) to confirm the quality of the benchmark data and simultaneously establish a human performance reference (Appendix~\ref{app:human_baseline}). For each of the 53 tasks, we randomly sampled 20 Cartesian--Polar paired instances (2,120 individual questions in total) and presented them independently to general-audience human raters, with the Cartesian and Polar versions of each pair assigned separately. For each instance, raters were required to evaluate three criteria and provide a complete step-by-step derivation:

\begin{enumerate}
    \item \textbf{Visual clarity:} Is the rendered image visually clear and legible, such that all relevant elements (labels, shapes, paths, grid lines) can be unambiguously perceived?
    \item \textbf{Logical correctness:} Is the task logic (rules, constraints, problem formulation) consistent and well-defined?
    \item \textbf{Problem solving and solution trace:} Raters were required to independently solve the question and write out a detailed, logically complete step-by-step solution trace justifying their final answer (or select \emph{I don't know} if unable to reach a solution with confidence).
\end{enumerate}

To guarantee annotation quality, we employed a two-stage annotator and quality-checker protocol. An independent quality checker reviewed every submission to verify that the written reasoning chain was complete, self-contained, and consistent with the submitted answer; any submission with incomplete reasoning or unjustified claims or incomplete response was returned to the rater for revision until accepted. 
Finally, for all instances where a rater's answer differed from the ground truth (\emph{Incorrect}) or where the rater opted out (\emph{I don't know}), the authors manually solved and re-audited each case. This audit confirmed that all such cases stemmed from rater reasoning errors or opt-outs on challenging problems rather than flaws in the benchmark, verifying the visual clarity, logical soundness, and ground-truth accuracy across all sampled instances.

\section{Evaluation Setup Details}
\label{sec:suppl_eval_setup_detail}

\subsection{Model Query Details}

All models are evaluated via their publicly available APIs using default generation configurations to ensure optimal performance. We set the maximum output token limit to each model's maximum supported value. For the Gemini, GPT, Grok and Gemma model families, we query their respective first-party APIs. For all remaining models (\eg Qwen, Kimi, Mistral), we access them through third-party serving platforms, specifically TogetherAI and OpenRouter.

\paragraph{Evaluated Models.}
Here is the full list of models we evaluated. 
For proprietary models, we include the Gemini family (\textit{Gemini-3.1-Pro}, \textit{Gemini-3-Flash}, \textit{Gemini-3.1-Flash-Lite}, \textit{Gemini-2.5-Pro}, \textit{Gemini-2.5-Flash})~\citep{comanici2025gemini,gemini3pro2025,gemini3flash_techreport,gemini31flashlite_techreport}, \textit{GPT-5.2}~\citep{singh2025openai}, \textit{Claude-Sonnet-4.6}~\citep{anthropic2024claude}, and \textit{Grok-4} family (\textit{Grok-4-0709}, \textit{Grok-4-Fast-Reasoning}, \textit{Grok-4-Fast-Non-Reasoning})~\citep{grok4modelcard2025}.
For open-source models, we evaluate \textit{Gemma-4-31B}, \textit{Gemma-4-26B}~\citep{gemma4}, \textit{Kimi-k2.5}~\citep{team2026kimi}, \textit{Qwen-3.5-397B-A17B}~\citep{qwen3.5}, and \textit{Mistral-Small-3.1-24B}~\citep{mistral2025small31}.

\paragraph{Reasoning Mode Configuration.}
For models supporting configurable reasoning, we evaluate under two inference settings: (1) a \textit{high reasoning} mode that activates the model's highest supported reasoning budget, and (2) a \textit{non-reasoning} mode that uses the minimal reasoning configuration. We note the following model-specific exceptions:
\begin{itemize}
    \item \textbf{Gemini-3.1-Pro}~\citep{gemini3pro2025}: Gemini-3.1-Pro does not support a fully non-reasoning setting. We therefore report its non-reasoning results under the low reasoning budget~(reasoning set to LOW).
    \item \textbf{Claude-Sonnet-4.6}~\citep{anthropic2024claude}: Due to query timeouts under the extended high reasoning mode, the high reasoning results reported in this paper are obtained using the default adaptive high reasoning mode.
    \item \textbf{Grok-4-Fast-Reasoning}~\citep{grok4modelcard2025}: We pair Grok-4-Fast-Reasoning and Grok-4-Fast-Non-Reasoning as the high reasoning and non-reasoning variants, respectively, in Table~\ref{tab:reasoning_performance}.
\end{itemize}

\subsection{Evaluation Setup for the Cartesian Shortcut}
\label{app:cartesian_shortcut_setup}

Here we provide implementation details for the empirical analyses presented in Section~\ref{sec:cartesian_shortcut}.

\paragraph{Chain-of-Thought Discretization Analysis.}
To quantify how frequently models discretize visual layouts into textual coordinates during reasoning, we conduct a Chain-of-Thought (CoT) analysis across 9 prominent visual reasoning benchmarks: BabyVision~\citep{chen2026babyvision}, OmniSpatial~\citep{jia2026omnispatial}, EMMA~\citep{hao2025emma}, MegaBench~\citep{chen2025megabench}, MMMU-Pro~\citep{yue2025mmmu}, BLINK~\citep{fu2024blink}, CharXiv~\citep{wang2024charxiv}, MUIRBench~\citep{wang2024muirbench}, and MMIU~\citep{meng2024mmiu}. We use Gemini-3-Flash as a representative model. To isolate synthetic, grid-based visual reasoning tasks, we first prompt the model to filter out photographic images and examples with domain-specific backgrounds (\eg, medical imaging, charts), yielding over 3,800 qualifying examples. We then evaluate these examples under the high reasoning mode of Gemini-3-Flash and collect the intermediate Chain-of-Thought outputs. To systematically detect spatial discretization, we provide the generated reasoning traces to Gemini-3.1-Pro as a judge, prompting it to identify whether the reasoning explicitly invokes Cartesian coordinates or axis-aligned positional references (\eg, ``row 2'', ``column 3'', ``(x,y)''). Results show that $56.58\%$ of examples exhibit explicit Cartesian coordinate usage in their reasoning traces, confirming the prevalence of the Cartesian Shortcut.

\paragraph{Two-Stage Perception-Reasoning Disentanglement.}
To further isolate the role of visual perception from deductive reasoning, we conduct a two-stage experiment on Gemini-3.1-Pro (Figure~\ref{fig:cartesian_shortcut}). In the first stage, we provide only the task image and prompt the model to generate a detailed descriptive caption, explicitly informing the model that this caption will subsequently be used to answer a question. In the second stage, we provide the original question together with the generated caption (without the image) and prompt the model to answer based solely on the textual description. To evaluate caption quality, we use the grid dimensions as a proxy for perceptual accuracy: we extract the ground-truth grid size from each example's metadata and prompt Gemini-3.1-Pro to verify whether the caption correctly reports the grid dimensions. 
We clarify that comprehensively evaluating caption fidelity would require manual human inspection; grid size verification serves as a coarse but scalable proxy for perceptual accuracy rather than a precise measure. 
A caption is marked as incorrect if it explicitly mentions a grid size that contradicts the ground truth. As reported in Figure~\ref{fig:cartesian_shortcut}, grid size extraction accuracy reaches 80.2\% on Cartesian layouts but drops to only 20.8\% on Polar layouts. Crucially, when evaluating task accuracy exclusively on the subset of correctly perceived instances, Cartesian examples maintain 86.9\% accuracy while Polar examples collapse to 47.6\%, demonstrating that the performance gap persists even when visual perception is controlled for.

\subsection{Evaluation Prompt Template}
\label{app:prompt_template}

\begin{figure}[h]
    \centering
    \includegraphics[width=0.8\linewidth]{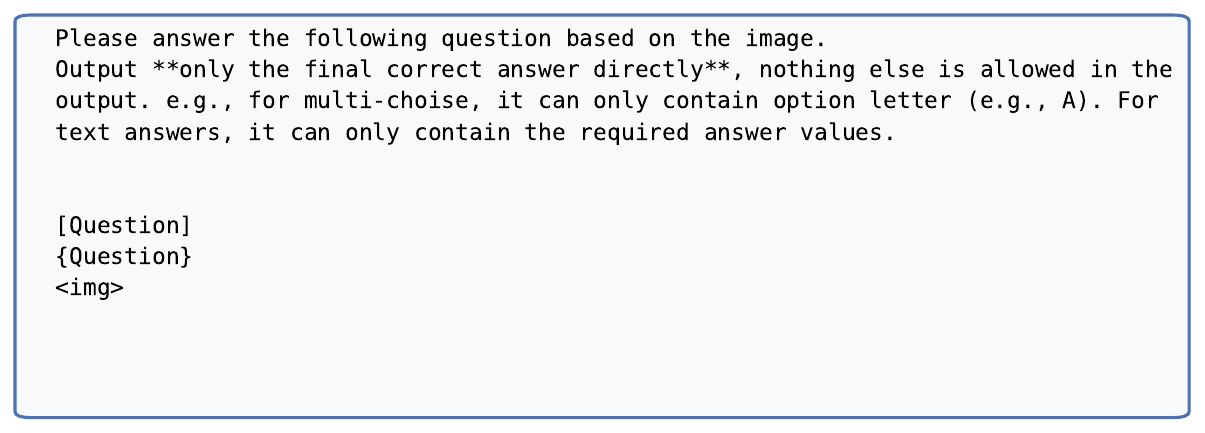}
    \caption{Prompt template used in the standard evaluation in our experiments.}
    \label{fig:prompt_standard}
\end{figure}
\begin{figure}[h]
    \centering
    \includegraphics[width=0.8\linewidth]{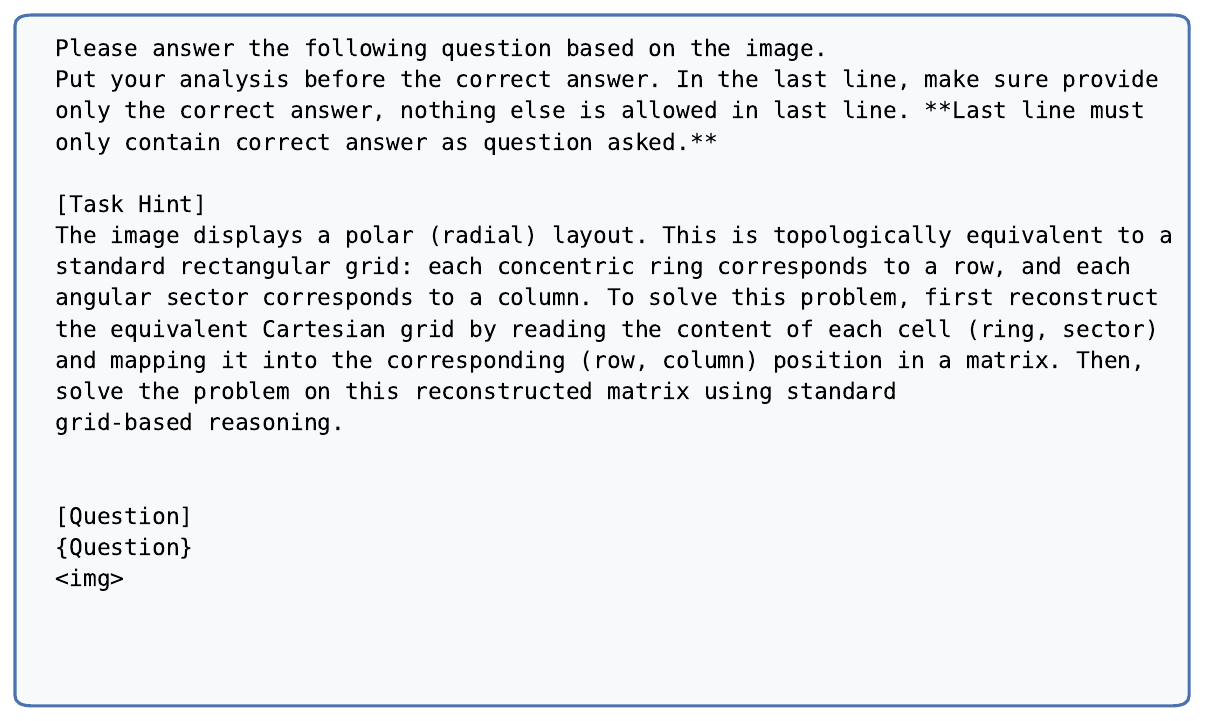}
    \caption{Prompt template used in conversion hint experiment.}
    \label{fig:prompt_hint}
\end{figure}
\begin{figure}[h]
    \centering
    \includegraphics[width=0.8\linewidth]{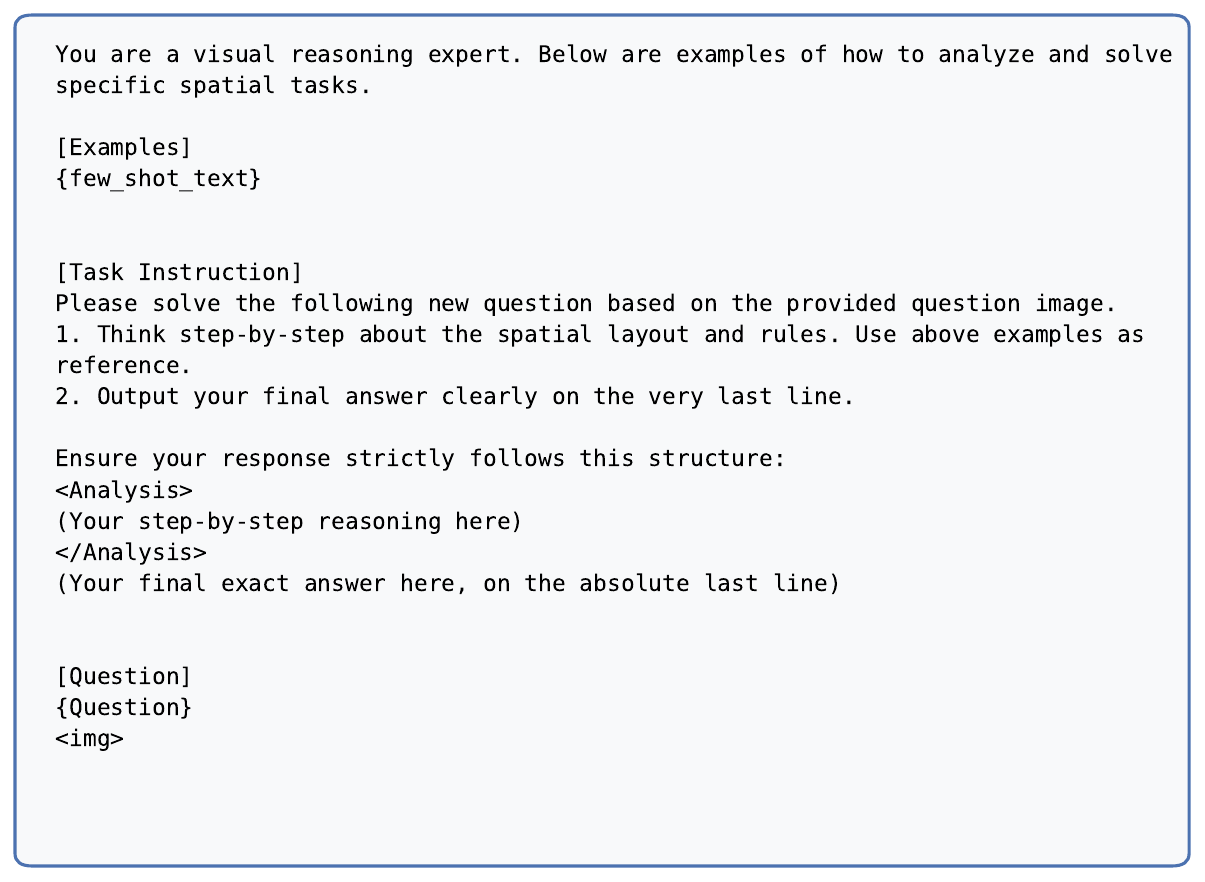}
    \caption{Prompt template used in the few-shot in context learning.}
    \label{fig:prompt_fewshot_cot}
\end{figure}

All tasks across both Cartesian and Polar layouts share a unified evaluation prompt template to ensure fair comparison. The template consists of: (1) a task-specific narrative providing contextual framing and rules, (2) the visual input image, and (3) the question with its answer format specification. The exact prompt template is provided in Figure~\ref{fig:prompt_standard}.

For the prompting interventions described in Section~\ref{sec:discussion}, we employ two additional prompt variants:
\begin{itemize}
    \item \textbf{Conversion Hint:} The prompt is augmented with an explicit instruction asking the model to first re-map the Polar layout into an equivalent Cartesian grid representation before attempting to solve the task.
    \item \textbf{5-Shot In-Context Learning:} The prompt is prepended with five solved Polar examples from the same task, each consisting of the Polar image, the question, and the correct answer, to scaffold the model's reasoning on Polar layouts.
\end{itemize}
\noindent The exact prompts for these two variants are provided in Figure~\ref{fig:prompt_hint} and Figure~\ref{fig:prompt_fewshot_cot} respectively..

\subsection{Evaluation Metrics}
\label{app:eval_metrics}

All tasks in the {\polar} are designed with deterministic, well-defined answer formats, enabling reliable automated evaluation. Depending on the task, ground-truth answers take one of the following standardized forms: multiple-choice labels, exact integers, coordinate pairs or axis positions, ordered lists, and directional sequences, as stated in Section~\ref{sec:task-taxonomy}. 
In this case, we employ rule-based programmatic parsers as the primary evaluation method. For each answer format, a dedicated parser extracts the model's predicted answer from its response and performs an exact-match comparison against the ground truth. Across all evaluated models, the vast majority of responses are successfully parsed and evaluated through this automated pipeline.
For responses where programmatic parsing fails, typically due to models not following the specified output format instructions, we apply an LLM-as-a-judge~\citep{zheng2023judging,gu2024surveyllmjudge} as a calibration fallback. Specifically, we prompt Gemini-3.1-Pro with both the ground-truth answer and the model's raw response, and ask it to determine whether the response is correct. This two-tier evaluation strategy ensures that models are not unfairly penalized for minor formatting deviations while maintaining rigorous evaluation standards.

\paragraph{Random Baseline.}
We additionally report a theoretical random baseline computed from the statistical expectation of each task's answer format. For multiple-choice tasks, this corresponds to $1/k$ where $k$ is the number of options; for integer and coordinate tasks, the baseline is derived from the uniform distribution over the valid answer space. This baseline serves as a lower-bound reference to contextualize model performance.

\paragraph{LLM-as-a-judge.}
Across all evaluated models, the overall fallback rate to the LLM judge (Gemini-3.1-Pro) is 14.47\% (e.g., Gemini-3.1-Pro: 5.0\%, Grok-4: 5.1\%, Qwen3.5-397B-A17B: 9.7\%, Gemma-4-26B: 15.3\%, GPT-5.2: 17.8\%, Claude-Sonnet-4.6: 29.6\%, and Kimi-k2.5: 47.3\%), primarily reflecting differences in how strictly each model adheres to the requested output format; notably, Gemini-3.1-Pro itself exhibits one of the lowest fallback rates, meaning the judge is rarely invoked on its own outputs. The responses routed to the fallback judge are predominantly surface-level format mismatches—for instance, ground truth ``7'' versus model output ``seven'', ``(18,2)'' versus ``18 and 2'', or ``C: 62cm'' versus ``62cm'', where the prediction is semantically correct but fails string-level exact matching. Because the judge only needs to determine whether the predicted answer semantically matches a deterministic ground truth, this verification task lies well within the reliable regime of frontier LLMs. To empirically validate the judge's reliability, we applied it to model predictions already scored by exact matching; across three independent runs on GPT-5.2's outputs those of which could be evaluated by exact match, the judge achieved 100\% agreement with exact-match scoring, 
and random human spot-checks on fallback outputs identified no grading errors. 
Combining this low self-fallback rate, the 100\% validation agreement, and the observation that Gemini-3.1-Pro exhibits the largest Cartesian-to-Polar accuracy drop among all evaluated models (Table~\ref{tab:overall_results}), demonstrating that this fallback mechanism does not introduce systematic bias favoring the judge model.

\subsection{Human Annotation Baseline}
\label{app:human_baseline}


\begin{table}[h]
\centering
\caption{Human performance and time cost on \polar{} by task category. $-\Delta Acc$ denotes the accuracy gap between Cartesian and Polar.}
\label{tab:suppl_human_validation}
\resizebox{0.8\linewidth}{!}{%
\begin{tabular}{l cc cc cc c cc}
\toprule
 & \multicolumn{2}{c}{Correct (\%)} & \multicolumn{2}{c}{I Don't Know (\%)} & \multicolumn{2}{c}{Incorrect (\%)} & & \multicolumn{2}{c}{Avg. Time (min)} \\
\cmidrule(lr){2-3} \cmidrule(lr){4-5} \cmidrule(lr){6-7} \cmidrule(lr){9-10}
Category & C & P & C & P & C & P & $-\Delta$ Acc. & C & P \\
\midrule
Algo. Logic & 92.5 & 87.5 & 5.8 & 10.0 & 1.7 & 2.5 & -5.0 & 19.7 & 26.6 \\
Combin. \& Prob. & 77.8 & 64.4 & 21.7 & 35.0 & 0.6 & 0.6 & -13.3 & 23.1 & 30.0 \\
Nav. \& Routing & 97.5 & 92.5 & 1.9 & 4.1 & 0.6 & 3.4 & -5.0 & 18.5 & 21.1 \\
Spatial Trans. & 98.8 & 94.6 & 0.8 & 4.2 & 0.4 & 1.2 & -4.2 & 16.4 & 20.5 \\
Visual Pattern & 100.0 & 97.3 & 0.0 & 2.3 & 0.0 & 0.5 & -2.7 & 18.6 & 17.4 \\
\midrule
\textbf{Overall} & 94.5 & 88.8 & 4.9 & 9.5 & 0.5 & 1.7 & -5.7 & 18.9 & 22.3 \\
\bottomrule
\end{tabular}}
\end{table}

\begin{figure}[h]
    \centering
    \begin{minipage}{0.48\linewidth}
        \centering
        \includegraphics[width=\linewidth]{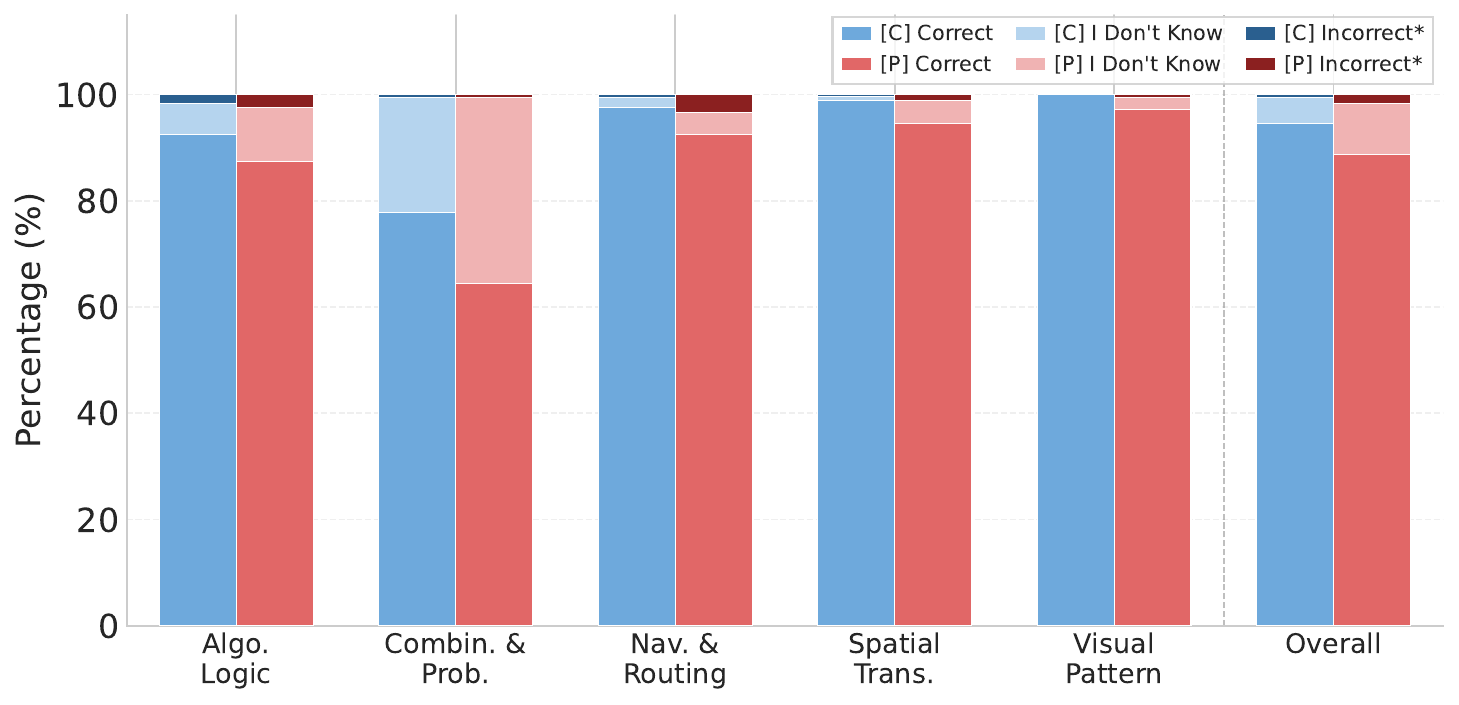}
        {(a)}
    \end{minipage}
    \hfill
    \begin{minipage}{0.48\linewidth}
        \centering
        \includegraphics[width=\linewidth]{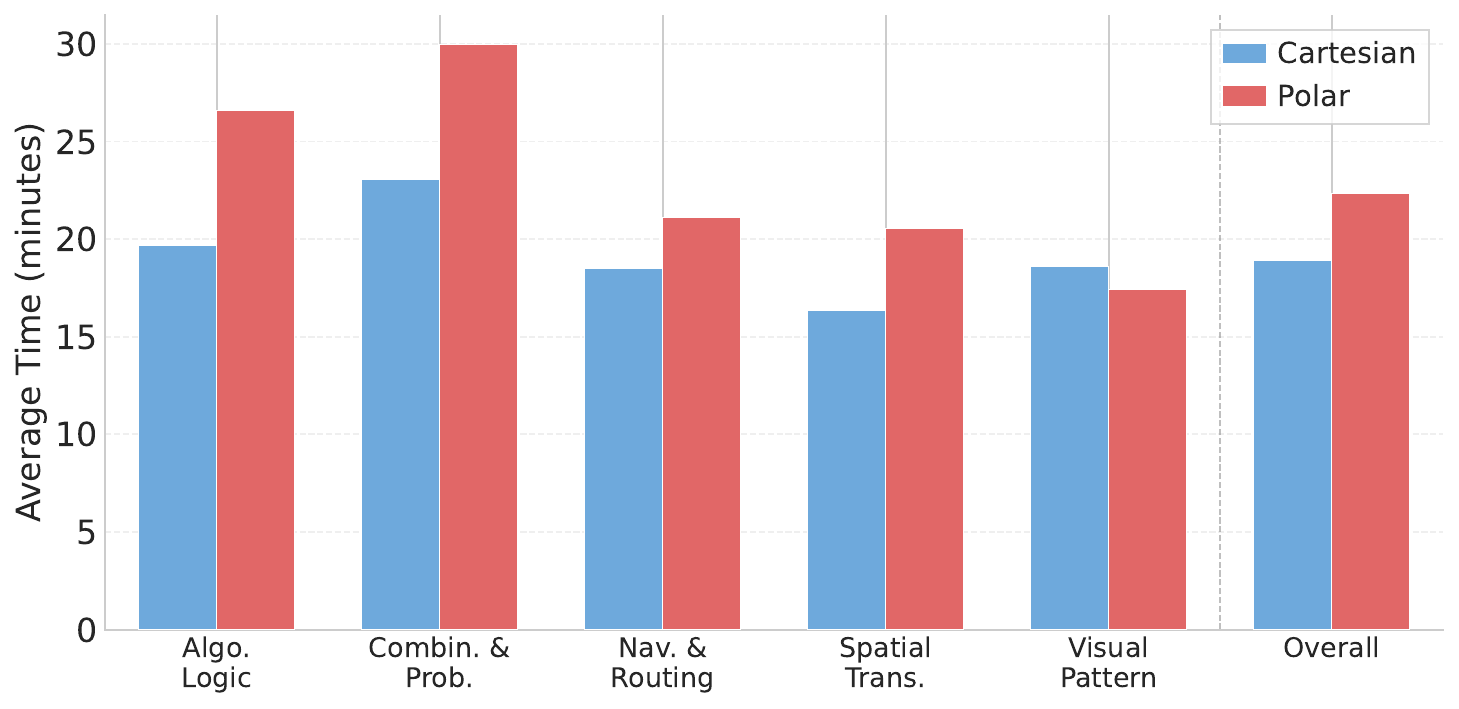}
        (b)
    \end{minipage}
    \caption{Human performance on \polar{} across task categories. (a) Answer correctness distribution, with stacked segments showing Correct, I Don't Know, and Incorrect rates for Cartesian (blue) and Polar (red) layouts. (b) Average annotation time per instance in minutes.}
    \label{fig:suppl_human_annotation}
\end{figure}

 

To establish a human performance reference, we analyze rater performance from our human validation study (Appendix~\ref{app:human_validation}), covering 20 randomly sampled Cartesian--Polar question pairs for each of the 53 tasks in Polaris-Bench. Raters were general-audience annotators without specialized STEM backgrounds, and Cartesian and Polar versions were evaluated separately. For unfamiliar problem types (e.g., Sudoku validity checking, N-Queens constraint verification), raters were permitted to perform basic web searches to understand task rules, but were required to independently derive and write a complete step-by-step solution trace. Raters were instructed to submit a final answer only when fully confident in their derivation, and to select ``I don't know'' otherwise. Since author verification confirmed the ground-truth validity of all sampled instances (Appendix~\ref{app:human_validation}), instances where a rater's complete derivation arrived at the ground truth are recorded as ``Correct,'' while ``Incorrect'' and ``I don't know'' reflect rater solving errors and opt-outs, respectively.

Table~\ref{tab:suppl_human_validation} and Figure~\ref{fig:suppl_human_annotation} report the results by task category. Overall, human raters achieve 94.5\% accuracy on Cartesian instances and 88.8\% on Polar instances, representing a modest 5.7 percentage point reduction, while the ``I don't know'' rate increases moderately from 4.9\% to 9.5\%. The average solving time is 18.9 minutes per Cartesian instance and 22.3 minutes per Polar instance; these reported times measure the rater's active problem-solving duration on the final accepted submission, excluding quality-checker review and intermediate revision cycles. Across individual categories, the Cartesian-to-Polar accuracy gap remains consistently small. \emph{Visual Pattern Matching} maintains near-ceiling accuracy on both layouts (100.0\% and 97.3\%) and requires less time on Polar instances (17.4 vs.\ 18.6 minutes), aligning with the intuition that visual pattern recognition is a fast, high-confidence process for humans. Conversely, \emph{Combinatorics and Probability} exhibits the lowest human accuracy (77.8\% and 64.4\%) and the highest ``I don't know'' rates (21.7\% and 35.0\%), reflecting the demanding mathematical counting and enumeration required for general-audience raters. Even in this most challenging category, however, the Cartesian-to-Polar gap remains moderate ($-13.3$ points) compared to the severe degradation seen in models.

Overall, human performance exhibits only a minor accuracy reduction and a modest 3.4-minute increase in solving time when transitioning from Cartesian to Polar layouts, in stark contrast to the 30--47 percentage point collapse observed across frontier MLLMs (Table~\ref{tab:overall_results}). This provides a clear human reference point contextualizing the model evaluation results in Section~\ref{sec:experiments}.

\subsection{Evaluation Results Statistics}

All results reported in this work are based on a single evaluation run per model, because repeated runs are very costly. The 53 tasks in {\polar} require complex visual reasoning, and under high reasoning mode models often produce long reasoning traces, so a single query can take several minutes and consume a large output token budget. Two situations required re-querying. First, when a response reached the maximum output token limit without a final answer, we re-ran that instance under the same setting. Second, some API requests failed (e.g., server errors or timeouts). If fewer than 5 of the 100 instances of a task failed, we computed accuracy over the completed responses; otherwise, we re-sent the failed requests. Both rules were applied in the same way to Cartesian and Polar instances.

\paragraph{Spot-check verification.}
To estimate run-to-run variation, we repeated the evaluation of selected models on randomly sampled tasks under identical settings. Task-level accuracy varied by a few percentage points~(approximately within 3) across runs. For example, two repeated runs of Gemini-3.1-Pro on Four Color gave 92\% and 90\% on Cartesian and 31\% and 30\% on Polar.

The following factors further support the reliability of our single-run results:
\begin{itemize}[leftmargin=*]
    \item \textbf{Large sample size.} Each task has 100 paired Cartesian--Polar instances, and aggregate results are computed over 5,300 instances per coordinate condition, which reduces instance-level noise.
    \item \textbf{Large effect sizes.} For frontier models, the Cartesian-to-Polar drops range from 37 to 47 percentage points (Table~1). Run-to-run differences of a few points at the task level, as observed in our spot checks, are small relative to these gaps and do not change the conclusions in the main text.

    \item \textbf{Cross-model consistency.} Every evaluated model scores lower on Polar than on Cartesian layouts, across both proprietary and open-source models. The only negligible drop is Mistral-Small-2503 (0.4 points), which is close to the random baseline on both layouts.

    \item \textbf{Consistency across conditions.} For frontier models, the drop appears in all five task categories (Table~1). For all six models in Figure~5, it appears in both fully and partially aligned tasks and under both bounded and wrapping conditions.
\end{itemize}

\section{Complete Experimental Results}

\subsection{Overall Performance Analysis}

\begin{figure}[h]
    \centering
    \includegraphics[width=\linewidth]{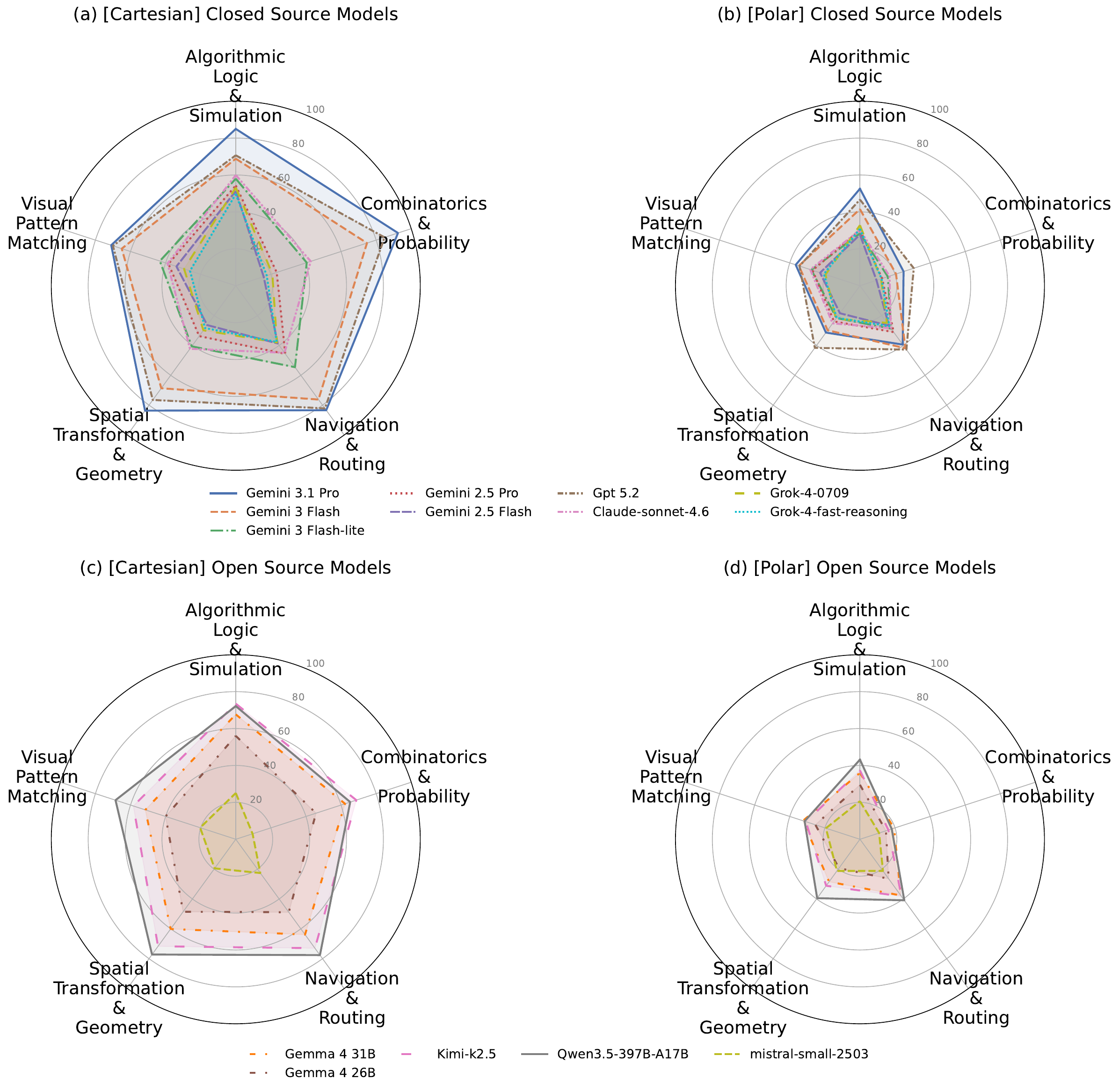}
    \caption{Per-category radar chart comparison of all evaluated models under high reasoning mode, corresponding to Table~\ref{tab:overall_results}. \textit{(a)} Closed-source models on Cartesian layouts. \textit{(b)} Closed-source models on Polar layouts. \textit{(c)} Open-source models on Cartesian layouts. \textit{(d)} Open-source models on Polar layouts. Each axis represents one of the five task categories, and the radial scale indicates accuracy (\%).}
    \label{fig:suppl_radar_chart_comparison}
\end{figure}

\begin{figure}[h]
    \centering
    \includegraphics[width=\linewidth]{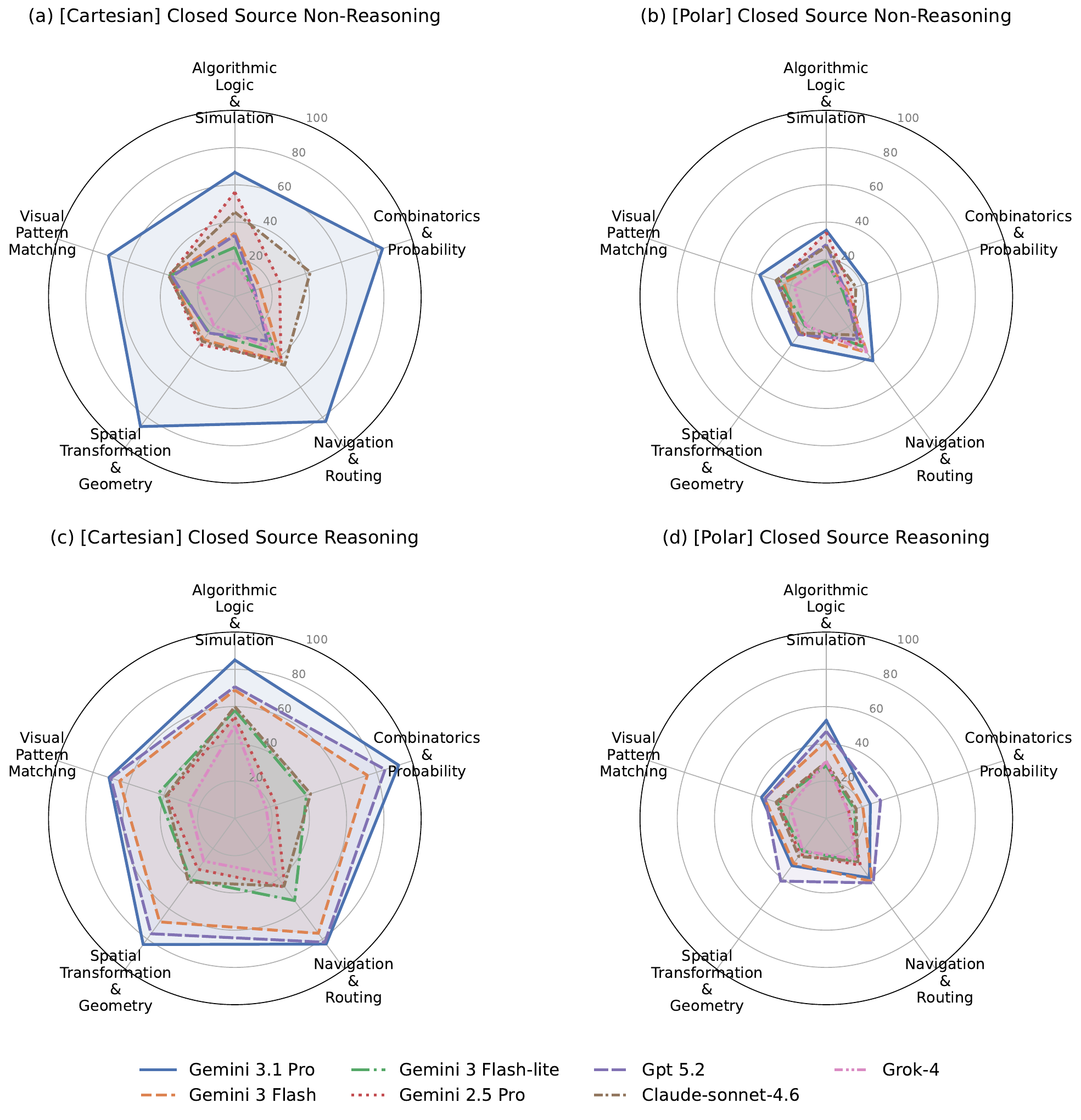}
    \caption{Effect of reasoning mode on per-category performance for closed-source models, corresponding to Table~\ref{tab:reasoning_performance}. \textit{(a)} Non-reasoning mode on Cartesian layouts. \textit{(b)} Non-reasoning mode on Polar layouts. \textit{(c)} High reasoning mode on Cartesian layouts. \textit{(d)} High reasoning mode on Polar layouts.}
    \label{fig:suppl_thinking_mode_radar_closed}
\end{figure}

\begin{figure}[h]
    \centering
    \includegraphics[width=\linewidth]{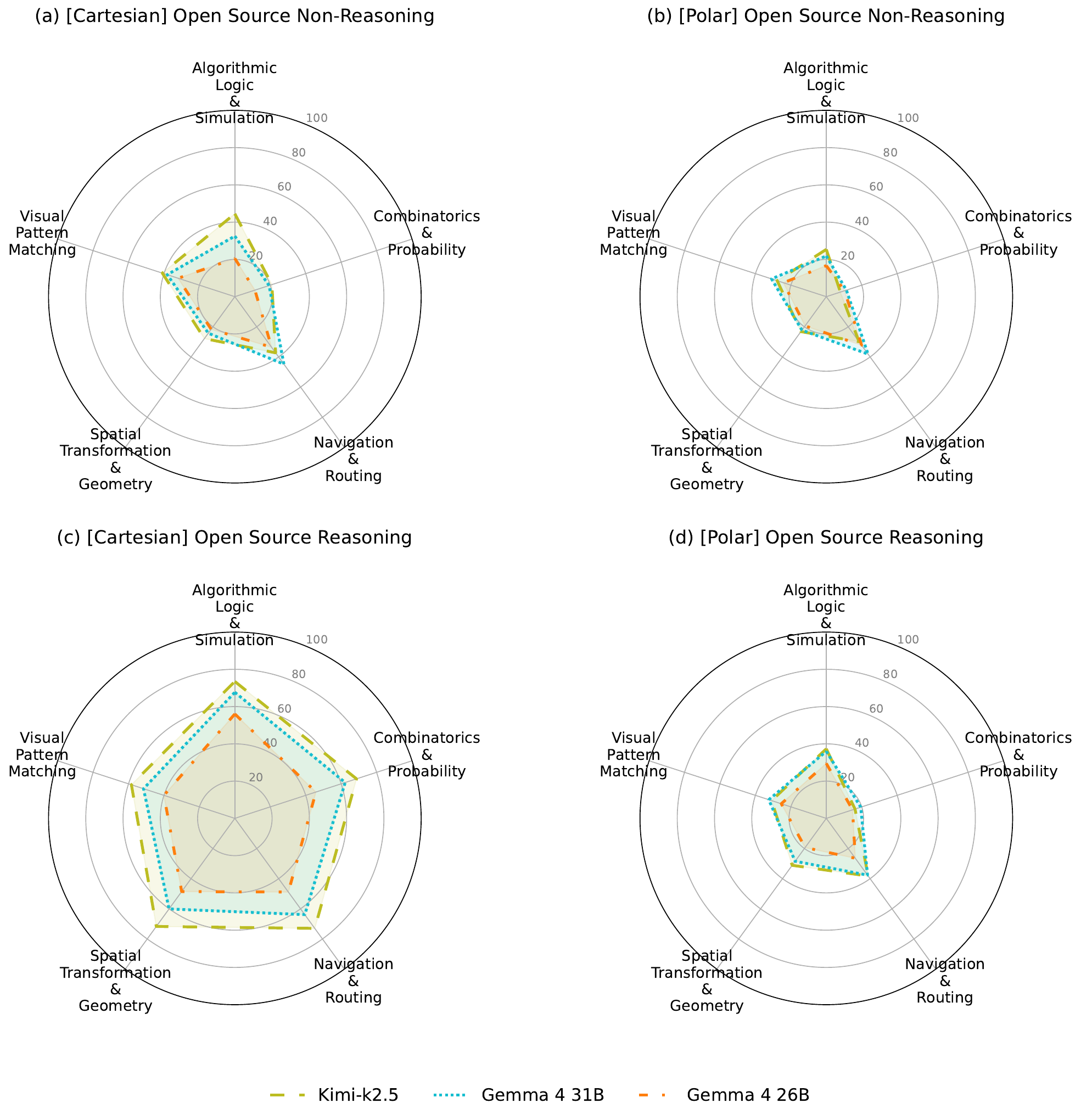}
    \caption{Effect of reasoning mode on per-category performance for open-source models, corresponding to Table~\ref{tab:reasoning_performance}. \textit{(a)} Non-reasoning mode on Cartesian layouts. \textit{(b)} Non-reasoning mode on Polar layouts. \textit{(c)} High reasoning mode on Cartesian layouts. \textit{(d)} High reasoning mode on Polar layouts.}
    \label{fig:suppl_thinking_mode_radar_open}
\end{figure}

\begin{figure}
    \centering
    \includegraphics[width=\linewidth]{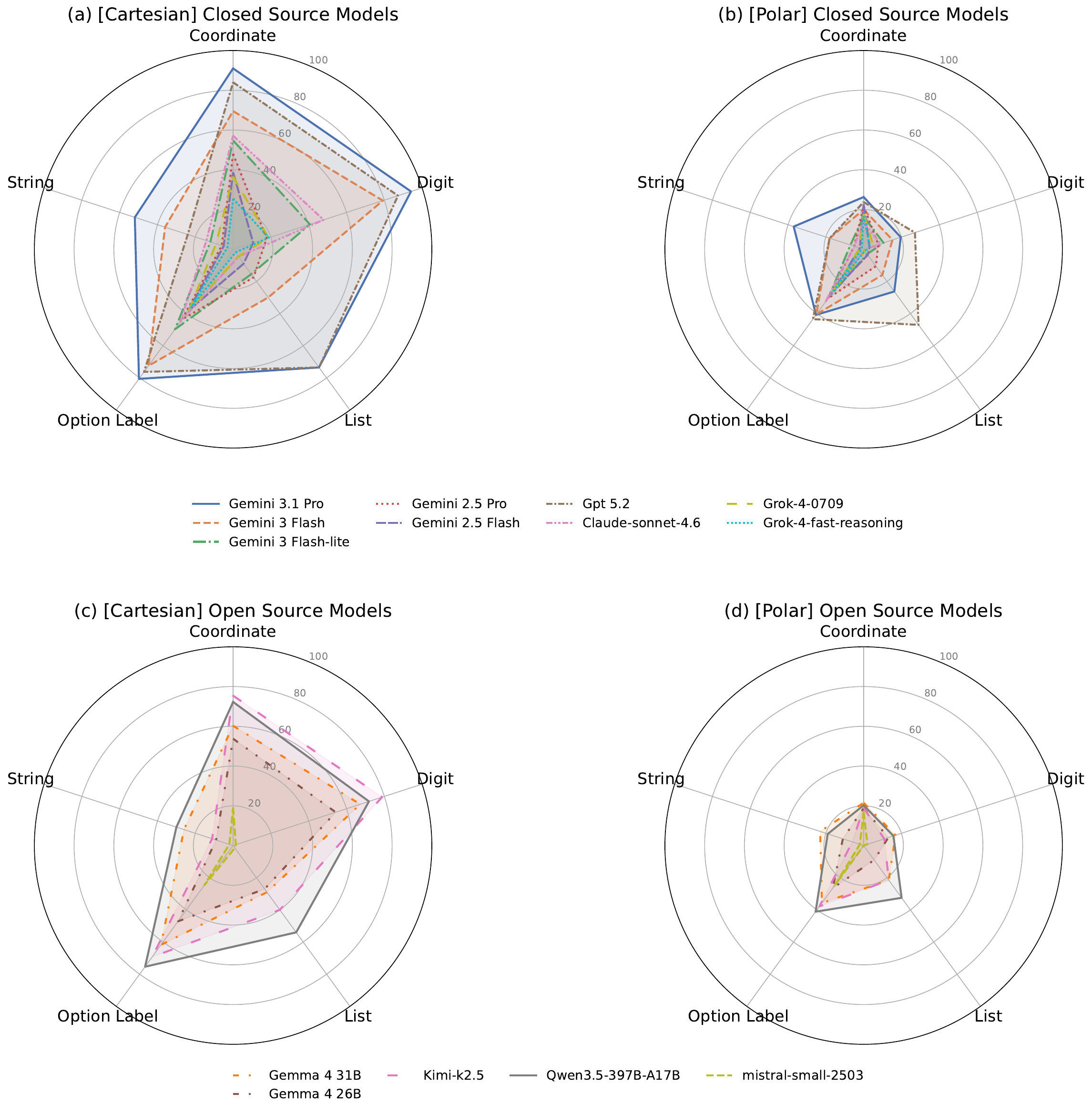}
    \caption{Per-answer-type radar chart comparison of all evaluated models under high reasoning mode. \textit{(a)} Closed-source models on Cartesian layouts. \textit{(b)} Closed-source models on Polar layouts. \textit{(c)} Open-source models on Cartesian layouts. \textit{(d)} Open-source models on Polar layouts. Each axis represents one of the five answer types (Coordinate, Digit, String, Option Label, List), and the radial scale indicates accuracy (\%).}
\label{fig:suppl_radar_chart_answer_type}
\end{figure}

We provide fine-grained radar chart visualizations to complement the aggregate results reported in Table~\ref{tab:overall_results} and Table~\ref{tab:reasoning_performance}. These charts display per-category accuracy across the five task taxonomies~(\ie visual pattern matching, spatial trasnformation and geometry, navigation and routing, combinatorics and probability, and algorithmic logic and simulation) enabling direct visual comparison of model performance profiles.

Figure~\ref{fig:suppl_radar_chart_comparison} presents the per-category performance of all evaluated models under high reasoning mode, separated into closed-source (top row) and open-source (bottom row) groups. The left column shows Cartesian accuracy and the right column shows Polar accuracy. On Cartesian layouts, closed-source models exhibit relatively expanded and overlapping radar profiles, with Gemini-3.1-Pro achieving the largest coverage. On Polar layouts, all radar profiles contract substantially and cluster closer to the center. Among open-source models, a similar contraction pattern is observed, with Polar profiles notably smaller and more compressed than their Cartesian counterparts across all five categories.

Figures~\ref{fig:suppl_thinking_mode_radar_closed} and~\ref{fig:suppl_thinking_mode_radar_open} present the effect of reasoning mode on per-category performance for closed-source and open-source models, respectively. Each figure contains four panels: the top row shows non-reasoning mode results on Cartesian (left) and Polar (right) layouts, while the bottom row shows the corresponding high reasoning mode results. For closed-source models (Figure~\ref{fig:suppl_thinking_mode_radar_closed}), the radar profiles expand visibly from non-reasoning to reasoning on Cartesian layouts, whereas the expansion on Polar layouts is comparatively modest. For open-source models (Figure~\ref{fig:suppl_thinking_mode_radar_open}), the reasoning mode similarly produces larger radar profiles on Cartesian layouts, while the Polar profiles remain relatively compact across both reasoning settings.

Figure~\ref{fig:suppl_radar_chart_answer_type} further disaggregates model performance by answer type rather than task category, with axes corresponding to Coordinate, Digit, String, Option Label, and List. On Cartesian layouts, both closed-source and open-source models exhibit relatively expanded radar profiles with visible inter-model variation across answer types. Upon transition to Polar layouts, all profiles contract substantially, and the degree of contraction varies across answer types, indicating that the Cartesian-to-Polar performance gap is not uniform across output formats. Among open-source models, the Polar profiles are particularly compressed, with reduced inter-model differentiation compared to their Cartesian counterparts.

\subsection{Per-Catagory Breakdown Performance}

\begin{table}[h]
\centering
\caption{Per-task results under the Navigation and Routing category, subcategory Rule-Based Navigation. All models are evaluated under high reasoning mode. C and P denote Cartesian and Polar accuracy (\%), respectively. $(-\Delta)$ indicates the accuracy change from Cartesian to Polar. Task abbreviations: Abs. Nav.~= Absolute Navigation; Ego. Nav.~= Egocentric Navigation; Rule Nav.~= Rule-Based Navigation; Wall Fol.~= Wall Follower; Wrap. Nav.~= Wrapping Navigation.}
\label{tab:suppl_nav_rule_nav}
\resizebox{\linewidth}{!}{
\begin{tabular}{@{} l c c@{\ }l c c@{\ }l c c@{\ }l c c@{\ }l c c@{\ }l @{}}
\toprule
\multirow{2}{*}{\textbf{Model}} & \multicolumn{3}{c}{\textbf{Abs. Nav.}} & \multicolumn{3}{c}{\textbf{Ego. Nav.}} & \multicolumn{3}{c}{\textbf{Rule Nav.}} & \multicolumn{3}{c}{\textbf{Wall Fol.}} & \multicolumn{3}{c}{\textbf{Wrap. Nav.}} \\
\cmidrule(lr){2-4} \cmidrule(lr){5-7} \cmidrule(lr){8-10} \cmidrule(lr){11-13} \cmidrule(lr){14-16} 
 & \textbf{C} & \textbf{P} & $(-\Delta)$ & \textbf{C} & \textbf{P} & $(-\Delta)$ & \textbf{C} & \textbf{P} & $(-\Delta)$ & \textbf{C} & \textbf{P} & $(-\Delta)$ & \textbf{C} & \textbf{P} & $(-\Delta)$ \\
\midrule
Human & 90.0 & 75.0 & \dt{$-15.0$} & 90.0 & 75.0 & \dt{$-15.0$} & 100.0 & 60.0 & \dt{$-40.0$} & 95.0 & 75.0 & \dt{$-20.0$} & 100.0 & 100.0 & \dt{$0.0$} \\
Random & 16.7 & 16.7 & \dt{$0.0$} & 16.7 & 16.7 & \dt{$0.0$} & 20.0 & 20.0 & \dt{$0.0$} & 20.0 & 20.0 & \dt{$0.0$} & 16.7 & 16.7 & \dt{$0.0$} \\
\midrule
\multicolumn{16}{c}{\textit{Closed-Source MLLMs}} \\
\midrule
Gemini-3.1-Pro & 96.0 & 31.0 & \dt{$-65.0$} & 94.0 & 32.0 & \dt{$-62.0$} & 95.0 & 12.0 & \dt{$-83.0$} & 89.0 & 52.5 & \dt{$-36.5$} & 99.0 & 24.5 & \dt{$-74.5$} \\
Gemini-3-Flash & 81.0 & 16.0 & \dt{$-65.0$} & 76.0 & 21.0 & \dt{$-55.0$} & 88.0 & 29.0 & \dt{$-59.0$} & 63.0 & 19.0 & \dt{$-44.0$} & 76.0 & 25.0 & \dt{$-51.0$} \\
Gemini-3-Flash-lite & 47.0 & 15.0 & \dt{$-32.0$} & 34.0 & 10.0 & \dt{$-24.0$} & 80.0 & 17.0 & \dt{$-63.0$} & 48.0 & 31.0 & \dt{$-17.0$} & 40.0 & 19.0 & \dt{$-21.0$} \\
Gemini-2.5-Pro & 23.0 & 16.0 & \dt{$-7.0$} & 13.0 & 7.0 & \dt{$-6.0$} & 85.0 & 15.0 & \dt{$-70.0$} & 48.0 & 47.0 & \dt{$-1.0$} & 36.0 & 25.0 & \dt{$-11.0$} \\
Gemini-2.5-Flash & 20.0 & 13.0 & \dt{$-7.0$} & 8.0 & 12.0 & \dt{$4.0$} & 64.0 & 16.0 & \dt{$-48.0$} & 52.0 & 50.0 & \dt{$-2.0$} & 15.0 & 15.0 & \dt{$0.0$} \\
GPT-5.2 & 91.0 & 29.0 & \dt{$-62.0$} & 71.0 & 27.0 & \dt{$-44.0$} & 96.0 & 32.0 & \dt{$-64.0$} & 88.0 & 29.0 & \dt{$-59.0$} & 93.0 & 36.0 & \dt{$-57.0$} \\
Claude-Sonnet-4.6 & 75.3 & 23.2 & \dt{$-52.1$} & 43.8 & 17.4 & \dt{$-26.5$} & 24.7 & 19.8 & \dt{$-5.0$} & 42.7 & 23.7 & \dt{$-19.0$} & 52.0 & 21.2 & \dt{$-30.8$} \\
Grok-4-0709 & 19.0 & 13.0 & \dt{$-6.0$} & 9.0 & 6.0 & \dt{$-3.0$} & 61.6 & 10.0 & \dt{$-51.6$} & 47.0 & 42.0 & \dt{$-5.0$} & 16.0 & 16.0 & \dt{$0.0$} \\
Grok-4-Fast-Reasoning & 25.0 & 25.0 & \dt{$0.0$} & 21.0 & 28.0 & \dt{$7.0$} & 56.0 & 31.0 & \dt{$-25.0$} & 23.0 & 23.0 & \dt{$0.0$} & 21.0 & 28.0 & \dt{$7.0$} \\
\midrule
\multicolumn{16}{c}{\textit{Open-Source MLLMs}} \\
\midrule
Gemma-4-31B & 75.0 & 15.0 & \dt{$-60.0$} & 52.0 & 22.0 & \dt{$-30.0$} & 75.0 & 12.0 & \dt{$-63.0$} & 55.0 & 48.0 & \dt{$-7.0$} & 71.0 & 28.0 & \dt{$-43.0$} \\
Gemma-4-26B & 52.0 & 8.0 & \dt{$-44.0$} & 13.1 & 9.0 & \dt{$-4.1$} & 66.0 & 11.0 & \dt{$-55.0$} & 44.0 & 37.0 & \dt{$-7.0$} & 59.0 & 10.0 & \dt{$-49.0$} \\
Kimi-k2.5 & 94.9 & 27.4 & \dt{$-67.5$} & 93.0 & 16.3 & \dt{$-76.7$} & 83.0 & 26.0 & \dt{$-57.0$} & 53.0 & 28.0 & \dt{$-25.0$} & 95.0 & 26.0 & \dt{$-69.0$} \\
Qwen3.5-397B-A17B & 95.0 & 17.0 & \dt{$-78.0$} & 95.0 & 25.0 & \dt{$-70.0$} & 80.0 & 14.0 & \dt{$-66.0$} & 73.0 & 40.0 & \dt{$-33.0$} & 91.0 & 33.0 & \dt{$-58.0$} \\
Mistral-Small-2503 & 23.9 & 21.3 & \dt{$-2.6$} & 6.1 & 8.1 & \dt{$2.0$} & 7.0 & 17.0 & \dt{$10.0$} & 51.0 & 51.0 & \dt{$0.0$} & 20.0 & 34.0 & \dt{$14.0$} \\
\bottomrule
\end{tabular}
}
\end{table}

\begin{table}[h]
\centering
\caption{Per-task results under the Navigation and Routing category, subcategory Optimal Pathfinding. All models are evaluated under high reasoning mode. C and P denote Cartesian and Polar accuracy (\%), respectively. $(-\Delta)$ indicates the accuracy change from Cartesian to Polar. Task abbreviations: Bound. Path~= Bounded Path Finding; Long. Path~= Longest Path; Short. Path~= Shortest Path; Wrap. Path~= Wrapping Path Finding.}
\label{tab:suppl_nav_optimal_path}
\resizebox{\linewidth}{!}{
\begin{tabular}{@{} l c c@{\ }l c c@{\ }l c c@{\ }l c c@{\ }l c c@{\ }l @{}}
\toprule
\multirow{2}{*}{\textbf{Model}} & \multicolumn{3}{c}{\textbf{Bound. Path}} & \multicolumn{3}{c}{\textbf{Long. Path}} & \multicolumn{3}{c}{\textbf{Maze}} & \multicolumn{3}{c}{\textbf{Short. Path}} & \multicolumn{3}{c}{\textbf{Wrap. Path}} \\
\cmidrule(lr){2-4} \cmidrule(lr){5-7} \cmidrule(lr){8-10} \cmidrule(lr){11-13} \cmidrule(lr){14-16} 
 & \textbf{C} & \textbf{P} & $(-\Delta)$ & \textbf{C} & \textbf{P} & $(-\Delta)$ & \textbf{C} & \textbf{P} & $(-\Delta)$ & \textbf{C} & \textbf{P} & $(-\Delta)$ & \textbf{C} & \textbf{P} & $(-\Delta)$ \\
\midrule
Human & 95.0 & 100.0 & \dt{$5.0$} & 100.0 & 100.0 & \dt{$0.0$} & 100.0 & 100.0 & \dt{$0.0$} & 100.0 & 100.0 & \dt{$0.0$} & 95.0 & 100.0 & \dt{$5.0$} \\
Random & 20.0 & 20.0 & \dt{$0.0$} & 20.0 & 20.0 & \dt{$0.0$} & 31.3 & 31.3 & \dt{$0.0$} & 20.0 & 20.0 & \dt{$0.0$} & 20.0 & 20.0 & \dt{$0.0$} \\
\midrule
\multicolumn{16}{c}{\textit{Closed-Source MLLMs}} \\
\midrule
Gemini-3.1-Pro & 98.0 & 37.0 & \dt{$-61.0$} & 78.0 & 55.0 & \dt{$-23.0$} & 34.0 & 37.5 & \dt{$3.5$} & 69.7 & 55.6 & \dt{$-14.1$} & 98.0 & 48.0 & \dt{$-50.0$} \\
Gemini-3-Flash & 86.0 & 63.0 & \dt{$-23.0$} & 77.0 & 56.0 & \dt{$-21.0$} & 37.5 & 40.5 & \dt{$3.0$} & 93.0 & 73.0 & \dt{$-20.0$} & 92.0 & 74.0 & \dt{$-18.0$} \\
Gemini-3-Flash-lite & 64.0 & 29.0 & \dt{$-35.0$} & 56.0 & 48.0 & \dt{$-8.0$} & 37.0 & 35.5 & \dt{$-1.5$} & 68.0 & 70.0 & \dt{$2.0$} & 51.0 & 27.0 & \dt{$-24.0$} \\
Gemini-2.5-Pro & 33.0 & 23.0 & \dt{$-10.0$} & 69.0 & 46.0 & \dt{$-23.0$} & 32.0 & 37.5 & \dt{$5.5$} & 65.0 & 54.0 & \dt{$-11.0$} & 29.0 & 18.0 & \dt{$-11.0$} \\
Gemini-2.5-Flash & 29.0 & 21.0 & \dt{$-8.0$} & 48.0 & 32.0 & \dt{$-16.0$} & 39.5 & 39.0 & \dt{$-0.5$} & 46.0 & 43.0 & \dt{$-3.0$} & 31.0 & 20.0 & \dt{$-11.0$} \\
GPT-5.2 & 86.0 & 31.0 & \dt{$-55.0$} & 80.0 & 69.0 & \dt{$-11.0$} & 44.5 & 36.5 & \dt{$-8.0$} & 82.0 & 87.0 & \dt{$5.0$} & 97.0 & 32.0 & \dt{$-65.0$} \\
Claude-Sonnet-4.6 & 59.8 & 49.0 & \dt{$-10.8$} & 48.5 & 44.4 & \dt{$-4.0$} & 34.7 & 33.7 & \dt{$-1.0$} & 65.0 & 56.1 & \dt{$-8.9$} & 56.0 & 54.1 & \dt{$-1.9$} \\
Grok-4-0709 & 20.0 & 25.0 & \dt{$5.0$} & 43.4 & 40.8 & \dt{$-2.6$} & 39.5 & 32.0 & \dt{$-7.5$} & 40.0 & 43.0 & \dt{$3.0$} & 29.0 & 33.0 & \dt{$4.0$} \\
Grok-4-Fast-Reasoning & 29.0 & 25.0 & \dt{$-4.0$} & 37.0 & 40.0 & \dt{$3.0$} & 33.5 & 39.0 & \dt{$5.5$} & 41.0 & 40.0 & \dt{$-1.0$} & 22.0 & 13.0 & \dt{$-9.0$} \\
\midrule
\multicolumn{16}{c}{\textit{Open-Source MLLMs}} \\
\midrule
Gemma-4-31B & 72.0 & 41.0 & \dt{$-31.0$} & 71.0 & 50.0 & \dt{$-21.0$} & 42.0 & 34.0 & \dt{$-8.0$} & 62.0 & 66.0 & \dt{$4.0$} & 72.0 & 43.0 & \dt{$-29.0$} \\
Gemma-4-26B & 48.0 & 20.2 & \dt{$-27.8$} & 48.0 & 26.0 & \dt{$-22.0$} & 27.5 & 35.0 & \dt{$7.5$} & 49.0 & 52.0 & \dt{$3.0$} & 43.4 & 30.0 & \dt{$-13.4$} \\
Kimi-k2.5 & 93.0 & 47.0 & \dt{$-46.0$} & 68.0 & 40.8 & \dt{$-27.2$} & 43.2 & 35.7 & \dt{$-7.5$} & 68.0 & 65.0 & \dt{$-3.0$} & 90.0 & 51.0 & \dt{$-39.0$} \\
Qwen3.5-397B-A17B & 92.0 & 52.0 & \dt{$-40.0$} & 76.0 & 57.0 & \dt{$-19.0$} & 38.5 & 42.5 & \dt{$4.0$} & 66.0 & 58.0 & \dt{$-8.0$} & 88.0 & 45.0 & \dt{$-43.0$} \\
Mistral-Small-2503 & 17.2 & 8.1 & \dt{$-9.1$} & 30.0 & 28.0 & \dt{$-2.0$} & 53.5 & 26.5 & \dt{$-27.0$} & 40.0 & 40.0 & \dt{$0.0$} & 19.0 & 16.0 & \dt{$-3.0$} \\
\bottomrule
\end{tabular}
}
\end{table}

\begin{table}[h]
\centering
\caption{Per-task results under the Navigation and Routing category, subcategory Constraint-Based Routing. All models are evaluated under high reasoning mode. C and P denote Cartesian and Polar accuracy (\%), respectively. $(-\Delta)$ indicates the accuracy change from Cartesian to Polar. Task abbreviations: Lrg. Num.~= Largest Number Path; Mono. Path~= Monotonic Path; Turn Cnt.~= Turn Counting; Word Srch.~= Word Search.}
\label{tab:suppl_nav_constraint_route}
\resizebox{0.85\linewidth}{!}{
\begin{tabular}{@{} l c c@{\ }l c c@{\ }l c c@{\ }l c c@{\ }l @{}}
\toprule
\multirow{2}{*}{\textbf{Model}} & \multicolumn{3}{c}{\textbf{Lrg. Num.}} & \multicolumn{3}{c}{\textbf{Mono. Path}} & \multicolumn{3}{c}{\textbf{Turn Cnt.}} & \multicolumn{3}{c}{\textbf{Word Srch.}} \\
\cmidrule(lr){2-4} \cmidrule(lr){5-7} \cmidrule(lr){8-10} \cmidrule(lr){11-13} 
 & \textbf{C} & \textbf{P} & $(-\Delta)$ & \textbf{C} & \textbf{P} & $(-\Delta)$ & \textbf{C} & \textbf{P} & $(-\Delta)$ & \textbf{C} & \textbf{P} & $(-\Delta)$ \\
\midrule
Human & 100.0 & 100.0 & \dt{$0.0$} & 100.0 & 100.0 & \dt{$0.0$} & 100.0 & 100.0 & \dt{$0.0$} & 95.0 & 95.0 & \dt{$0.0$} \\
Random & 20.0 & 20.0 & \dt{$0.0$} & 16.7 & 16.7 & \dt{$0.0$} & 20.0 & 20.0 & \dt{$0.0$} & 0.0 & 0.0 & \dt{$0.0$} \\
\midrule
\multicolumn{13}{c}{\textit{Closed-Source MLLMs}} \\
\midrule
Gemini-3.1-Pro & 71.0 & 29.0 & \dt{$-42.0$} & 89.0 & 91.0 & \dt{$2.0$} & 57.0 & 27.0 & \dt{$-30.0$} & 100.0 & 20.0 & \dt{$-80.0$} \\
Gemini-3-Flash & 48.0 & 26.0 & \dt{$-22.0$} & 87.0 & 94.0 & \dt{$7.0$} & 65.0 & 23.0 & \dt{$-42.0$} & 98.0 & 27.0 & \dt{$-71.0$} \\
Gemini-3-Flash-lite & 34.0 & 22.0 & \dt{$-12.0$} & 90.0 & 52.0 & \dt{$-38.0$} & 18.0 & 13.0 & \dt{$-5.0$} & 97.0 & 14.0 & \dt{$-83.0$} \\
Gemini-2.5-Pro & 27.0 & 23.0 & \dt{$-4.0$} & 89.0 & 80.0 & \dt{$-9.0$} & 37.0 & 21.0 & \dt{$-16.0$} & 53.0 & 23.0 & \dt{$-30.0$} \\
Gemini-2.5-Flash & 28.0 & 23.0 & \dt{$-5.0$} & 90.0 & 63.0 & \dt{$-27.0$} & 21.0 & 21.0 & \dt{$0.0$} & 50.0 & 7.0 & \dt{$-43.0$} \\
GPT-5.2 & 69.0 & 29.0 & \dt{$-40.0$} & 84.0 & 93.0 & \dt{$9.0$} & 71.0 & 33.0 & \dt{$-38.0$} & 98.0 & 36.0 & \dt{$-62.0$} \\
Claude-Sonnet-4.6 & 25.3 & 19.0 & \dt{$-6.3$} & 1.0 & 0.0 & \dt{$-1.0$} & 29.0 & 24.2 & \dt{$-4.8$} & 71.6 & 13.4 & \dt{$-58.2$} \\
Grok-4-0709 & 17.0 & 22.0 & \dt{$5.0$} & 82.0 & 45.9 & \dt{$-36.1$} & 16.0 & 7.0 & \dt{$-9.0$} & 92.0 & 13.0 & \dt{$-79.0$} \\
Grok-4-Fast-Reasoning & 23.0 & 18.0 & \dt{$-5.0$} & 86.0 & 43.0 & \dt{$-43.0$} & 21.0 & 22.0 & \dt{$1.0$} & 89.0 & 4.0 & \dt{$-85.0$} \\
\midrule
\multicolumn{13}{c}{\textit{Open-Source MLLMs}} \\
\midrule
Gemma-4-31B & 31.0 & 22.0 & \dt{$-9.0$} & 84.0 & 90.0 & \dt{$6.0$} & 36.0 & 17.0 & \dt{$-19.0$} & 96.0 & 39.0 & \dt{$-57.0$} \\
Gemma-4-26B & 27.0 & 27.0 & \dt{$0.0$} & 89.0 & 77.0 & \dt{$-12.0$} & 20.2 & 14.0 & \dt{$-6.2$} & 98.0 & 14.1 & \dt{$-83.9$} \\
Kimi-k2.5 & 35.7 & 24.0 & \dt{$-11.7$} & 83.0 & 87.0 & \dt{$4.0$} & 25.0 & 18.0 & \dt{$-7.0$} & 96.9 & 43.4 & \dt{$-53.5$} \\
Qwen3.5-397B-A17B & 68.0 & 34.0 & \dt{$-34.0$} & 82.0 & 94.0 & \dt{$12.0$} & 54.0 & 21.0 & \dt{$-33.0$} & 88.0 & 40.0 & \dt{$-48.0$} \\
Mistral-Small-2503 & 21.0 & 19.0 & \dt{$-2.0$} & 0.0 & 0.0 & \dt{$0.0$} & 23.0 & 16.0 & \dt{$-7.0$} & 5.0 & 12.0 & \dt{$7.0$} \\
\bottomrule
\end{tabular}
}
\end{table}

\begin{table}[h]
\centering
\caption{Per-task results under the Visual Pattern Matching category, subcategory Visual Spotting and Search / Pattern Continuation. All models are evaluated under high reasoning mode. C and P denote Cartesian and Polar accuracy (\%), respectively. $(-\Delta)$ indicates the accuracy change from Cartesian to Polar. Task abbreviations: Anom. Det.~= Anomaly Detection; Letter Col.~= Letter Collection; Odd Piece~= Odd Piece Out; Pat. Pred.~= Pattern Prediction.}
\label{tab:suppl_vpm_spotting_pattern}
\resizebox{0.85\linewidth}{!}{
\begin{tabular}{@{} l c c@{\ }l c c@{\ }l c c@{\ }l c c@{\ }l @{}}
\toprule
\multirow{2}{*}{\textbf{Model}} & \multicolumn{3}{c}{\textbf{Anom. Det.}} & \multicolumn{3}{c}{\textbf{Letter Col.}} & \multicolumn{3}{c}{\textbf{Odd Piece}} & \multicolumn{3}{c}{\textbf{Pat. Pred.}} \\
\cmidrule(lr){2-4} \cmidrule(lr){5-7} \cmidrule(lr){8-10} \cmidrule(lr){11-13} 
 & \textbf{C} & \textbf{P} & $(-\Delta)$ & \textbf{C} & \textbf{P} & $(-\Delta)$ & \textbf{C} & \textbf{P} & $(-\Delta)$ & \textbf{C} & \textbf{P} & $(-\Delta)$ \\
\midrule
Human & 100.0 & 85.0 & \dt{$-15.0$} & 100.0 & 100.0 & \dt{$0.0$} & 100.0 & 100.0 & \dt{$0.0$} & 100.0 & 95.0 & \dt{$-5.0$} \\
Random & 0.9 & 1.0 & \dt{$0.0$} & 0.0 & 0.0 & \dt{$0.0$} & 20.0 & 20.0 & \dt{$0.0$} & 19.7 & 19.6 & \dt{$-0.0$} \\
\midrule
\multicolumn{13}{c}{\textit{Closed-Source MLLMs}} \\
\midrule
Gemini-3.1-Pro & 83.0 & 10.0 & \dt{$-73.0$} & 52.0 & 37.0 & \dt{$-15.0$} & 69.0 & 42.0 & \dt{$-27.0$} & 90.0 & 27.3 & \dt{$-62.7$} \\
Gemini-3-Flash & 78.0 & 4.0 & \dt{$-74.0$} & 36.0 & 18.0 & \dt{$-18.0$} & 65.0 & 36.0 & \dt{$-29.0$} & 83.0 & 31.0 & \dt{$-52.0$} \\
Gemini-3-Flash-lite & 57.0 & 2.0 & \dt{$-55.0$} & 12.0 & 7.0 & \dt{$-5.0$} & 37.0 & 40.0 & \dt{$3.0$} & 32.0 & 26.0 & \dt{$-6.0$} \\
Gemini-2.5-Pro & 27.0 & 0.0 & \dt{$-27.0$} & 6.0 & 5.0 & \dt{$-1.0$} & 36.0 & 46.0 & \dt{$10.0$} & 37.0 & 17.0 & \dt{$-20.0$} \\
Gemini-2.5-Flash & 20.0 & 1.0 & \dt{$-19.0$} & 5.0 & 2.0 & \dt{$-3.0$} & 36.0 & 41.0 & \dt{$5.0$} & 26.0 & 29.0 & \dt{$3.0$} \\
GPT-5.2 & 73.0 & 2.0 & \dt{$-71.0$} & 24.0 & 18.0 & \dt{$-6.0$} & 88.0 & 49.0 & \dt{$-39.0$} & 54.0 & 13.0 & \dt{$-41.0$} \\
Claude-Sonnet-4.6 & 74.8 & 1.0 & \dt{$-73.7$} & 14.1 & 5.1 & \dt{$-9.1$} & 26.0 & 36.0 & \dt{$10.0$} & 19.0 & 14.0 & \dt{$-5.0$} \\
Grok-4-0709 & 16.2 & 2.1 & \dt{$-14.0$} & 9.0 & 2.0 & \dt{$-7.0$} & 26.0 & 29.0 & \dt{$3.0$} & 26.0 & 23.0 & \dt{$-3.0$} \\
Grok-4-Fast-Reasoning & 6.0 & 0.0 & \dt{$-6.0$} & 3.0 & 1.0 & \dt{$-2.0$} & 29.0 & 23.0 & \dt{$-6.0$} & 22.0 & 34.0 & \dt{$12.0$} \\
\midrule
\multicolumn{13}{c}{\textit{Open-Source MLLMs}} \\
\midrule
Gemma-4-31B & 63.0 & 5.0 & \dt{$-58.0$} & 26.0 & 23.0 & \dt{$-3.0$} & 39.0 & 32.0 & \dt{$-7.0$} & 41.0 & 21.0 & \dt{$-20.0$} \\
Gemma-4-26B & 63.0 & 4.1 & \dt{$-58.9$} & 9.0 & 11.0 & \dt{$2.0$} & 26.3 & 29.0 & \dt{$2.7$} & 17.0 & 14.1 & \dt{$-2.9$} \\
Kimi-k2.5 & 75.0 & 4.0 & \dt{$-71.0$} & 11.1 & 6.0 & \dt{$-5.1$} & 53.0 & 47.0 & \dt{$-6.0$} & 19.0 & 13.0 & \dt{$-6.0$} \\
Qwen3.5-397B-A17B & 78.0 & 2.0 & \dt{$-76.0$} & 30.0 & 19.0 & \dt{$-11.0$} & 74.0 & 37.0 & \dt{$-37.0$} & 49.0 & 24.0 & \dt{$-25.0$} \\
Mistral-Small-2503 & 14.1 & 0.0 & \dt{$-14.1$} & 2.0 & 2.0 & \dt{$0.0$} & 27.0 & 38.0 & \dt{$11.0$} & 18.0 & 27.0 & \dt{$9.0$} \\
\bottomrule
\end{tabular}
}
\end{table}

\begin{table}[h]
\centering
\caption{Per-task results under the Visual Pattern Matching category, subcategory Shape and Jigsaw Assembly. All models are evaluated under high reasoning mode. C and P denote Cartesian and Polar accuracy (\%), respectively. $(-\Delta)$ indicates the accuracy change from Cartesian to Polar. Task abbreviations: Frag. Match.~= Fragment Matching; Jigsaw~= Jigsaw Matching; Layer Comp.~= Layer Completion; Pat. Comp.~= Pattern Completion; Shape Comp.~= Shape Completion; Shape Fit.~= Shape Fitting; Tmpl. Match.~= Template Matching.}
\label{tab:suppl_vpm_shape_jigsaw}
\resizebox{\linewidth}{!}{
\begin{tabular}{@{} l c c@{\ }l c c@{\ }l c c@{\ }l c c@{\ }l c c@{\ }l c c@{\ }l c c@{\ }l @{}}
\toprule
\multirow{2}{*}{\textbf{Model}} & \multicolumn{3}{c}{\textbf{Frag. Match.}} & \multicolumn{3}{c}{\textbf{Jigsaw}} & \multicolumn{3}{c}{\textbf{Layer Comp.}} & \multicolumn{3}{c}{\textbf{Pat. Comp.}} & \multicolumn{3}{c}{\textbf{Shape Comp.}} & \multicolumn{3}{c}{\textbf{Shape Fit.}} & \multicolumn{3}{c}{\textbf{Tmpl. Match.}} \\
\cmidrule(lr){2-4} \cmidrule(lr){5-7} \cmidrule(lr){8-10} \cmidrule(lr){11-13} \cmidrule(lr){14-16} \cmidrule(lr){17-19} \cmidrule(lr){20-22} 
 & \textbf{C} & \textbf{P} & $(-\Delta)$ & \textbf{C} & \textbf{P} & $(-\Delta)$ & \textbf{C} & \textbf{P} & $(-\Delta)$ & \textbf{C} & \textbf{P} & $(-\Delta)$ & \textbf{C} & \textbf{P} & $(-\Delta)$ & \textbf{C} & \textbf{P} & $(-\Delta)$ & \textbf{C} & \textbf{P} & $(-\Delta)$ \\
\midrule
Human & 100.0 & 100.0 & \dt{$0.0$} & 100.0 & 100.0 & \dt{$0.0$} & 100.0 & 95.0 & \dt{$-5.0$} & 100.0 & 100.0 & \dt{$0.0$} & 100.0 & 100.0 & \dt{$0.0$} & 100.0 & 100.0 & \dt{$0.0$} & 100.0 & 95.0 & \dt{$-5.0$} \\
Random & 20.0 & 20.0 & \dt{$0.0$} & 20.0 & 20.0 & \dt{$0.0$} & 25.0 & 25.0 & \dt{$0.0$} & 25.0 & 25.0 & \dt{$0.0$} & 25.0 & 25.0 & \dt{$0.0$} & 33.0 & 33.0 & \dt{$0.0$} & 20.0 & 20.0 & \dt{$0.0$} \\
\midrule
\multicolumn{22}{c}{\textit{Closed-Source MLLMs}} \\
\midrule
Gemini-3.1-Pro & 59.0 & 31.0 & \dt{$-28.0$} & 80.0 & 44.0 & \dt{$-36.0$} & 93.0 & 69.7 & \dt{$-23.3$} & 55.0 & 26.0 & \dt{$-29.0$} & 43.0 & 29.3 & \dt{$-13.7$} & 60.0 & 33.0 & \dt{$-27.0$} & 97.0 & 53.5 & \dt{$-43.5$} \\
Gemini-3-Flash & 53.0 & 25.0 & \dt{$-28.0$} & 74.0 & 50.0 & \dt{$-24.0$} & 91.0 & 70.0 & \dt{$-21.0$} & 41.0 & 20.0 & \dt{$-21.0$} & 33.0 & 22.0 & \dt{$-11.0$} & 61.0 & 45.0 & \dt{$-16.0$} & 100.0 & 62.0 & \dt{$-38.0$} \\
Gemini-3-Flash-lite & 24.0 & 30.0 & \dt{$6.0$} & 40.0 & 34.0 & \dt{$-6.0$} & 58.0 & 47.0 & \dt{$-11.0$} & 29.0 & 19.0 & \dt{$-10.0$} & 49.0 & 29.0 & \dt{$-20.0$} & 46.0 & 34.0 & \dt{$-12.0$} & 92.0 & 37.0 & \dt{$-55.0$} \\
Gemini-2.5-Pro & 21.0 & 19.0 & \dt{$-2.0$} & 34.0 & 34.0 & \dt{$0.0$} & 76.0 & 52.0 & \dt{$-24.0$} & 29.0 & 27.0 & \dt{$-2.0$} & 28.0 & 26.0 & \dt{$-2.0$} & 43.0 & 39.0 & \dt{$-4.0$} & 85.0 & 32.0 & \dt{$-53.0$} \\
Gemini-2.5-Flash & 23.0 & 21.0 & \dt{$-2.0$} & 18.0 & 16.0 & \dt{$-2.0$} & 71.0 & 36.0 & \dt{$-35.0$} & 30.0 & 28.0 & \dt{$-2.0$} & 37.0 & 24.0 & \dt{$-13.0$} & 30.0 & 33.0 & \dt{$3.0$} & 76.0 & 19.0 & \dt{$-57.0$} \\
GPT-5.2 & 95.0 & 30.0 & \dt{$-65.0$} & 54.0 & 46.0 & \dt{$-8.0$} & 88.0 & 64.0 & \dt{$-24.0$} & 94.0 & 24.0 & \dt{$-70.0$} & 52.0 & 40.0 & \dt{$-12.0$} & 48.0 & 34.0 & \dt{$-14.0$} & 100.0 & 59.0 & \dt{$-41.0$} \\
Claude-Sonnet-4.6 & 35.0 & 18.6 & \dt{$-16.5$} & 39.2 & 30.9 & \dt{$-8.2$} & 67.0 & 74.8 & \dt{$7.8$} & 31.6 & 24.0 & \dt{$-7.6$} & 41.0 & 29.3 & \dt{$-11.7$} & 49.5 & 42.4 & \dt{$-7.1$} & 31.0 & 35.7 & \dt{$4.7$} \\
Grok-4-0709 & 26.0 & 21.0 & \dt{$-5.0$} & 19.0 & 17.0 & \dt{$-2.0$} & 38.0 & 26.0 & \dt{$-12.0$} & 27.0 & 18.0 & \dt{$-9.0$} & 32.0 & 21.0 & \dt{$-11.0$} & 38.0 & 34.0 & \dt{$-4.0$} & 67.0 & 22.6 & \dt{$-44.4$} \\
Grok-4-Fast-Reasoning & 15.0 & 17.0 & \dt{$2.0$} & 22.0 & 17.0 & \dt{$-5.0$} & 52.0 & 26.0 & \dt{$-26.0$} & 25.0 & 23.0 & \dt{$-2.0$} & 24.0 & 26.0 & \dt{$2.0$} & 38.0 & 37.0 & \dt{$-1.0$} & 51.0 & 22.0 & \dt{$-29.0$} \\
\midrule
\multicolumn{22}{c}{\textit{Open-Source MLLMs}} \\
\midrule
Gemma-4-31B & 34.0 & 19.0 & \dt{$-15.0$} & 50.0 & 47.0 & \dt{$-3.0$} & 80.0 & 60.0 & \dt{$-20.0$} & 50.0 & 34.0 & \dt{$-16.0$} & 29.0 & 26.0 & \dt{$-3.0$} & 60.0 & 39.0 & \dt{$-21.0$} & 95.0 & 51.0 & \dt{$-44.0$} \\
Gemma-4-26B & 31.0 & 19.0 & \dt{$-12.0$} & 34.3 & 26.0 & \dt{$-8.3$} & 64.0 & 54.0 & \dt{$-10.0$} & 29.0 & 27.0 & \dt{$-2.0$} & 22.2 & 22.2 & \dt{$0.0$} & 52.5 & 37.0 & \dt{$-15.5$} & 91.9 & 36.0 & \dt{$-55.9$} \\
Kimi-k2.5 & 70.4 & 31.0 & \dt{$-39.4$} & 46.0 & 28.3 & \dt{$-17.7$} & 91.0 & 66.0 & \dt{$-25.0$} & 81.6 & 25.0 & \dt{$-56.6$} & 37.0 & 29.0 & \dt{$-8.0$} & 63.0 & 40.0 & \dt{$-23.0$} & 95.0 & 50.0 & \dt{$-45.0$} \\
Qwen3.5-397B-A17B & 84.0 & 33.0 & \dt{$-51.0$} & 69.0 & 55.0 & \dt{$-14.0$} & 96.0 & 49.0 & \dt{$-47.0$} & 89.0 & 30.0 & \dt{$-59.0$} & 43.0 & 19.0 & \dt{$-24.0$} & 54.0 & 23.0 & \dt{$-31.0$} & 88.0 & 56.0 & \dt{$-32.0$} \\
Mistral-Small-2503 & 31.3 & 11.3 & \dt{$-20.0$} & 14.0 & 17.0 & \dt{$3.0$} & 7.0 & 17.0 & \dt{$10.0$} & 26.0 & 19.0 & \dt{$-7.0$} & 30.0 & 30.0 & \dt{$0.0$} & 38.0 & 33.0 & \dt{$-5.0$} & 16.0 & 19.0 & \dt{$3.0$} \\
\bottomrule
\end{tabular}
}
\end{table}

\begin{table}[h]
\centering
\caption{Per-task results under the Spatial Transformation and Geometry category, subcategory Geometric Measurement and Counting. All models are evaluated under high reasoning mode. C and P denote Cartesian and Polar accuracy (\%), respectively. $(-\Delta)$ indicates the accuracy change from Cartesian to Polar. Task abbreviations: Area Bal.~= Area Balancing; Area Cnt.~= Area Counting; Curve Len.~= Curve Length; Pipe Len.~= Pipe Lengths.}
\label{tab:suppl_stg_geometric}
\resizebox{\linewidth}{!}{
\begin{tabular}{@{} l c c@{\ }l c c@{\ }l c c@{\ }l c c@{\ }l c c@{\ }l @{}}
\toprule
\multirow{2}{*}{\textbf{Model}} & \multicolumn{3}{c}{\textbf{Area Bal.}} & \multicolumn{3}{c}{\textbf{Area Cnt.}} & \multicolumn{3}{c}{\textbf{Curve Len.}} & \multicolumn{3}{c}{\textbf{Pipe Len.}} & \multicolumn{3}{c}{\textbf{Uncut Cells}} \\
\cmidrule(lr){2-4} \cmidrule(lr){5-7} \cmidrule(lr){8-10} \cmidrule(lr){11-13} \cmidrule(lr){14-16} 
 & \textbf{C} & \textbf{P} & $(-\Delta)$ & \textbf{C} & \textbf{P} & $(-\Delta)$ & \textbf{C} & \textbf{P} & $(-\Delta)$ & \textbf{C} & \textbf{P} & $(-\Delta)$ & \textbf{C} & \textbf{P} & $(-\Delta)$ \\
\midrule
Human & 95.0 & 75.0 & \dt{$-20.0$} & 100.0 & 100.0 & \dt{$0.0$} & 100.0 & 70.0 & \dt{$-30.0$} & 95.0 & 100.0 & \dt{$5.0$} & 100.0 & 100.0 & \dt{$0.0$} \\
Random & 20.0 & 19.9 & \dt{$-0.1$} & 0.1 & 0.1 & \dt{$-0.0$} & 20.0 & 20.0 & \dt{$0.0$} & 0.0 & 0.0 & \dt{$0.0$} & 20.0 & 20.0 & \dt{$0.0$} \\
\midrule
\multicolumn{16}{c}{\textit{Closed-Source MLLMs}} \\
\midrule
Gemini-3.1-Pro & 66.0 & 22.0 & \dt{$-44.0$} & 94.0 & 2.0 & \dt{$-92.0$} & 69.0 & 20.0 & \dt{$-49.0$} & 75.0 & 1.0 & \dt{$-74.0$} & 89.0 & 44.0 & \dt{$-45.0$} \\
Gemini-3-Flash & 51.0 & 22.0 & \dt{$-29.0$} & 80.0 & 4.0 & \dt{$-76.0$} & 59.0 & 39.0 & \dt{$-20.0$} & 12.0 & 2.0 & \dt{$-10.0$} & 63.0 & 41.0 & \dt{$-22.0$} \\
Gemini-3-Flash-lite & 29.0 & 22.0 & \dt{$-7.0$} & 8.0 & 6.0 & \dt{$-2.0$} & 26.0 & 26.0 & \dt{$0.0$} & 4.0 & 0.0 & \dt{$-4.0$} & 40.0 & 35.0 & \dt{$-5.0$} \\
Gemini-2.5-Pro & 20.0 & 18.0 & \dt{$-2.0$} & 6.0 & 4.0 & \dt{$-2.0$} & 38.0 & 40.0 & \dt{$2.0$} & 1.0 & 1.0 & \dt{$0.0$} & 45.0 & 34.0 & \dt{$-11.0$} \\
Gemini-2.5-Flash & 23.0 & 16.0 & \dt{$-7.0$} & 1.0 & 2.0 & \dt{$1.0$} & 23.0 & 15.0 & \dt{$-8.0$} & 0.0 & 2.0 & \dt{$2.0$} & 34.0 & 28.0 & \dt{$-6.0$} \\
GPT-5.2 & 46.0 & 21.0 & \dt{$-25.0$} & 88.0 & 7.0 & \dt{$-81.0$} & 35.0 & 29.0 & \dt{$-6.0$} & 56.0 & 7.0 & \dt{$-49.0$} & 50.0 & 41.0 & \dt{$-9.0$} \\
Claude-Sonnet-4.6 & 18.2 & 10.0 & \dt{$-8.2$} & 37.4 & 1.0 & \dt{$-36.4$} & 75.8 & 87.6 & \dt{$11.8$} & 2.0 & 0.0 & \dt{$-2.0$} & 43.3 & 47.4 & \dt{$4.1$} \\
Grok-4-0709 & 20.0 & 20.0 & \dt{$0.0$} & 0.0 & 2.0 & \dt{$2.0$} & 60.0 & 67.0 & \dt{$7.0$} & 0.0 & 0.0 & \dt{$0.0$} & 34.7 & 36.4 & \dt{$1.7$} \\
Grok-4-Fast-Reasoning & 17.0 & 13.0 & \dt{$-4.0$} & 1.0 & 0.0 & \dt{$-1.0$} & 94.0 & 90.0 & \dt{$-4.0$} & 0.0 & 0.0 & \dt{$0.0$} & 38.0 & 43.0 & \dt{$5.0$} \\
\midrule
\multicolumn{16}{c}{\textit{Open-Source MLLMs}} \\
\midrule
Gemma-4-31B & 48.0 & 24.0 & \dt{$-24.0$} & 53.0 & 6.0 & \dt{$-47.0$} & 46.0 & 25.0 & \dt{$-21.0$} & 1.0 & 0.0 & \dt{$-1.0$} & 62.0 & 41.0 & \dt{$-21.0$} \\
Gemma-4-26B & 20.0 & 22.0 & \dt{$2.0$} & 40.0 & 5.0 & \dt{$-35.0$} & 9.1 & 5.0 & \dt{$-4.0$} & 11.0 & 0.0 & \dt{$-11.0$} & 52.0 & 20.2 & \dt{$-31.8$} \\
Kimi-k2.5 & 46.0 & 22.4 & \dt{$-23.6$} & 98.0 & 3.0 & \dt{$-95.0$} & 52.0 & 70.7 & \dt{$18.7$} & 12.0 & 1.0 & \dt{$-11.0$} & 58.6 & 36.0 & \dt{$-22.6$} \\
Qwen3.5-397B-A17B & 52.0 & 20.0 & \dt{$-32.0$} & 72.0 & 7.0 & \dt{$-65.0$} & 47.0 & 32.0 & \dt{$-15.0$} & 27.0 & 3.0 & \dt{$-24.0$} & 75.0 & 32.0 & \dt{$-43.0$} \\
Mistral-Small-2503 & 20.0 & 27.0 & \dt{$7.0$} & 0.0 & 0.0 & \dt{$0.0$} & 54.7 & 74.0 & \dt{$19.3$} & 0.0 & 0.0 & \dt{$0.0$} & 19.0 & 17.0 & \dt{$-2.0$} \\
\bottomrule
\end{tabular}
}
\end{table}

\begin{table}[h]
\centering
\caption{Per-task results under the Spatial Transformation and Geometry category, subcategory Spatial Constraints and Topology. All models are evaluated under high reasoning mode. C and P denote Cartesian and Polar accuracy (\%), respectively. $(-\Delta)$ indicates the accuracy change from Cartesian to Polar. Task abbreviations: Grid Fold.~= Grid Folding; Imp. Shape~= Impossible Shape.}
\label{tab:suppl_stg_spatial}
\resizebox{0.7\linewidth}{!}{
\begin{tabular}{@{} l c c@{\ }l c c@{\ }l c c@{\ }l @{}}
\toprule
\multirow{2}{*}{\textbf{Model}} & \multicolumn{3}{c}{\textbf{Four Color}} & \multicolumn{3}{c}{\textbf{Grid Fold.}} & \multicolumn{3}{c}{\textbf{Imp. Shape}} \\
\cmidrule(lr){2-4} \cmidrule(lr){5-7} \cmidrule(lr){8-10} 
 & \textbf{C} & \textbf{P} & $(-\Delta)$ & \textbf{C} & \textbf{P} & $(-\Delta)$ & \textbf{C} & \textbf{P} & $(-\Delta)$ \\
\midrule
Human & 100.0 & 95.0 & \dt{$-5.0$} & 100.0 & 100.0 & \dt{$0.0$} & 100.0 & 100.0 & \dt{$0.0$} \\
Random & 20.0 & 20.0 & \dt{$0.0$} & 20.0 & 20.0 & \dt{$0.0$} & 20.0 & 20.0 & \dt{$0.0$} \\
\midrule
\multicolumn{10}{c}{\textit{Closed-Source MLLMs}} \\
\midrule
Gemini-3.1-Pro & 91.9 & 28.3 & \dt{$-63.6$} & 100.0 & 54.1 & \dt{$-45.9$} & 62.0 & 20.0 & \dt{$-42.0$} \\
Gemini-3-Flash & 86.0 & 19.0 & \dt{$-67.0$} & 99.0 & 47.0 & \dt{$-52.0$} & 41.0 & 20.0 & \dt{$-21.0$} \\
Gemini-3-Flash-lite & 44.0 & 18.0 & \dt{$-26.0$} & 86.0 & 41.0 & \dt{$-45.0$} & 28.0 & 22.0 & \dt{$-6.0$} \\
Gemini-2.5-Pro & 30.0 & 20.0 & \dt{$-10.0$} & 60.0 & 42.0 & \dt{$-18.0$} & 14.0 & 10.0 & \dt{$-4.0$} \\
Gemini-2.5-Flash & 24.0 & 25.0 & \dt{$1.0$} & 40.0 & 26.0 & \dt{$-14.0$} & 23.0 & 22.0 & \dt{$-1.0$} \\
GPT-5.2 & 98.0 & 27.0 & \dt{$-71.0$} & 100.0 & 59.0 & \dt{$-41.0$} & 43.0 & 22.0 & \dt{$-21.0$} \\
Claude-Sonnet-4.6 & 45.4 & 19.2 & \dt{$-26.2$} & 100.0 & 37.1 & \dt{$-62.9$} & 19.2 & 17.2 & \dt{$-2.0$} \\
Grok-4-0709 & 34.0 & 25.0 & \dt{$-9.0$} & 45.0 & 23.0 & \dt{$-22.0$} & 24.0 & 24.0 & \dt{$0.0$} \\
Grok-4-Fast-Reasoning & 28.0 & 28.0 & \dt{$0.0$} & 51.0 & 18.0 & \dt{$-33.0$} & 21.0 & 15.0 & \dt{$-6.0$} \\
\midrule
\multicolumn{10}{c}{\textit{Open-Source MLLMs}} \\
\midrule
Gemma-4-31B & 65.0 & 32.0 & \dt{$-33.0$} & 97.0 & 44.0 & \dt{$-53.0$} & 33.0 & 22.0 & \dt{$-11.0$} \\
Gemma-4-26B & 51.5 & 19.0 & \dt{$-32.5$} & 98.0 & 36.0 & \dt{$-62.0$} & 25.5 & 13.0 & \dt{$-12.5$} \\
Kimi-k2.5 & 81.8 & 26.0 & \dt{$-55.8$} & 99.0 & 39.0 & \dt{$-60.0$} & 56.0 & 13.0 & \dt{$-43.0$} \\
Qwen3.5-397B-A17B & 92.0 & 43.0 & \dt{$-49.0$} & 100.0 & 64.0 & \dt{$-36.0$} & 82.0 & 26.0 & \dt{$-56.0$} \\
Mistral-Small-2503 & 18.2 & 17.0 & \dt{$-1.2$} & 20.0 & 21.2 & \dt{$1.2$} & 18.6 & 28.3 & \dt{$9.7$} \\
\bottomrule
\end{tabular}
}
\end{table}

\begin{table}[h]
\centering
\caption{Per-task results under the Spatial Transformation and Geometry category, subcategory Rotations and Reflections. All models are evaluated under high reasoning mode. C and P denote Cartesian and Polar accuracy (\%), respectively. $(-\Delta)$ indicates the accuracy change from Cartesian to Polar. Task abbreviations: Grid Rot.~= Grid Rotation; Mirror Ref.~= Mirror Reflection; Pivot Rot.~= Pivot Rotation; Rot. Center~= Rotation Center; Rot. Match.~= Rotation Matching.}
\label{tab:suppl_stg_rotations}
\resizebox{\linewidth}{!}{
\begin{tabular}{@{} l c c@{\ }l c c@{\ }l c c@{\ }l c c@{\ }l c c@{\ }l @{}}
\toprule
\multirow{2}{*}{\textbf{Model}} & \multicolumn{3}{c}{\textbf{Grid Rot.}} & \multicolumn{3}{c}{\textbf{Mirror Ref.}} & \multicolumn{3}{c}{\textbf{Pivot Rot.}} & \multicolumn{3}{c}{\textbf{Rot. Center}} & \multicolumn{3}{c}{\textbf{Rot. Match.}} \\
\cmidrule(lr){2-4} \cmidrule(lr){5-7} \cmidrule(lr){8-10} \cmidrule(lr){11-13} \cmidrule(lr){14-16} 
 & \textbf{C} & \textbf{P} & $(-\Delta)$ & \textbf{C} & \textbf{P} & $(-\Delta)$ & \textbf{C} & \textbf{P} & $(-\Delta)$ & \textbf{C} & \textbf{P} & $(-\Delta)$ & \textbf{C} & \textbf{P} & $(-\Delta)$ \\
\midrule
Human & 100.0 & 100.0 & \dt{$0.0$} & 100.0 & 100.0 & \dt{$0.0$} & 100.0 & 100.0 & \dt{$0.0$} & 95.0 & 90.0 & \dt{$-5.0$} & 100.0 & 100.0 & \dt{$0.0$} \\
Random & 20.0 & 20.0 & \dt{$0.0$} & 25.0 & 25.0 & \dt{$0.0$} & 20.0 & 20.0 & \dt{$0.0$} & 25.0 & 25.0 & \dt{$0.0$} & 0.0 & 0.0 & \dt{$0.0$} \\
\midrule
\multicolumn{16}{c}{\textit{Closed-Source MLLMs}} \\
\midrule
Gemini-3.1-Pro & 98.0 & 31.0 & \dt{$-67.0$} & 84.0 & 42.0 & \dt{$-42.0$} & 87.0 & 51.5 & \dt{$-35.5$} & 100.0 & 40.8 & \dt{$-59.2$} & 72.0 & 51.5 & \dt{$-20.5$} \\
Gemini-3-Flash & 99.0 & 35.0 & \dt{$-64.0$} & 76.0 & 47.0 & \dt{$-29.0$} & 83.0 & 62.0 & \dt{$-21.0$} & 96.0 & 20.0 & \dt{$-76.0$} & 48.0 & 30.0 & \dt{$-18.0$} \\
Gemini-3-Flash-lite & 73.0 & 23.0 & \dt{$-50.0$} & 33.0 & 36.0 & \dt{$3.0$} & 52.0 & 36.0 & \dt{$-16.0$} & 78.0 & 11.0 & \dt{$-67.0$} & 27.0 & 6.0 & \dt{$-21.0$} \\
Gemini-2.5-Pro & 30.0 & 20.0 & \dt{$-10.0$} & 38.0 & 35.0 & \dt{$-3.0$} & 48.0 & 33.0 & \dt{$-15.0$} & 75.0 & 34.0 & \dt{$-41.0$} & 34.0 & 20.0 & \dt{$-14.0$} \\
Gemini-2.5-Flash & 27.0 & 23.0 & \dt{$-4.0$} & 32.0 & 27.0 & \dt{$-5.0$} & 32.0 & 22.0 & \dt{$-10.0$} & 66.0 & 27.0 & \dt{$-39.0$} & 18.0 & 5.0 & \dt{$-13.0$} \\
GPT-5.2 & 100.0 & 87.0 & \dt{$-13.0$} & 98.0 & 58.0 & \dt{$-40.0$} & 94.0 & 68.0 & \dt{$-26.0$} & 96.0 & 28.0 & \dt{$-68.0$} & 91.0 & 87.0 & \dt{$-4.0$} \\
Claude-Sonnet-4.6 & 53.1 & 19.2 & \dt{$-33.9$} & 39.0 & 43.0 & \dt{$4.0$} & 40.0 & 32.0 & \dt{$-8.0$} & 68.7 & 13.3 & \dt{$-55.4$} & 7.2 & 1.0 & \dt{$-6.2$} \\
Grok-4-0709 & 34.0 & 21.0 & \dt{$-13.0$} & 35.0 & 32.0 & \dt{$-3.0$} & 33.0 & 21.2 & \dt{$-11.8$} & 59.6 & 15.0 & \dt{$-44.6$} & 9.0 & 2.0 & \dt{$-7.0$} \\
Grok-4-Fast-Reasoning & 21.0 & 18.0 & \dt{$-3.0$} & 25.0 & 25.0 & \dt{$0.0$} & 28.0 & 13.0 & \dt{$-15.0$} & 41.0 & 15.0 & \dt{$-26.0$} & 4.0 & 2.0 & \dt{$-2.0$} \\
\midrule
\multicolumn{16}{c}{\textit{Open-Source MLLMs}} \\
\midrule
Gemma-4-31B & 97.0 & 28.0 & \dt{$-69.0$} & 64.0 & 35.0 & \dt{$-29.0$} & 76.0 & 44.0 & \dt{$-32.0$} & 82.0 & 26.0 & \dt{$-56.0$} & 57.0 & 43.0 & \dt{$-14.0$} \\
Gemma-4-26B & 92.0 & 25.2 & \dt{$-66.8$} & 62.0 & 40.0 & \dt{$-22.0$} & 57.0 & 42.0 & \dt{$-15.0$} & 70.0 & 6.1 & \dt{$-63.9$} & 43.0 & 18.0 & \dt{$-25.0$} \\
Kimi-k2.5 & 100.0 & 29.0 & \dt{$-71.0$} & 91.9 & 53.5 & \dt{$-38.4$} & 78.0 & 52.5 & \dt{$-25.5$} & 90.7 & 16.7 & \dt{$-74.0$} & 68.0 & 42.0 & \dt{$-26.0$} \\
Qwen3.5-397B-A17B & 100.0 & 73.0 & \dt{$-27.0$} & 94.0 & 56.0 & \dt{$-38.0$} & 89.0 & 60.0 & \dt{$-29.0$} & 93.0 & 35.0 & \dt{$-58.0$} & 81.0 & 62.0 & \dt{$-19.0$} \\
Mistral-Small-2503 & 23.0 & 15.3 & \dt{$-7.7$} & 27.0 & 34.0 & \dt{$7.0$} & 13.0 & 20.0 & \dt{$7.0$} & 37.0 & 22.0 & \dt{$-15.0$} & 3.0 & 0.0 & \dt{$-3.0$} \\
\bottomrule
\end{tabular}
}
\end{table}

\begin{table}[h]
\centering
\caption{Per-task results under the Algorithmic Logic and Simulation category. All models are evaluated under high reasoning mode. C and P denote Cartesian and Polar accuracy (\%), respectively. $(-\Delta)$ indicates the accuracy change from Cartesian to Polar. Task abbreviations: Bounce~= Bouncing Point; Collision~= Collision Detection; Max. Coll.~= Maximum Collection; Min. Flips~= Minimum Flips.}
\label{tab:suppl_algo_all}
\resizebox{\linewidth}{!}{
\begin{tabular}{@{} l c c@{\ }l c c@{\ }l c c@{\ }l c c@{\ }l c c@{\ }l c c@{\ }l @{}}
\toprule
\multirow{2}{*}{\textbf{Model}} & \multicolumn{3}{c}{\textbf{Bounce}} & \multicolumn{3}{c}{\textbf{Collision}} & \multicolumn{3}{c}{\textbf{Max. Coll.}} & \multicolumn{3}{c}{\textbf{Min. Flips}} & \multicolumn{3}{c}{\textbf{N-Queens}} & \multicolumn{3}{c}{\textbf{Sudoku}} \\
\cmidrule(lr){2-4} \cmidrule(lr){5-7} \cmidrule(lr){8-10} \cmidrule(lr){11-13} \cmidrule(lr){14-16} \cmidrule(lr){17-19} 
 & \textbf{C} & \textbf{P} & $(-\Delta)$ & \textbf{C} & \textbf{P} & $(-\Delta)$ & \textbf{C} & \textbf{P} & $(-\Delta)$ & \textbf{C} & \textbf{P} & $(-\Delta)$ & \textbf{C} & \textbf{P} & $(-\Delta)$ & \textbf{C} & \textbf{P} & $(-\Delta)$ \\
\midrule
Human & 85.0 & 70.0 & \dt{$-15.0$} & 95.0 & 95.0 & \dt{$0.0$} & 95.0 & 95.0 & \dt{$0.0$} & 85.0 & 80.0 & \dt{$-5.0$} & 95.0 & 95.0 & \dt{$0.0$} & 100.0 & 90.0 & \dt{$-10.0$} \\
Random & 1.8 & 1.8 & \dt{$-0.0$} & 20.0 & 20.0 & \dt{$0.0$} & 20.0 & 20.0 & \dt{$0.0$} & 20.0 & 20.0 & \dt{$0.0$} & 20.0 & 20.0 & \dt{$0.0$} & 20.0 & 20.0 & \dt{$0.0$} \\
\midrule
\multicolumn{19}{c}{\textit{Closed-Source MLLMs}} \\
\midrule
Gemini-3.1-Pro & 97.0 & 30.6 & \dt{$-66.4$} & 58.0 & 34.0 & \dt{$-24.0$} & 66.0 & 31.0 & \dt{$-35.0$} & 88.9 & 85.7 & \dt{$-3.2$} & 100.0 & 74.0 & \dt{$-26.0$} & 100.0 & 60.0 & \dt{$-40.0$} \\
Gemini-3-Flash & 49.0 & 28.0 & \dt{$-21.0$} & 67.0 & 31.0 & \dt{$-36.0$} & 30.0 & 20.0 & \dt{$-10.0$} & 68.0 & 53.0 & \dt{$-15.0$} & 100.0 & 50.0 & \dt{$-50.0$} & 99.0 & 67.0 & \dt{$-32.0$} \\
Gemini-3-Flash-lite & 34.0 & 16.0 & \dt{$-18.0$} & 54.0 & 27.0 & \dt{$-27.0$} & 27.0 & 18.0 & \dt{$-9.0$} & 37.0 & 32.0 & \dt{$-5.0$} & 100.0 & 33.0 & \dt{$-67.0$} & 96.0 & 44.0 & \dt{$-52.0$} \\
Gemini-2.5-Pro & 32.0 & 25.0 & \dt{$-7.0$} & 48.0 & 29.0 & \dt{$-19.0$} & 18.0 & 17.0 & \dt{$-1.0$} & 36.0 & 24.0 & \dt{$-12.0$} & 99.0 & 34.0 & \dt{$-65.0$} & 93.0 & 52.0 & \dt{$-41.0$} \\
Gemini-2.5-Flash & 19.0 & 24.0 & \dt{$5.0$} & 46.0 & 25.0 & \dt{$-21.0$} & 20.0 & 14.0 & \dt{$-6.0$} & 31.0 & 31.0 & \dt{$0.0$} & 99.0 & 35.0 & \dt{$-64.0$} & 93.0 & 37.0 & \dt{$-56.0$} \\
GPT-5.2 & 79.0 & 32.0 & \dt{$-47.0$} & 61.0 & 40.0 & \dt{$-21.0$} & 43.0 & 17.0 & \dt{$-26.0$} & 41.0 & 30.0 & \dt{$-11.0$} & 100.0 & 90.0 & \dt{$-10.0$} & 100.0 & 71.0 & \dt{$-29.0$} \\
Claude-Sonnet-4.6 & 86.3 & 9.6 & \dt{$-76.8$} & 51.0 & 32.3 & \dt{$-18.7$} & 20.6 & 33.3 & \dt{$12.7$} & 18.9 & 22.7 & \dt{$3.7$} & 99.0 & 34.7 & \dt{$-64.2$} & 83.8 & 49.0 & \dt{$-34.8$} \\
Grok-4-0709 & 21.9 & 17.2 & \dt{$-4.7$} & 38.4 & 29.0 & \dt{$-9.4$} & 26.0 & 29.3 & \dt{$3.3$} & 55.7 & 56.1 & \dt{$0.5$} & 96.0 & 29.0 & \dt{$-67.0$} & 81.0 & 36.0 & \dt{$-45.0$} \\
Grok-4-Fast-Reasoning & 17.0 & 13.0 & \dt{$-4.0$} & 30.0 & 25.0 & \dt{$-5.0$} & 34.0 & 30.0 & \dt{$-4.0$} & 57.0 & 55.0 & \dt{$-2.0$} & 76.0 & 26.0 & \dt{$-50.0$} & 84.0 & 36.0 & \dt{$-48.0$} \\
\midrule
\multicolumn{19}{c}{\textit{Open-Source MLLMs}} \\
\midrule
Gemma-4-31B & 48.0 & 21.0 & \dt{$-27.0$} & 53.0 & 26.0 & \dt{$-27.0$} & 52.0 & 27.0 & \dt{$-25.0$} & 55.0 & 36.0 & \dt{$-19.0$} & 100.0 & 56.0 & \dt{$-44.0$} & 98.0 & 50.0 & \dt{$-48.0$} \\
Gemma-4-26B & 42.0 & 23.0 & \dt{$-19.0$} & 39.0 & 14.0 & \dt{$-25.0$} & 14.1 & 8.0 & \dt{$-6.1$} & 43.0 & 36.0 & \dt{$-7.0$} & 100.0 & 46.0 & \dt{$-54.0$} & 98.0 & 50.0 & \dt{$-48.0$} \\
Kimi-k2.5 & 90.6 & 17.8 & \dt{$-72.8$} & 56.0 & 32.0 & \dt{$-24.0$} & 36.8 & 22.3 & \dt{$-14.5$} & 59.4 & 45.2 & \dt{$-14.2$} & 99.0 & 53.7 & \dt{$-45.3$} & 99.0 & 53.0 & \dt{$-46.0$} \\
Qwen3.5-397B-A17B & 58.0 & 25.0 & \dt{$-33.0$} & 58.0 & 37.0 & \dt{$-21.0$} & 70.0 & 29.0 & \dt{$-41.0$} & 56.0 & 51.0 & \dt{$-5.0$} & 100.0 & 46.0 & \dt{$-54.0$} & 91.0 & 71.0 & \dt{$-20.0$} \\
Mistral-Small-2503 & 4.5 & 0.0 & \dt{$-4.5$} & 31.0 & 35.7 & \dt{$4.7$} & 22.0 & 18.0 & \dt{$-4.0$} & 15.8 & 24.2 & \dt{$8.4$} & 38.0 & 20.0 & \dt{$-18.0$} & 39.0 & 26.0 & \dt{$-13.0$} \\
\bottomrule
\end{tabular}
}
\end{table}

\begin{table}[h]
\centering
\caption{Per-task results under the Combinatorics and Probability category, subcategory Bounded Combinatorics. All models are evaluated under high reasoning mode. C and P denote Cartesian and Polar accuracy (\%), respectively. $(-\Delta)$ indicates the accuracy change from Cartesian to Polar. Task abbreviations: Bnd. Diag.~= Bounded Diagonal Paths; Bnd. Knight~= Bounded Knight Paths; Chkpt. Path~= Checkpoint Paths; Lattice~= Lattice Paths; Path Cnt.~= Path Counting.}
\label{tab:suppl_cp_bounded}
\resizebox{\linewidth}{!}{
\begin{tabular}{@{} l c c@{\ }l c c@{\ }l c c@{\ }l c c@{\ }l c c@{\ }l @{}}
\toprule
\multirow{2}{*}{\textbf{Model}} & \multicolumn{3}{c}{\textbf{Bnd. Diag.}} & \multicolumn{3}{c}{\textbf{Bnd. Knight}} & \multicolumn{3}{c}{\textbf{Chkpt. Path}} & \multicolumn{3}{c}{\textbf{Lattice}} & \multicolumn{3}{c}{\textbf{Path Cnt.}} \\
\cmidrule(lr){2-4} \cmidrule(lr){5-7} \cmidrule(lr){8-10} \cmidrule(lr){11-13} \cmidrule(lr){14-16} 
 & \textbf{C} & \textbf{P} & $(-\Delta)$ & \textbf{C} & \textbf{P} & $(-\Delta)$ & \textbf{C} & \textbf{P} & $(-\Delta)$ & \textbf{C} & \textbf{P} & $(-\Delta)$ & \textbf{C} & \textbf{P} & $(-\Delta)$ \\
\midrule
Human & 65.0 & 50.0 & \dt{$-15.0$} & 75.0 & 45.0 & \dt{$-30.0$} & 75.0 & 45.0 & \dt{$-30.0$} & 95.0 & 90.0 & \dt{$-5.0$} & 85.0 & 70.0 & \dt{$-15.0$} \\
Random & 0.0 & 0.0 & \dt{$0.0$} & 0.0 & 0.0 & \dt{$0.0$} & 0.0 & 0.0 & \dt{$0.0$} & 0.0 & 0.0 & \dt{$0.0$} & 20.0 & 20.0 & \dt{$0.0$} \\
\midrule
\multicolumn{16}{c}{\textit{Closed-Source MLLMs}} \\
\midrule
Gemini-3.1-Pro & 100.0 & 64.3 & \dt{$-35.7$} & 92.0 & 4.0 & \dt{$-88.0$} & 99.0 & 10.0 & \dt{$-89.0$} & 88.0 & 19.0 & \dt{$-69.0$} & 82.0 & 23.0 & \dt{$-59.0$} \\
Gemini-3-Flash & 97.0 & 39.0 & \dt{$-58.0$} & 76.0 & 6.0 & \dt{$-70.0$} & 81.0 & 5.0 & \dt{$-76.0$} & 76.0 & 13.0 & \dt{$-63.0$} & 59.0 & 34.0 & \dt{$-25.0$} \\
Gemini-3-Flash-lite & 53.0 & 42.0 & \dt{$-11.0$} & 23.0 & 1.0 & \dt{$-22.0$} & 55.0 & 1.0 & \dt{$-54.0$} & 40.0 & 4.0 & \dt{$-36.0$} & 38.0 & 27.0 & \dt{$-11.0$} \\
Gemini-2.5-Pro & 29.0 & 23.0 & \dt{$-6.0$} & 4.0 & 0.0 & \dt{$-4.0$} & 25.0 & 1.0 & \dt{$-24.0$} & 9.0 & 2.0 & \dt{$-7.0$} & 32.0 & 28.0 & \dt{$-4.0$} \\
Gemini-2.5-Flash & 12.0 & 11.0 & \dt{$-1.0$} & 1.0 & 1.0 & \dt{$0.0$} & 9.0 & 0.0 & \dt{$-9.0$} & 6.0 & 1.0 & \dt{$-5.0$} & 20.0 & 18.0 & \dt{$-2.0$} \\
GPT-5.2 & 98.0 & 78.0 & \dt{$-20.0$} & 88.0 & 33.0 & \dt{$-55.0$} & 71.0 & 0.0 & \dt{$-71.0$} & 83.0 & 9.0 & \dt{$-74.0$} & 85.0 & 22.0 & \dt{$-63.0$} \\
Claude-Sonnet-4.6 & 74.7 & 38.0 & \dt{$-36.7$} & 66.7 & 9.5 & \dt{$-57.2$} & 62.5 & 4.3 & \dt{$-58.2$} & 29.4 & 3.1 & \dt{$-26.2$} & 27.6 & 32.0 & \dt{$4.5$} \\
Grok-4-0709 & 17.2 & 13.0 & \dt{$-4.1$} & 4.5 & 1.0 & \dt{$-3.5$} & 7.3 & 0.0 & \dt{$-7.3$} & 10.0 & 2.0 & \dt{$-8.0$} & 28.0 & 25.0 & \dt{$-3.0$} \\
Grok-4-Fast-Reasoning & 27.0 & 13.0 & \dt{$-14.0$} & 9.0 & 0.0 & \dt{$-9.0$} & 5.0 & 0.0 & \dt{$-5.0$} & 2.0 & 2.0 & \dt{$0.0$} & 31.0 & 32.0 & \dt{$1.0$} \\
\midrule
\multicolumn{16}{c}{\textit{Open-Source MLLMs}} \\
\midrule
Gemma-4-31B & 91.0 & 52.0 & \dt{$-39.0$} & 56.0 & 8.0 & \dt{$-48.0$} & 48.0 & 7.0 & \dt{$-41.0$} & 75.0 & 7.0 & \dt{$-68.0$} & 41.0 & 22.0 & \dt{$-19.0$} \\
Gemma-4-26B & 80.0 & 50.0 & \dt{$-30.0$} & 51.0 & 4.1 & \dt{$-46.9$} & 22.0 & 5.0 & \dt{$-17.0$} & 48.0 & 9.0 & \dt{$-39.0$} & 10.2 & 11.2 & \dt{$1.0$} \\
Kimi-k2.5 & 96.9 & 16.7 & \dt{$-80.2$} & 79.4 & 4.1 & \dt{$-75.3$} & 51.6 & 1.1 & \dt{$-50.6$} & 52.0 & 5.0 & \dt{$-47.0$} & 49.0 & 27.6 & \dt{$-21.4$} \\
Qwen3.5-397B-A17B & 99.0 & 35.0 & \dt{$-64.0$} & 41.0 & 9.0 & \dt{$-32.0$} & 81.0 & 4.0 & \dt{$-77.0$} & 77.0 & 6.0 & \dt{$-71.0$} & 8.0 & 4.0 & \dt{$-4.0$} \\
Mistral-Small-2503 & 0.0 & 0.0 & \dt{$0.0$} & 2.1 & 1.0 & \dt{$-1.1$} & 3.5 & 0.0 & \dt{$-3.5$} & 0.0 & 1.0 & \dt{$1.0$} & 23.0 & 27.0 & \dt{$4.0$} \\
\bottomrule
\end{tabular}
}
\end{table}

\begin{table}[h]
\centering
\caption{Per-task results under the Combinatorics and Probability category, subcategories Stochastic Processes and Topological Combinatorics. All models are evaluated under high reasoning mode. C and P denote Cartesian and Polar accuracy (\%), respectively. $(-\Delta)$ indicates the accuracy change from Cartesian to Polar. Task abbreviations: Edge Cnt.~= Edge Counting; Rand. Walk~= Random Walk; Knight Path~= Knight Paths; Wrap. Diag.~= Wrapping Diagonal Paths.}
\label{tab:suppl_cp_stoch_topo}
\resizebox{0.85\linewidth}{!}{
\begin{tabular}{@{} l c c@{\ }l c c@{\ }l c c@{\ }l c c@{\ }l @{}}
\toprule
\multirow{2}{*}{\textbf{Model}} & \multicolumn{3}{c}{\textbf{Edge Cnt.}} & \multicolumn{3}{c}{\textbf{Rand. Walk}} & \multicolumn{3}{c}{\textbf{Knight Path}} & \multicolumn{3}{c}{\textbf{Wrap. Diag.}} \\
\cmidrule(lr){2-4} \cmidrule(lr){5-7} \cmidrule(lr){8-10} \cmidrule(lr){11-13} 
 & \textbf{C} & \textbf{P} & $(-\Delta)$ & \textbf{C} & \textbf{P} & $(-\Delta)$ & \textbf{C} & \textbf{P} & $(-\Delta)$ & \textbf{C} & \textbf{P} & $(-\Delta)$ \\
\midrule
Human & 100.0 & 95.0 & \dt{$-5.0$} & 70.0 & 80.0 & \dt{$10.0$} & 70.0 & 45.0 & \dt{$-25.0$} & 65.0 & 60.0 & \dt{$-5.0$} \\
Random & 20.0 & 20.0 & \dt{$0.0$} & 20.0 & 20.0 & \dt{$0.0$} & 0.0 & 0.0 & \dt{$0.0$} & 0.0 & 0.0 & \dt{$0.0$} \\
\midrule
\multicolumn{13}{c}{\textit{Closed-Source MLLMs}} \\
\midrule
Gemini-3.1-Pro & 91.0 & 25.0 & \dt{$-66.0$} & 100.0 & 40.8 & \dt{$-59.2$} & 90.0 & 2.0 & \dt{$-88.0$} & 90.0 & 35.7 & \dt{$-54.3$} \\
Gemini-3-Flash & 65.0 & 34.0 & \dt{$-31.0$} & 93.0 & 26.0 & \dt{$-67.0$} & 74.0 & 5.0 & \dt{$-69.0$} & 52.0 & 23.0 & \dt{$-29.0$} \\
Gemini-3-Flash-lite & 28.0 & 29.0 & \dt{$1.0$} & 78.0 & 23.0 & \dt{$-55.0$} & 28.0 & 1.0 & \dt{$-27.0$} & 21.0 & 15.0 & \dt{$-6.0$} \\
Gemini-2.5-Pro & 29.0 & 27.0 & \dt{$-2.0$} & 66.0 & 24.0 & \dt{$-42.0$} & 3.0 & 1.0 & \dt{$-2.0$} & 13.0 & 8.0 & \dt{$-5.0$} \\
Gemini-2.5-Flash & 21.0 & 26.0 & \dt{$5.0$} & 68.0 & 25.0 & \dt{$-43.0$} & 2.0 & 2.0 & \dt{$0.0$} & 4.0 & 1.0 & \dt{$-3.0$} \\
GPT-5.2 & 67.0 & 36.0 & \dt{$-31.0$} & 99.0 & 44.0 & \dt{$-55.0$} & 88.0 & 28.0 & \dt{$-60.0$} & 84.0 & 26.0 & \dt{$-58.0$} \\
Claude-Sonnet-4.6 & 40.4 & 39.4 & \dt{$-1.0$} & 42.5 & 23.2 & \dt{$-19.4$} & 18.8 & 0.0 & \dt{$-18.8$} & 22.1 & 2.1 & \dt{$-20.0$} \\
Grok-4-0709 & 39.0 & 40.0 & \dt{$1.0$} & 62.2 & 17.2 & \dt{$-45.0$} & 5.2 & 1.0 & \dt{$-4.2$} & 14.0 & 9.0 & \dt{$-5.0$} \\
Grok-4-Fast-Reasoning & 41.0 & 36.0 & \dt{$-5.0$} & 26.0 & 20.0 & \dt{$-6.0$} & 7.0 & 0.0 & \dt{$-7.0$} & 11.0 & 8.0 & \dt{$-3.0$} \\
\midrule
\multicolumn{13}{c}{\textit{Open-Source MLLMs}} \\
\midrule
Gemma-4-31B & 41.4 & 18.0 & \dt{$-23.4$} & 89.0 & 45.0 & \dt{$-44.0$} & 66.0 & 7.0 & \dt{$-59.0$} & 53.0 & 9.0 & \dt{$-44.0$} \\
Gemma-4-26B & 22.2 & 14.0 & \dt{$-8.2$} & 83.0 & 23.0 & \dt{$-60.0$} & 62.0 & 4.0 & \dt{$-58.0$} & 33.3 & 13.0 & \dt{$-20.3$} \\
Kimi-k2.5 & 41.0 & 34.0 & \dt{$-7.0$} & 92.6 & 43.3 & \dt{$-49.3$} & 85.6 & 7.1 & \dt{$-78.5$} & 70.0 & 7.0 & \dt{$-63.0$} \\
Qwen3.5-397B-A17B & 68.0 & 31.0 & \dt{$-37.0$} & 96.0 & 49.0 & \dt{$-47.0$} & 53.0 & 6.0 & \dt{$-47.0$} & 64.0 & 19.0 & \dt{$-45.0$} \\
Mistral-Small-2503 & 38.5 & 41.2 & \dt{$2.7$} & 18.0 & 26.0 & \dt{$8.0$} & 0.0 & 2.0 & \dt{$2.0$} & 1.0 & 0.0 & \dt{$-1.0$} \\
\bottomrule
\end{tabular}
}
\end{table}

We report detailed per-task performance for all five task categories, with results organized by subcategory where applicable.
\paragraph{Navigation and Routing.}
Tables~\ref{tab:suppl_nav_rule_nav}, \ref{tab:suppl_nav_optimal_path}, and~\ref{tab:suppl_nav_constraint_route} present the per-task results under this category, grouped by the Rule-Based Navigation, Optimal Pathfinding, and Constraint-Based Routing subcategories, respectively.
\paragraph{Visual Pattern Matching.}
Tables~\ref{tab:suppl_vpm_spotting_pattern} and~\ref{tab:suppl_vpm_shape_jigsaw} present the per-task results, with the latter merging subcategories that contain fewer tasks into a single table.
\paragraph{Spatial Transformation and Geometry.}
Tables~\ref{tab:suppl_stg_geometric}, \ref{tab:suppl_stg_spatial}, and~\ref{tab:suppl_stg_rotations} present the per-task results, grouped by the Geometric Measurement and Counting, Spatial Constraints and Topology, and Rotations and Reflections subcategories, respectively.
\paragraph{Algorithmic Logic and Simulation.}
Table~\ref{tab:suppl_algo_all} presents the per-task results for this category. Due to the small number of tasks per subcategory, all results are consolidated into a single table.
\paragraph{Combinatorics and Probability.}
Tables~\ref{tab:suppl_cp_bounded} and~\ref{tab:suppl_cp_stoch_topo} present the per-task results, with Table~\ref{tab:suppl_cp_bounded} covering the Bounded Combinatorics subcategory and Table~\ref{tab:suppl_cp_stoch_topo} merging the Stochastic Processes and Topological Combinatorics subcategories.

Notably, while the overall Cartesian-to-Polar accuracy gap remains substantial across all models, each model exhibits a small subset of tasks on which its performance drop is comparatively moderate relative to other models evaluated on the same task. 
These task-level exceptions are largely \emph{model-specific}: the particular tasks on which a given model shows a smaller gap differ considerably from one model to another, with little consistent overlap across the evaluated model set.
We think this may stem from idiosyncratic biases in individual 
models' training distributions.
Moreover, the rare cases in which Polar accuracy approaches Cartesian accuracy almost exclusively occur when both values are near the random baseline, 
indicating that the model is effectively guessing under both layouts rather than maintaining high accuracy across the layout change.

\subsection{Experimental Details for Discussion}
\label{app:discussion_details}

Below we provide more details of the experiments discussed in Section~\ref{sec:discussion}.

\paragraph{Prompting interventions.}
We randomly select six tasks from those exhibiting large Cartesian-to-Polar accuracy gaps: Bouncing Point, Edge Counting, Four Color, Grid Rotation, Pattern Prediction, and Rule-Based Navigation. For each task we generate four variation runs, each containing 100 Polar instances (400 instances per task in total). We evaluate six representative models spanning closed-source and open-source families: Gemini-3.1-Pro, Gemini-3-Flash, GPT-5.2, Claude-Sonnet-4.6, Kimi-k2.5, and Gemma-4-31B. Each model is evaluated under three conditions: (1)~vanilla Polar (the standard Polar prompt), (2)~Polar + Conversion Hint, where the prompt is augmented with an explicit instruction asking the model to first re-map the Polar layout into an equivalent Cartesian grid before solving the task (Figure~\ref{fig:prompt_hint}), and (3)~Polar + 5-shot ICL, where the prompt is prepended with five solved Polar examples from the same task (Figure~\ref{fig:prompt_fewshot_cot}). 
The Cartesian baseline uses the same 100 instances per run under the standard prompt. Table~\ref{tab:ood_interventions} reports each model's accuracy averaged across the six tasks and four runs. Figure~\ref{fig:discussion}(f) visualizes the same data. The six selected tasks span all five task categories: Visual Pattern Matching (Pattern Prediction), Spatial Transformation \& Geometry (Four Color, Grid Rotation), Navigation \& Routing (Rule-Based Navigation), Combinatorics \& Probability (Edge Counting), and Algorithmic Logic \& Simulation (Bouncing Point).

\paragraph{Geometry deformation.}
To test whether the performance gap is specific to the Polar transformation, we construct a gradient of intermediate geometric deformations applied to Cartesian images. We randomly select three tasks---Pattern Prediction, Pipe Lengths, and Path Counting---and apply five deformation levels of increasing severity: (1)~\textbf{Wave}: a sinusoidal distortion applied to the vertical axis of the Cartesian grid, which introduces curvature without radial compression or angular wrapping; and (2)~\textbf{Arc-$\boldsymbol{\theta}$} ($\theta \in \{45^\circ, 90^\circ, 135^\circ, 180^\circ\}$): the Cartesian grid is bent into a circular arc of the specified angular extent, progressively departing from orthogonal structure. All deformations are generated programmatically from the same Cartesian source images while ensuring that task-relevant visual elements (labels, grid lines, paths) remain legible. The same six models are evaluated on each deformation level, as well as on the original Cartesian and full Polar versions for reference. Table~\ref{tab:geometry_deformation_full} reports accuracy averaged over the three tasks. Figure~\ref{fig:discussion}(e) visualizes the per-deformation-level accuracy for each model.

\begin{table}[t]
\centering
\caption{Prompting intervention accuracy (\%) across 6 representative models on 6 randomly selected tasks (4 runs $\times$ 100 instances per task). Neither conversion hints nor 5-shot ICL recovers Polar accuracy.}
\label{tab:ood_interventions}
\resizebox{0.8\linewidth}{!}{
\begin{tabular}{lcccc}
\toprule
\textbf{Model} & \textbf{Cartesian} & \textbf{Polar} & \textbf{Polar+Conversion Hint} & \textbf{Polar+5-shot ICL} \\
\midrule
Gemini-3.1-Pro & 94.2 & 28.0 & 27.8 & 24.7 \\
Gemini-3-Flash & 78.3 & 30.5 & 28.2 & 28.1 \\
GPT-5.2 & 82.3 & 38.3 & 40.0 & 26.0 \\
Claude-Sonnet-4.6 & 45.2 & 22.0 & 22.3 & 24.6 \\
Kimi-k2.5 & 69.2 & 25.1 & 21.2 & 19.1 \\
Gemma-4-31B & 61.2 & 27.8 & 24.3 & 22.6 \\
\bottomrule
\end{tabular}
}
\end{table}

\begin{table}[h]
\centering
\caption{Complete results (accuracy \%) under geometry deformation across 6 evaluated models (averaged over 3 tasks: Pattern Prediction, Pipe Lengths, and Path Counting).}
\label{tab:geometry_deformation_full}
\resizebox{0.85\linewidth}{!}{
\begin{tabular}{lccccccc}
\toprule
\textbf{Model} & \textbf{Cartesian} & \textbf{Wave} & \textbf{Arc-$45^\circ$} & \textbf{Arc-$90^\circ$} & \textbf{Arc-$135^\circ$} & \textbf{Arc-$180^\circ$} & \textbf{Polar} \\
\midrule
Gemini-3.1-Pro & 89.0 & 44.0 & 35.0 & 20.0 & 15.3 & 16.3 & 17.1 \\
Gemini-3-Flash & 51.3 & 23.9 & 30.7 & 21.1 & 15.8 & 16.5 & 23.3 \\
GPT-5.2 & 65.0 & 29.0 & 30.3 & 24.8 & 21.0 & 20.3 & 16.7 \\
Claude-Sonnet-4.6 & 18.7 & 12.3 & 16.3 & 13.7 & 14.7 & 12.0 & 15.7 \\
Gemma-4-31B & 27.7 & 17.7 & 22.3 & 15.9 & 19.2 & 16.3 & 18.0 \\
Kimi-k2.5 & 26.7 & 16.0 & 19.0 & 18.3 & 14.7 & 14.0 & 15.3 \\
\bottomrule
\end{tabular}
}
\end{table}

\section{Task Examples}

We provide examples with complete image and question, answer from more tasks in {\polar}.

\begin{figure}
    \centering
    \includegraphics[width=\linewidth]{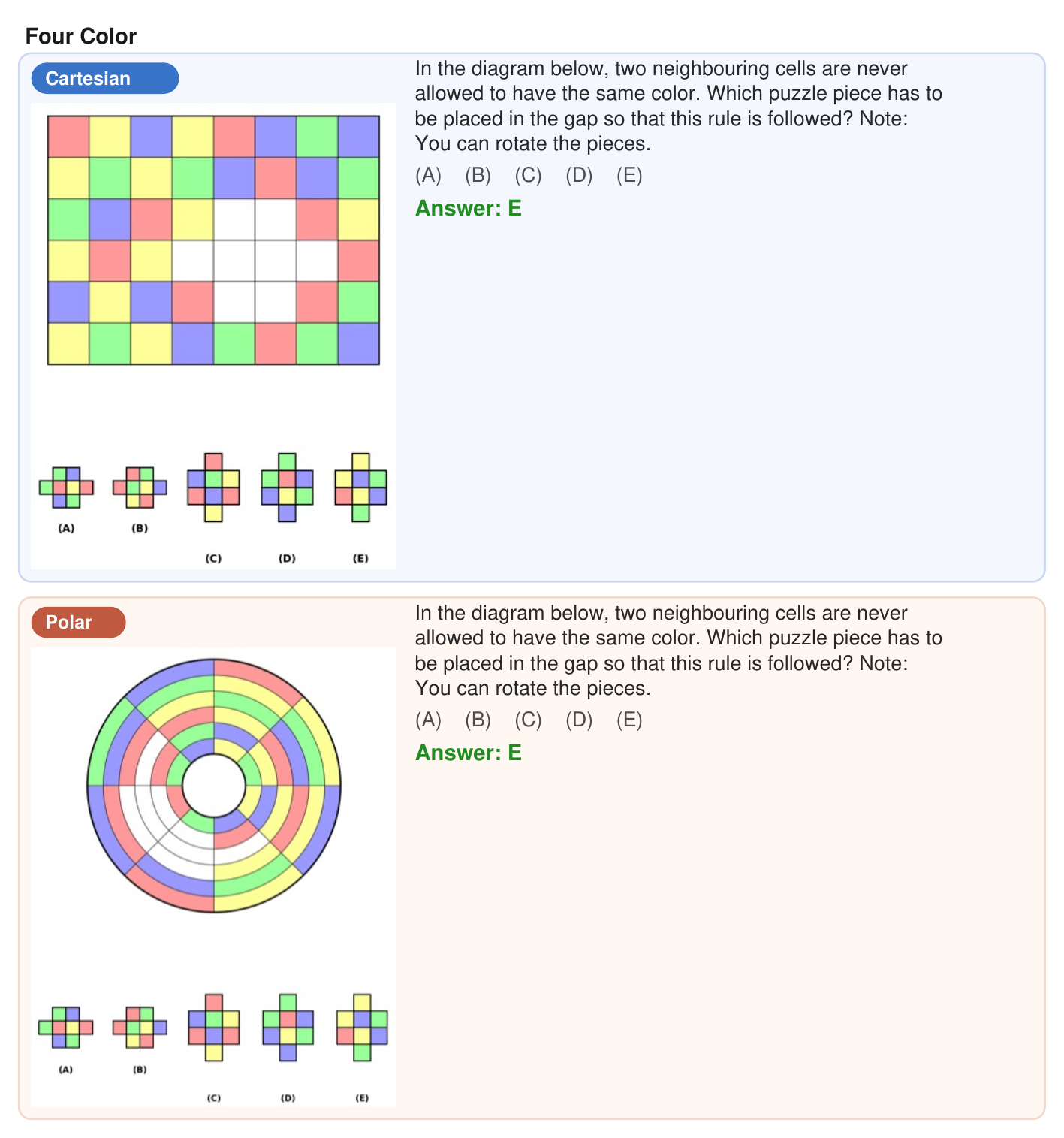}
    \caption{Example from four color task.}
    \label{fig:placeholder}
\end{figure}

\begin{figure}
    \centering
    \includegraphics[width=\linewidth]{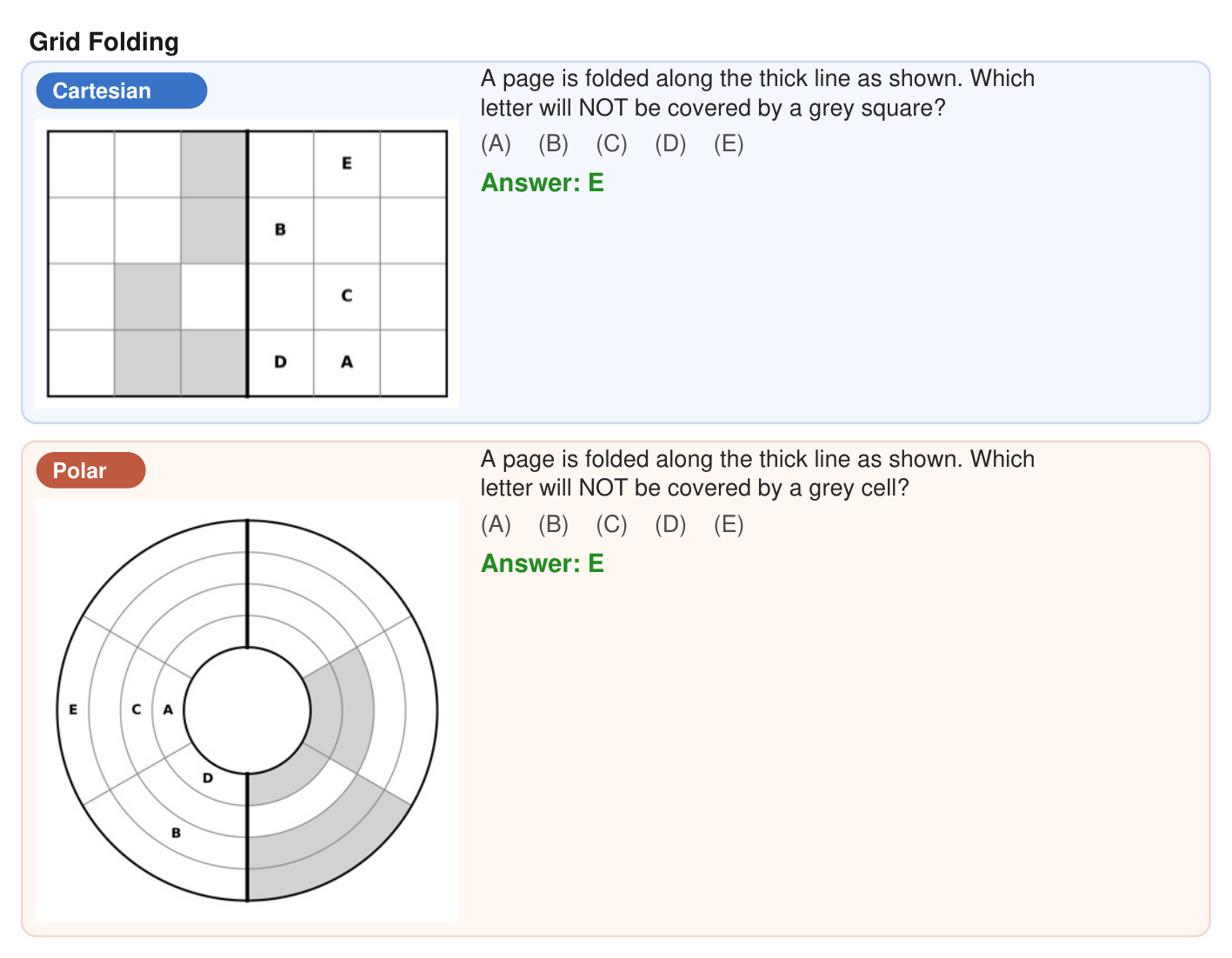}
    \caption{Example from grid folding task.}
    \label{fig:placeholder}
\end{figure}

\begin{figure}[h]
    \centering
    \includegraphics[width=\linewidth]{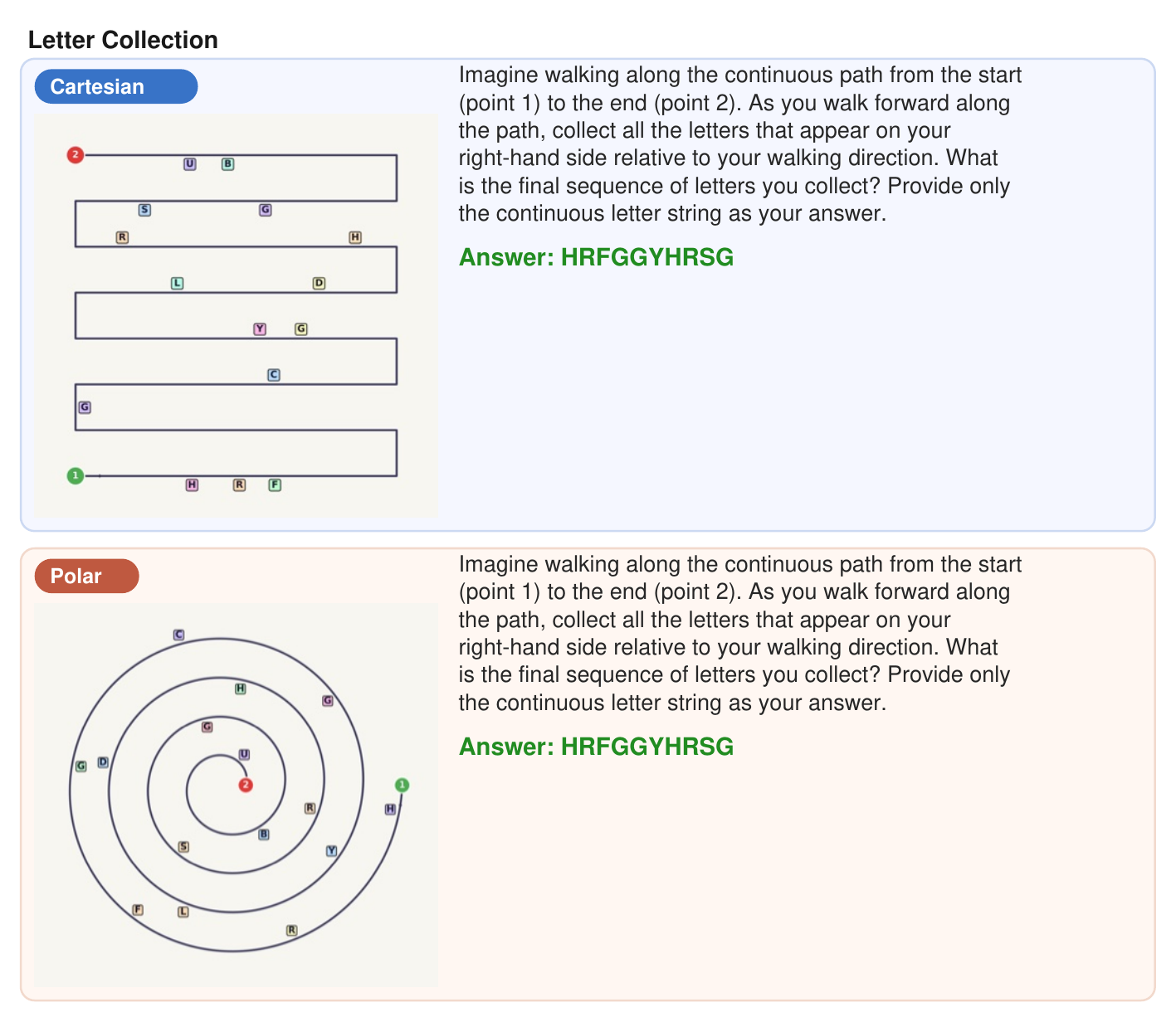}
    \caption{Example from letter collection task.}
    \label{fig:suppl_example_letter_collection}
\end{figure}

\begin{figure}[h]
    \centering
    \includegraphics[width=\linewidth]{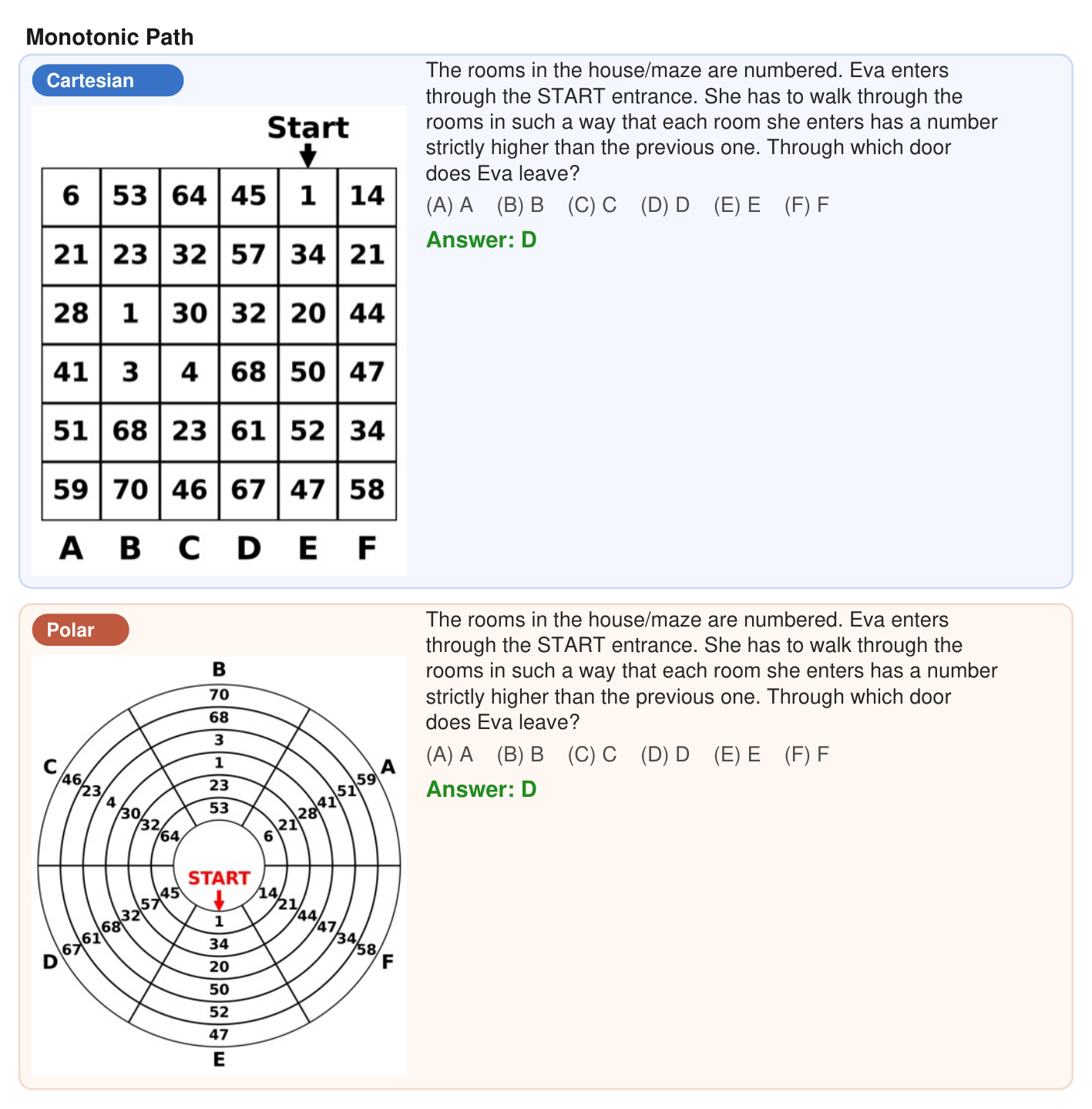}
    \caption{Example from monotonic path task.}
    \label{fig:suppl_example_wraping_pathfinding}
\end{figure}

\begin{figure}[h]
    \centering
    \includegraphics[width=\linewidth]{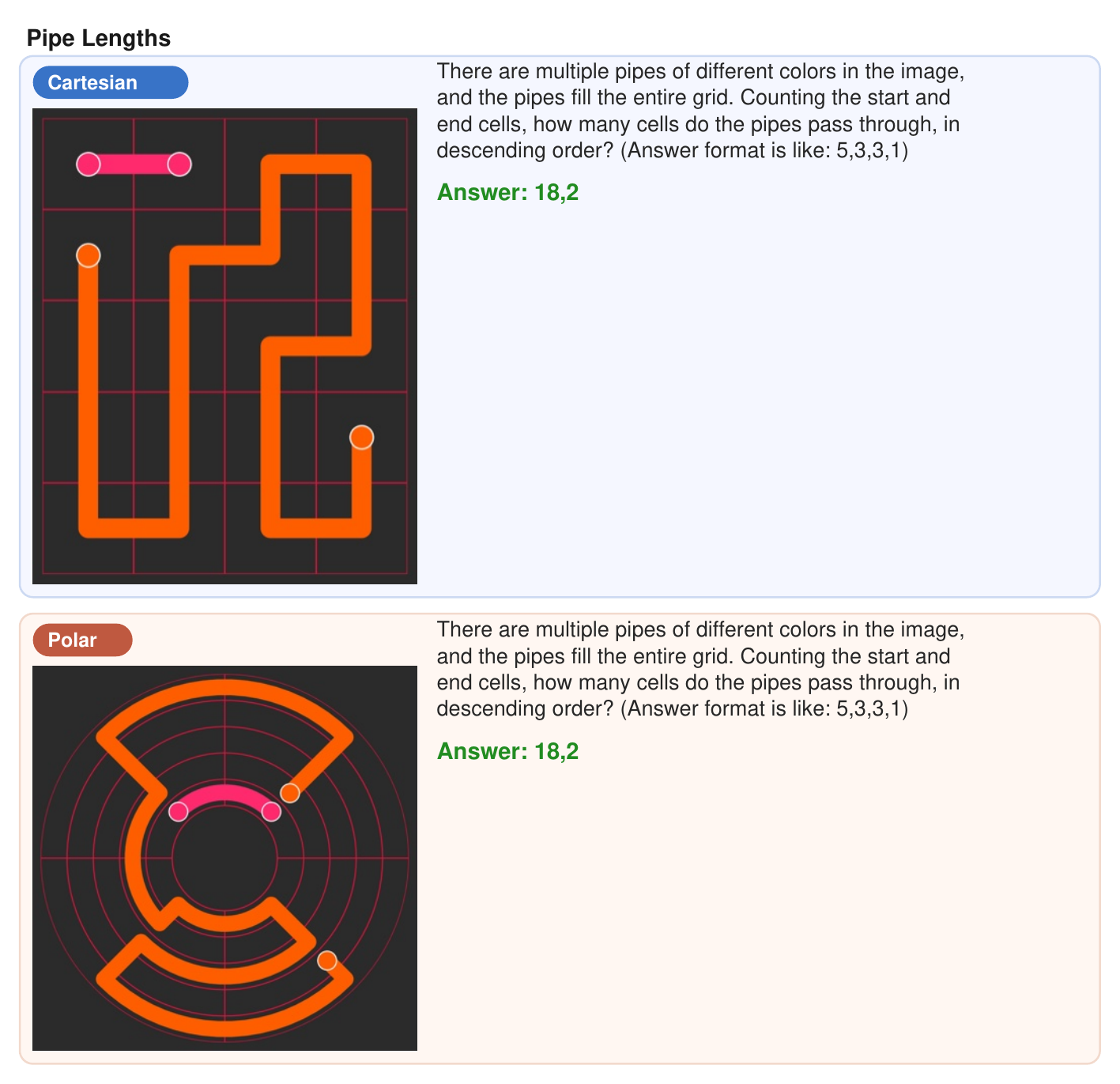}
    \caption{Example from pipe lengths task.}
    \label{fig:suppl_example_pipe_length}
\end{figure}

\begin{figure}[h]
    \centering
    \includegraphics[width=\linewidth]{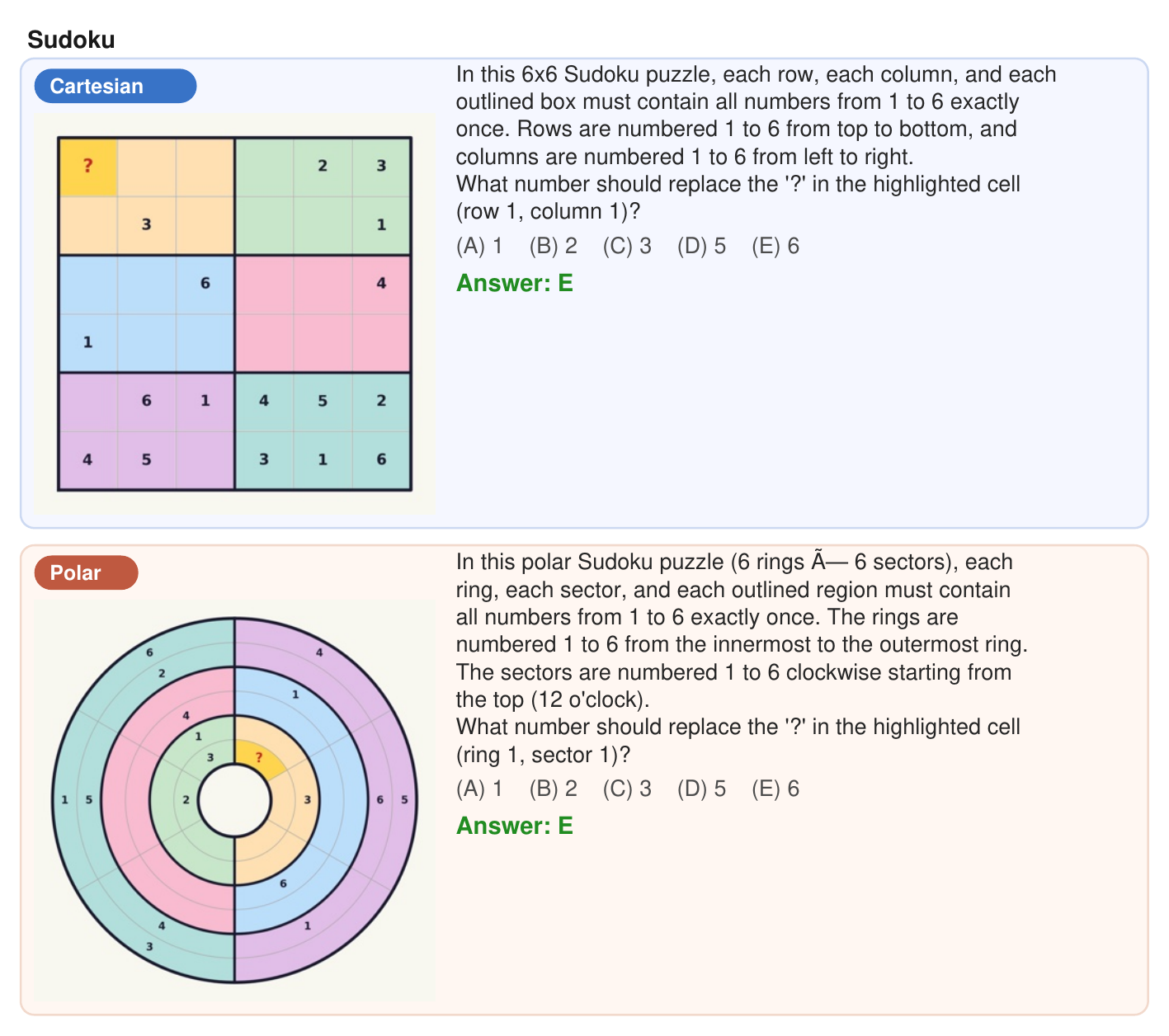}
    \caption{Example from sudoku task.}
    \label{fig:suppl_example_sudoku}
\end{figure}

\begin{figure}[h]
    \centering
    \includegraphics[width=\linewidth]{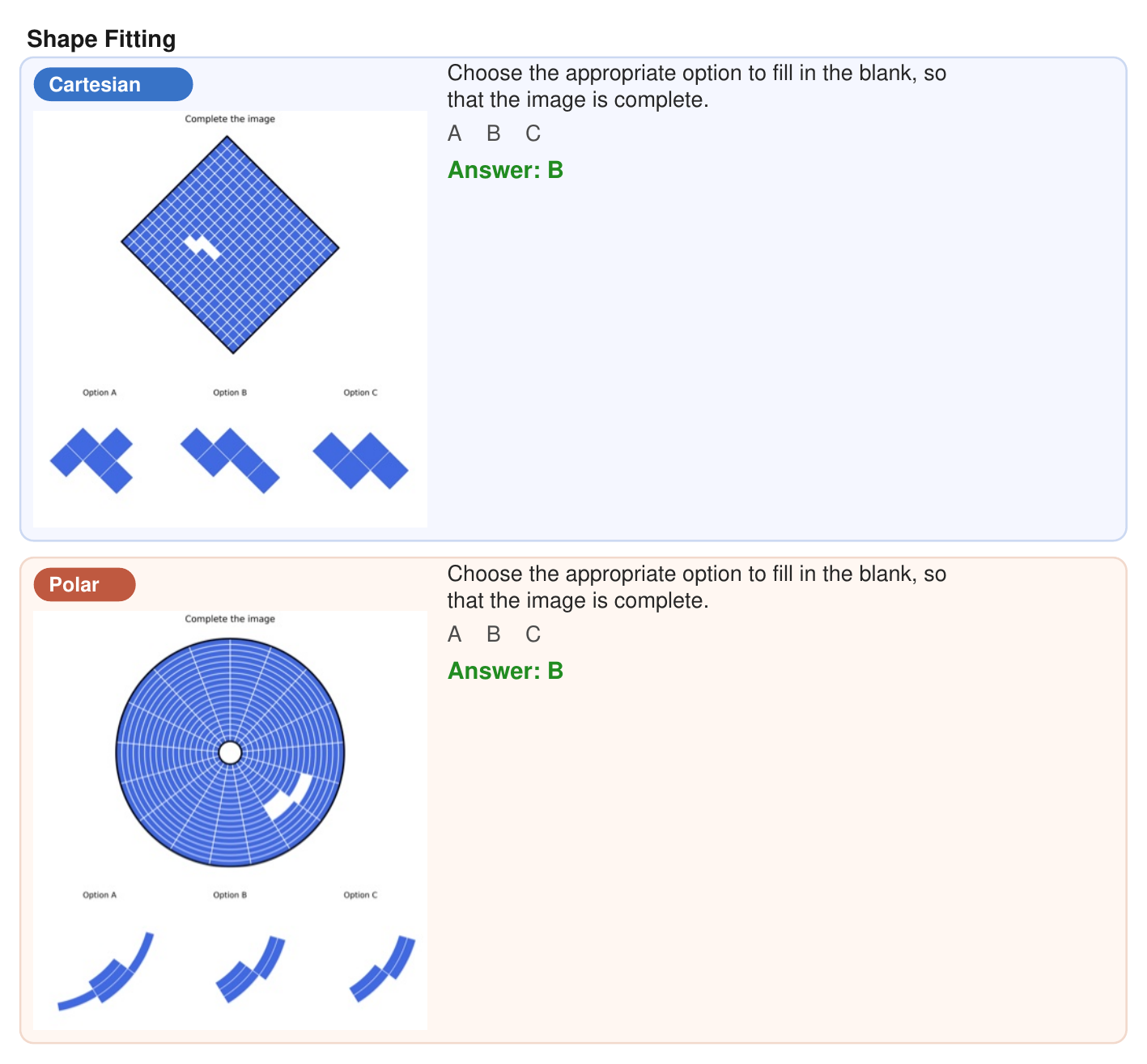}
    \caption{Example from shape fitting task.}
    \label{fig:suppl_example_shape_fitting}
\end{figure}

\begin{figure}[h]
    \centering
    \includegraphics[width=\linewidth]{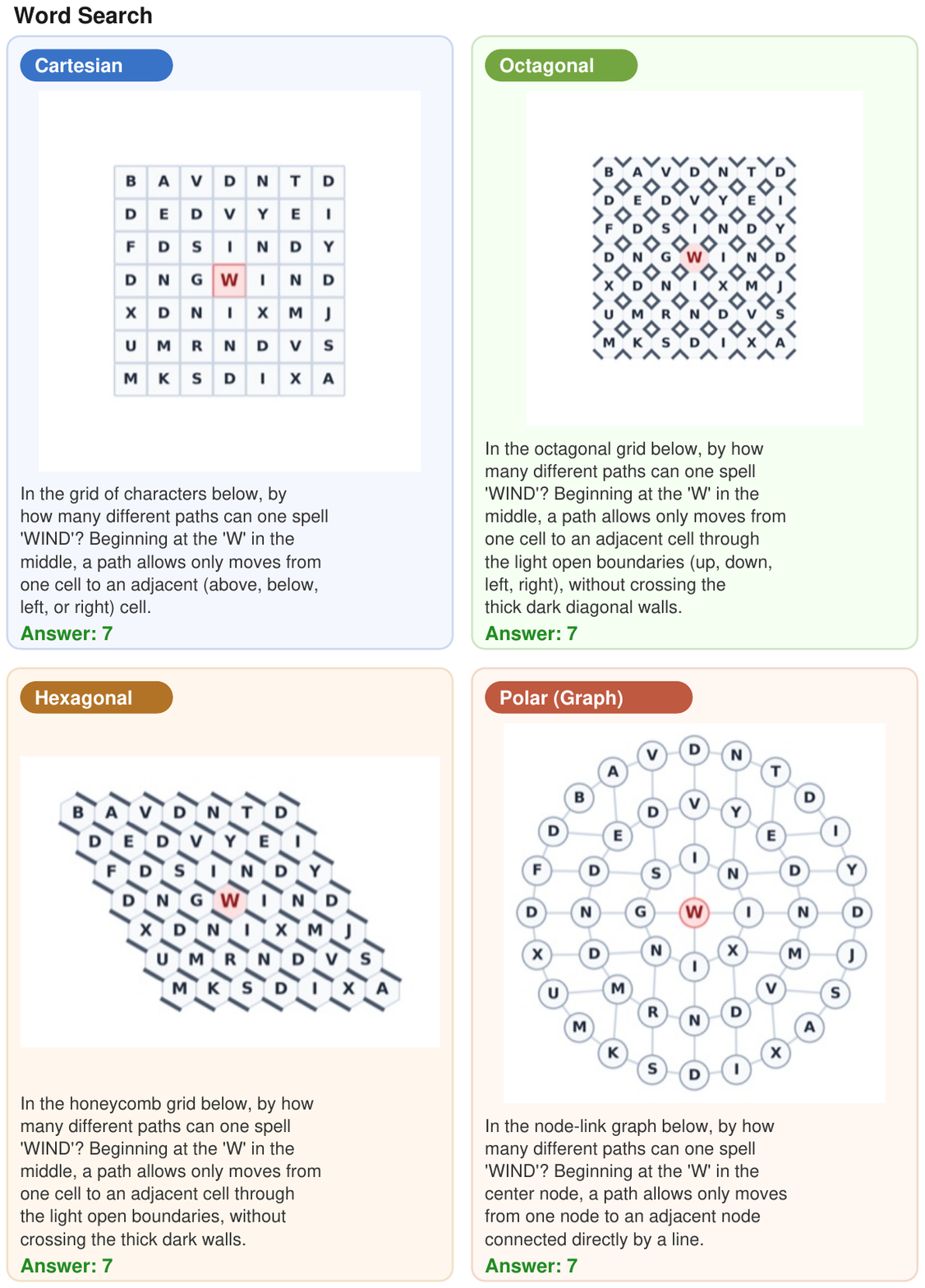}
    \caption{Example from word search task.}
    \label{fig:suppl_example_word_search}
\end{figure}

\begin{figure}[h]
    \centering
    \includegraphics[width=\linewidth]{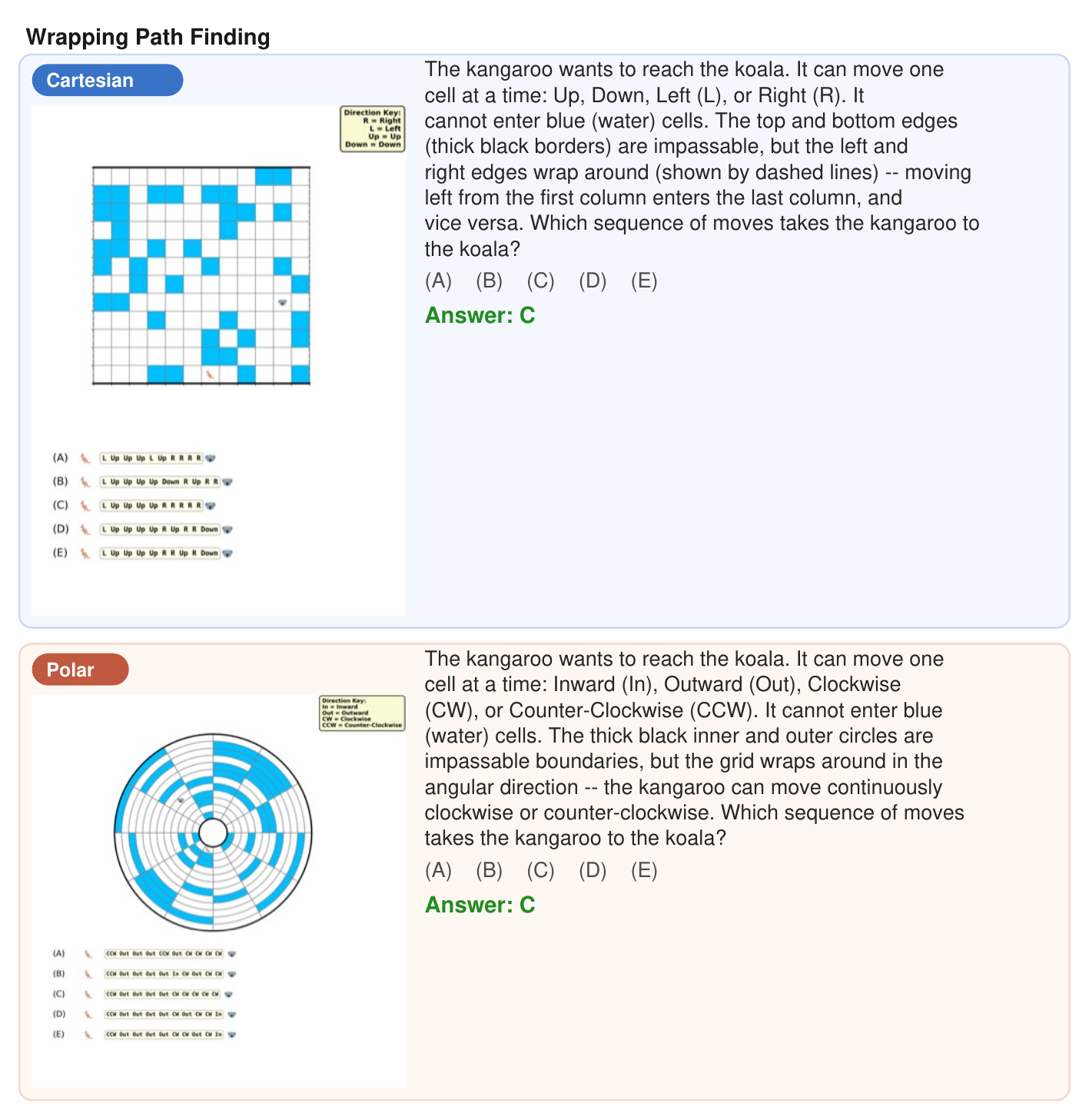}
    \caption{Example from wrapping path finding task.}
    \label{fig:suppl_example_wraping_pathfinding}
\end{figure}

\begin{figure}[h]
    \centering
    \includegraphics[width=\linewidth]{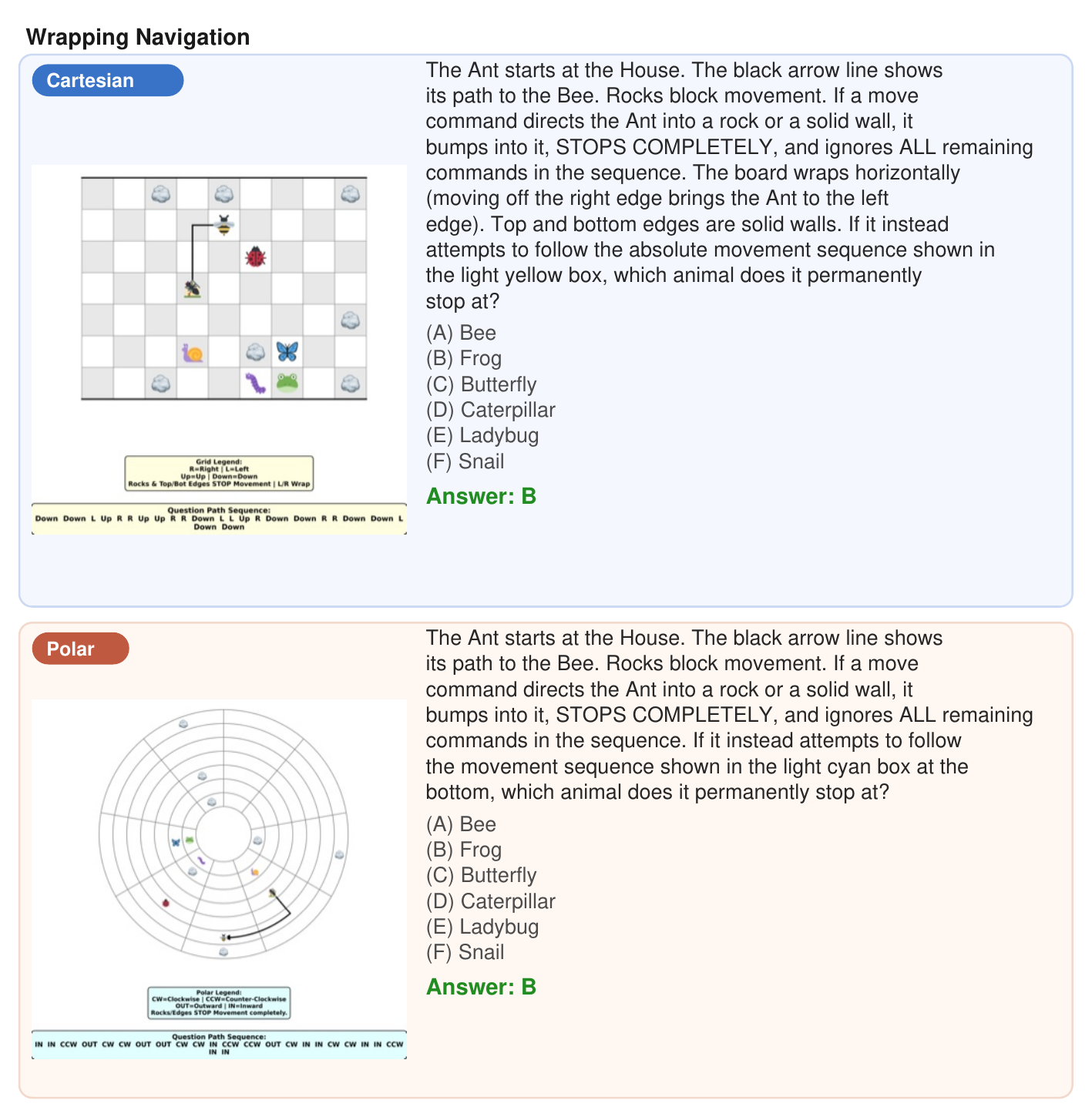}
    \caption{Example from wrapping navigation task.}
    \label{fig:suppl_example_wrapping_navigation}
\end{figure}

\end{document}